\PassOptionsToPackage{table}{xcolor}
\documentclass{article}

\PassOptionsToPackage{numbers, compress}{natbib}

 \usepackage[eandd, final]{neurips_2026}

\usepackage[utf8]{inputenc} %
\usepackage[T1]{fontenc}    %
\usepackage{hyperref}       %
\usepackage{url}            %
\usepackage{booktabs}       %
\usepackage{amsfonts}       %
\usepackage{nicefrac}       %
\usepackage{microtype}      %
\usepackage{xcolor}         %
\usepackage{tikz}
\usepackage{subcaption}
\usetikzlibrary{arrows}
\usepackage{adjustbox}
\usepackage{amsmath}
\usepackage{enumitem}
\usepackage{amssymb}
\usepackage{wrapfig}
\usepackage{float}

\usepackage{pifont}

\usepackage[colorinlistoftodos,prependcaption,textsize=tiny]{todonotes}

\usepackage[table]{xcolor}

\usepackage{listings}
\usepackage[most]{tcolorbox}

\lstdefinelanguage{Markdown}{
    basicstyle=\ttfamily\small,
    breaklines=true,
    breakatwhitespace=true,
    columns=fullflexible,
    keepspaces=true,
    showstringspaces=false,
    morekeywords={*, \#, \#\#, \#\#\#, -, >},
    keywordstyle=\color{blue}\bfseries,
}

\title{

Benchmarking Open-Ended Multi-Agent Coordination in Language Agents }

\author{%
\begin{tabular}{@{}c@{}}
Kale-ab Abebe Tessera$^{1}$\thanks{Correspondence to: \texttt{k.tessera@ed.ac.uk}}
\quad
Andras Szecsenyi$^{1}$ \quad
Cameron Barker$^{1}$ \\[0.4em]
Alexander Rutherford$^{2}$ \quad
Davide Paglieri$^{3}$ \quad
Aidan Scannell$^{1}$ \\[0.4em]
Henry Gouk$^{1}$ \quad
Elliot J. Crowley$^{1}$ \quad
Tim Rocktäschel$^{3}$ \quad
Amos Storkey$^{1}$ \\[0.6em]
{\normalfont
$^{1}$University of Edinburgh \quad
$^{2}$University of Oxford \quad
$^{3}$University College London}
\end{tabular}%
}

\begin{document}

\maketitle

\begin{abstract}
  As language models are increasingly deployed as autonomous agents, they must coordinate with others over long horizons in open-ended interactive tasks. Yet existing evaluations rarely test these demands together, focusing instead on single-agent tasks, short interactions, or highly structured multi-agent settings. We introduce \textsc{alem}, a JAX-based benchmark for open-ended multi-agent coordination built on Craftax-like dynamics. \textsc{alem} embeds procedurally generated coordination tasks, soft specialisation, communication, and controllable coordination difficulty into a long-horizon survival world with exploration, crafting, trading, and combat. We evaluate $13$ LLMs zero-shot within homogeneous teams, with trained MARL agents as reference points. Current LLM agents remain far from solving \textsc{alem}, averaging only $\sim6\%$ normalised return, but their failures are not uniform. On the hardest coordination setting, zero-shot Gemini-3.1-Pro-High approaches MARL agents trained for one billion steps, while GPT-5.4-High achieves strong base-task reward but much lower coordination reward. This contrast shows that individual task competence does not imply coordination competence. Further experiments reveal that communication is the largest contributor to coordination, while memory and reasoning help when used to maintain multi-step plans. In heterogeneous teams, however, combining different models improves base-task progress but lowers coordination reward relative to their homogeneous counterparts. Overall, our results identify coordination as a distinct bottleneck for frontier LLM agents, separate from single-agent capabilities. \textsc{alem} makes this bottleneck measurable and provides a controlled testbed for developing agents that communicate, allocate roles, and execute shared plans. Code is available at \url{https://github.com/alem-world/alem-env}, with checkpoints, LLM traces and a leaderboard at \url{https://alem-world.github.io}.  
\end{abstract}

\section{Introduction}

\newcommand{\alemTwo}{\textcolor[HTML]{4FB4FF}{$\bullet$}}
\newcommand{\alemAll}{\textcolor[HTML]{E03030}{$\blacksquare$}}
\newcommand{\alemHo}{\textcolor[HTML]{B5732C}{$\blacklozenge$}}

\begin{figure}[t]
  \centering
  \begingroup
  \newcommand{\alemWorldW}{0.305\linewidth}
  \newcommand{\alemZoomW}{0.232\linewidth}

  \makebox[\linewidth]{%
    \makebox[\alemWorldW]{\normalsize Overworld}\hfill
    \makebox[\alemWorldW]{\normalsize Gnomish Mines}\hfill
    \makebox[\alemWorldW]{\normalsize Fire Realm}%
  }\\[0.5pt]

  \makebox[\linewidth]{%
    \includegraphics[width=\alemWorldW]{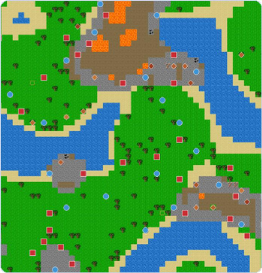}\hfill
    \includegraphics[width=\alemWorldW]{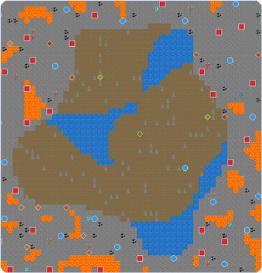}\hfill
    \includegraphics[width=\alemWorldW]{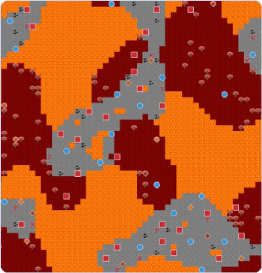}%
  }\\[2pt]

  \makebox[\linewidth]{%
    \makebox[\alemZoomW]{\normalsize Mining}\hfill
    \makebox[\alemZoomW]{\normalsize Construction}\hfill
    \makebox[\alemZoomW]{\normalsize Combat}\hfill
    \makebox[\alemZoomW]{\normalsize Crafting}%
  }\\[0.5pt]

  \makebox[\linewidth]{%
    \includegraphics[width=\alemZoomW]{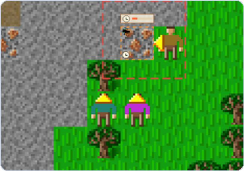}\hfill
    \includegraphics[width=\alemZoomW]{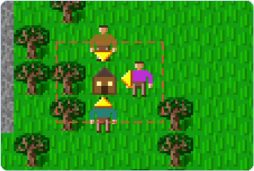}\hfill
    \includegraphics[width=\alemZoomW]{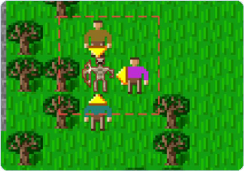}\hfill
    \includegraphics[width=\alemZoomW]{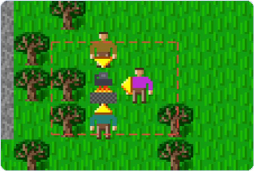}%
  }
  \endgroup

  \caption{\textbf{\textsc{alem} extends Craftax-like open worlds~\citep{matthews2024craftax,omari2025multi} into a controllable multi-agent coordination benchmark.}
  Top: procedurally generated levels with sampled coordination tasks (\alemTwo~2-agent sync, \alemAll~all-agent sync, \alemHo~handover). Bottom: coordinated mining, construction, combat, and crafting examples. By resampling tasks and coordination structure each episode, \textsc{alem} evaluates agents' ability to infer coordination needs from observation rather than memorise fixed task patterns.}
  \label{fig:alem-hero}
\end{figure}

Large Language Models (LLMs) have evolved rapidly, from systems primarily used for text generation~\citep{brown2020language,vaswani2017attention,devlin2019bert} to agentic systems capable of completing \textit{long-horizon}, \textit{open-ended tasks}~\citep{kwa2025measuring,yamada2025ai,chan2025mle,lu2026towards,nakano2021webgpt,wang2023voyager}. As these systems are deployed as autonomous agents, they will increasingly operate in multi-agent settings where success depends on coordinating with others~\citep{hammond2025multi}. Yet the mechanisms required for sustained multi-agent coordination remain poorly understood, especially when agents must communicate intent, allocate roles, maintain shared plans, and align actions over long horizons~\citep{guo2024large}. This creates a need to evaluate whether LLM agents can coordinate effectively in long-horizon, open-ended multi-agent environments, and to characterise the failures that arise when they cannot.

Existing benchmarks cover only part of this space. MultiAgentBench, Collab-Overcooked, and related multi-agent LLM benchmarks evaluate collaboration, but typically in shorter-horizon and more narrowly specified tasks~\citep{zhu2025multiagentbench,sun2025collab,agashe2025llm,wang2024battleagentbench,grotschla2025agentsnet,meireles2025the}. Others focus on long-horizon task completion in primarily single-agent settings~\citep{paglieri2025balrog,zhang2026deepplanning,sinha2026the,wang2025odysseybench,zhao2026ama,he2026texttt,grady2026kellybench}, or include multiple agents but lack explicit coordination tasks or controllable coordination difficulty~\citep{omari2025multi,ye2025efficient,yang2025llmpowered}. The community therefore still lacks evaluations that combine \textit{long-horizon} interaction and \textit{open-ended} task structure with \textit{explicit, procedurally generated, and controllable coordination demands}. Games are established testbeds for sequential decision-making~\citep{bellemare2013arcade,brown2018superhuman,schrittwieser2020mastering,vinyals2019grandmaster,campbell2002deep}, and procedural generation is especially useful because it reduces memorisation and contamination while enabling controlled variation across tasks and difficulty levels~\citep{paglieri2025balrog,shojaee2025illusion}.

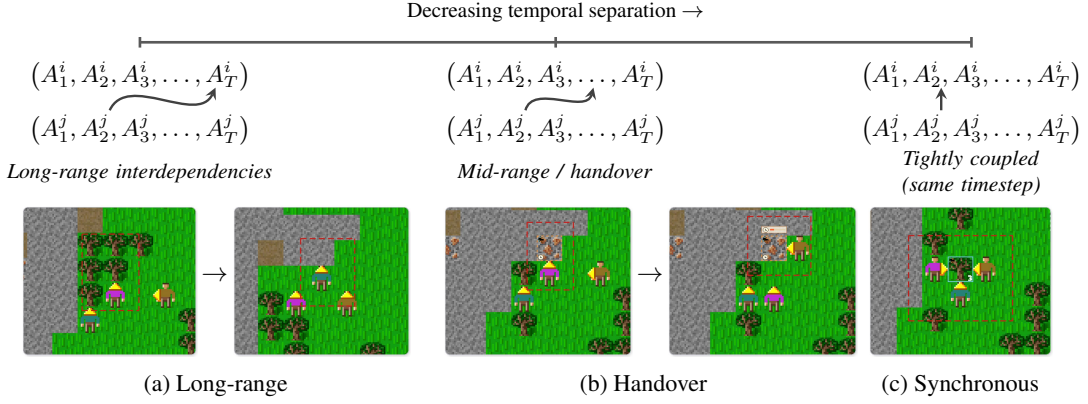
\begin{figure}[t]
\centering
\captionsetup[subfigure]{justification=centering,singlelinecheck=false}
\def\imgheight{2.0cm}

\begin{tikzpicture}[
    font=\small,
    >=stealth,
    every node/.style={inner sep=1pt}
]

\draw[line width=1pt, black!65] (0, 1.5) -- (11, 1.5);
\draw[line width=1pt, black!65] (0, 1.43) -- (0, 1.57);
\draw[line width=1pt, black!65] (5.5, 1.43) -- (5.5, 1.57);
\draw[line width=1pt, black!65] (11, 1.43) -- (11, 1.57);

\node[font=\footnotesize] at (5.5, 1.9)
  {Decreasing temporal separation $\rightarrow$};

\node[anchor=center] at (0, 1.05) {$\left(A^i_1, A^i_2, A^i_3, \ldots, A^i_T\right)$};
\node[anchor=center] at (0, 0.35) {$\left(A^j_1, A^j_2, A^j_3, \ldots, A^j_T\right)$};
\draw[->, line width=0.9pt, black!75]
  (-0.4, 0.60) to[out=65, in=-115] (1.0, 0.90);
\node[text width=3.5cm, align=center, font=\footnotesize] at (0, -0.23)
  {\textit{Long-range interdependencies}};

\node[anchor=center] at (5.5, 1.05) {$\left(A^i_1, A^i_2, A^i_3, \ldots, A^i_T\right)$};
\node[anchor=center] at (5.5, 0.35) {$\left(A^j_1, A^j_2, A^j_3, \ldots, A^j_T\right)$};
\draw[->, line width=0.9pt, black!75]
  (5.1, 0.6) to[out=75, in=-105] (6.0, 0.90);
\node[text width=3.5cm, align=center, font=\footnotesize] at (5.5, -0.23)
  {\textit{Mid-range / handover}};

\node[anchor=center] at (11, 1.05) {$\left(A^i_1, A^i_2, A^i_3, \ldots, A^i_T\right)$};
\node[anchor=center] at (11, 0.35) {$\left(A^j_1, A^j_2, A^j_3, \ldots, A^j_T\right)$};
\draw[->, line width=0.9pt, black!75] (10.6, 0.6) -- (10.6, 0.90);
\node[text width=3.5cm, align=center, font=\footnotesize] at (11, -0.23)
  {\textit{Tightly coupled (same timestep)}};

\end{tikzpicture}

\vspace{2pt}

\subcaptionbox{Long-range\label{fig:coord-lr}}[0.40\linewidth][c]{%
  \includegraphics[height=\imgheight]{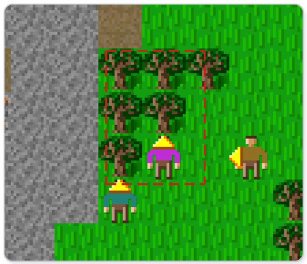}%
  \,\raisebox{1.0cm}{$\rightarrow$}\,%
  \includegraphics[height=\imgheight]{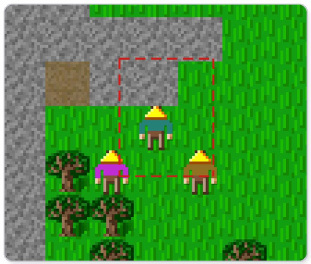}%
}%
\hfill
\subcaptionbox{Handover\label{fig:coord-ho}}[0.38\linewidth][c]{%
  \includegraphics[height=\imgheight]{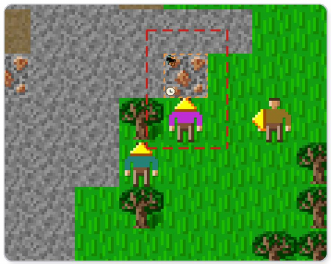}%
  \,\raisebox{1.0cm}{$\rightarrow$}\,%
  \includegraphics[height=\imgheight]{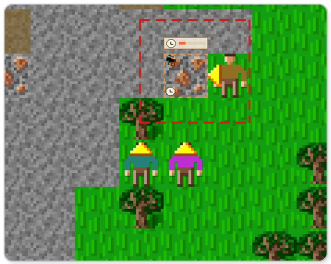}%
}%
\hfill
\subcaptionbox{Synchronous\label{fig:coord-sy}}[0.19\linewidth][c]{%
  \includegraphics[height=\imgheight]{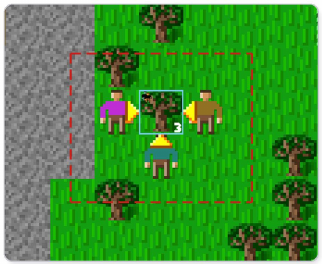}%
}
\caption{
\textbf{Coordination coupling spectrum in \textsc{alem}.}
\textsc{alem} tasks vary the temporal separation $\Delta t = |t - t'|$ between coupled actions $A^i_t$ and $A^j_{t'}$. We show representative examples: \textit{long-range}~(\ref{fig:coord-lr}, large $\Delta t$), where one agent gathers wood used to craft a pickaxe, enabling another agent to mine stone later; \textit{handover}~(\ref{fig:coord-ho}, small $\Delta t$), where agent~$i$ initiates a mining task that agent~$j$ must complete within a short window; and \textit{synchronous}~(\ref{fig:coord-sy}, $\Delta t = 0$), where agents act on the same target at the same timestep. Dashed boxes mark the coordination-critical regions.
}
\label{fig:coordination_spectrum}
\end{figure}

To address these gaps, we introduce \textsc{alem}\footnote{Alem translates to "world" in Amharic.}(Fig.~\ref{fig:alem-hero}), a benchmark for open-ended\footnote{Following Craftax-style benchmarks~\citep{matthews2024craftax,omari2025multi}, we use \emph{open-ended} to mean a large, procedurally generated but finite goal space, not unbounded novel goals as in~\citet{pmlr-v235-hughes24a}.} multi-agent coordination built on Craftax-like dynamics~\citep{matthews2024craftax,omari2025multi}. \textsc{alem} procedurally generates complex environments where coordination difficulty is modulated by a single parameter, ranging from loose temporal dependencies and handovers to tightly coupled joint actions (Fig.~\ref{fig:coordination_spectrum}). Information-theoretic diagnostics show that these mechanics induce genuine inter-agent dependence, not just added single-agent difficulty (App.~\ref{app:env_probing}). Implemented in JAX~\citep{jax2018github}, the benchmark facilitates scalable and reproducible evaluation of how multi-agent failure modes emerge as coordination demands increase.

Using \textsc{alem}, we evaluate 13 modern LLMs in zero-shot homogeneous teams, with trained MARL agents as reference points. Current LLM agents remain far from solving the benchmark, averaging only $\sim6\%$ normalised return. Gemini-3.1-Pro-High performs comparably to MARL agents trained for one billion environment steps on the hardest coordination setting ($17.5\%$ vs.\ $17.6\%$ Coord.\%), while GPT-5.4-High achieves strong base-task reward but much lower coordination reward. This contrast shows that individual task competence is not sufficient for coordinated progress. Ablations further show that communication has the largest effect on coordination performance, while memory and reasoning help when they support multi-step planning. Finally, in initial heterogeneous-team experiments, mixed teams score lower on coordination reward but higher on base reward than their constituent homogeneous baselines. Together, these results identify long-horizon, open-ended coordination as a distinct bottleneck for current LLM agents.

Our contributions are: \textbf{i}) \textsc{alem}, a JAX-based benchmark for long-horizon, open-ended multi-agent coordination with procedural tasks, soft specialisation, communication, and controllable difficulty; \textbf{ii}) pixel, symbolic, and text interfaces for evaluating MARL and zero-shot LLM agents under matched coordination demands; and \textbf{iii}) a systematic evaluation of 13 modern LLMs and 4 MARL baselines, showing that coordination remains a distinct bottleneck from base-task competence and exposing failure modes across communication, memory, reasoning, and team composition.

\section{Background}
\vspace{-0.1in}

We model the environment as a partially observable stochastic game (POSG)~\citep{hansen2004dynamic}, denoted as the tuple $\langle I, S, A, O, \Omega, T, R \rangle$. Here, $I = \{1, \dots, n\}$ is the set of agents, $S$ is the set of states, $A = A^1 \times \dots \times A^n$ is the joint action space, and $O = O^1 \times \dots \times O^n$ is the joint observation space. The observation function $\Omega = (\Omega^1, \dots, \Omega^n)$ maps states to per-agent observations: at each state $s \in S$, agent $i$ receives an individual observation $o^i = \Omega^i(s) \in O^i$ and selects an action $a^i \in A^i$. The environment transitions according to $T : S \times A \to \Delta(S)$, where $\Delta(\cdot)$ is the set of probability distributions over a set. Each agent receives an individual reward given by $R^i : S \times A \times S \to \mathbb{R}$.

\vspace{-0.1in}

\section{\textsc{alem}: An Open-Ended World for Multi-Agent Coordination}

We introduce \textsc{alem}, a benchmark for studying coordination in open-ended multi-agent worlds. Built on Craftax-Coop~\citep{omari2025multi}, \textsc{alem} extends prior work along four axes: diverse \emph{coordination types}, \emph{procedural generation} of coordination tasks, \emph{soft specialisation}, and an explicit \emph{communication} channel. Across nine procedurally generated levels, agents explore, trade resources, craft tools, build structures, and fight mobs under controllable coordination demands. By scaling these demands while preserving the underlying world structure, \textsc{alem} enables controlled evaluation of how coordination failures emerge as coordination demands increase. Information-theoretic diagnostics provide evidence that these mechanics induce inter-agent dependence beyond added single-agent difficulty~\citep{tessera2026probing} (App.~\ref{app:env_probing}), while our JAX implementation supports billion-step MARL training on a single GPU in under two days (App.~\ref{app:sps}).

\subsection{Coordination Types}\label{subsec:coordination}

Following \citet{malone1994interdisciplinary}, we view coordination as \textit{interdependence between activities}. In multi-agent systems, these activities correspond to agents' actions. We characterise the dependencies between coupled actions $A_t^i$ and $A_{t'}^j$ by their \emph{temporal separation}, $\Delta t = |t - t'|$, resulting in a \emph{coordination spectrum} from long-range dependencies to simultaneous joint action (Fig.~\ref{fig:coordination_spectrum}).

Existing MARL benchmarks typically isolate specific regions of this spectrum: SMAC~\citep{samvelyan2019starcraft} emphasises synchronous focus-fire ($\Delta t \approx 0$), Overcooked~\citep{carroll2019utility} tests short-window handovers (small $\Delta t$), and Melting Pot's Clean Up~\citep{agapiou2022melting} targets long-range resource interdependence (large $\Delta t$). \textsc{alem} instantiates this entire spectrum within a single open-ended environment, using procedural generation to vary both the temporal structure of coordination and the number of agents involved (details App.~\ref{app:coord_details}).

\subsection{Procedural Coordination Generation}

Existing environments procedurally generate spatial layouts to reduce memorisation and encourage generalisation~\citep{omari2025multi,matthews2024craftax,ye2025efficient,hafner2022benchmarking}. However, if coordination requirements are fixed, agents can still exploit recurring interaction patterns. \textsc{alem} therefore procedurally generates the \emph{coordination structure} itself.

Each episode, eligible entities across tasks such as mining, construction, combat, and crafting (Fig.~\ref{fig:alem-hero}) are independently assigned a coordination type (synchronous or handover) and a required agent count ($2$ to $N$). Synchronous tasks also adopt game-theoretic incentives: \emph{hard} tasks act as pure coordination problems requiring joint action for any reward~\citep{schelling1980strategy}, while \emph{soft} tasks function as assurance games, balancing a safe solo payoff against a larger joint-action bonus~\citep{skyrms2004stag} (details App.~\ref{app:coord_details}).

\subsection{Soft Specialisation and Difficulty Scaling}\label{subsec:difficulty}
Instead of enforcing rigid division of labour via hard role gates~\citep{agapiou2022melting,omari2025multi}, \textsc{alem} uses \emph{soft specialisation}. Any agent can attempt any role-gated action, but specialists succeed deterministically ($\eta_{\text{spec}}=1$) while non-specialists succeed with probability $\eta_{\text{non}} = 1 - \alpha$. This rewards effective role allocation while preserving the capacity for online adaptation. 

\begin{figure*}[t]
  \centering
  \raisebox{-0.5\height}{%
    \includegraphics[width=0.19\linewidth]{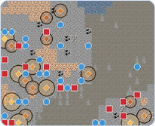}%
  }\hfill%
  \raisebox{-0.5\height}{%
    \includegraphics[width=0.19\linewidth]{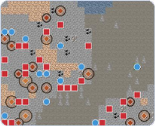}%
  }\hfill%
  \raisebox{-0.5\height}{%
    \includegraphics[width=0.19\linewidth]{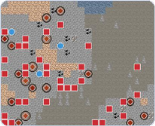}%
  }\hfill%
  \raisebox{-0.5\height}{%
    \includegraphics[width=0.38\linewidth]{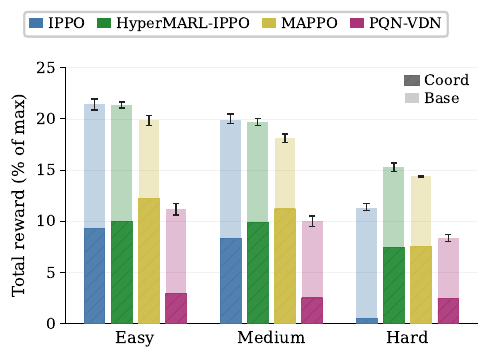}%
  }%
  \\[2pt]
  \makebox[0.19\linewidth]{\small Easy\;($\alpha{=}0.30$)}%
  \hfill%
  \makebox[0.19\linewidth]{\small Medium\;($\alpha{=}0.60$)}%
  \hfill%
  \makebox[0.19\linewidth]{\small Hard\;($\alpha{=}0.90$)}%
  \hfill%
  \makebox[0.38\linewidth]{}%

    \caption{
  \textbf{\textsc{alem} difficulty settings and MARL baselines.}
  \textbf{Left:} The same generated world under Easy, Medium, and Hard settings, with coordination sites marked as \alemTwo~2-agent sync, \alemAll~all-agent sync, and \alemHo~handover. Increasing $\alpha$ preserves the layout and coordination opportunities, but tightens execution constraints through more all-agent requirements, shorter handover windows, and stronger specialisation. \textbf{Right:} 1B-step trained MARL baseline performance, reported as mean percentage of max achievable reward with 95\% CIs and decomposed into coordination (dark) and base (light) reward.
}

  \label{fig:difficulty-and-rl}
\end{figure*}

We scale overall coordination difficulty with the single scalar $\alpha \in [0,1]$. Because world generation in JAX is deterministic, varying $\alpha$ preserves the spatial layout and frequency of coordination opportunities (Fig.~\ref{fig:difficulty-and-rl}), cleanly isolating execution difficulty from world variation. Higher $\alpha$ tightens execution constraints: it increases the probability of $N$-agent requirements, shortens handover windows, increases solo failure rates on soft tasks, and lowers non-specialist efficiency. We evaluate on Easy ($\alpha=0.3$), Medium ($\alpha=0.6$), and Hard ($\alpha=0.9$) settings (parameters detailed in App.~\ref{app:coordination_difficulty}).

\subsection{Environment Interface}

\subsubsection{RL Interface}
\paragraph{Observation space.} \textsc{alem} exposes both pixel (see Fig.~\ref{fig:alem_pixel_obs}, App~\ref{app:env_details}) and symbolic observations of the underlying state. Each egocentric observation includes a local map, inventory, and a teammate dashboard (health, specialisation, requested resources, off-screen direction, and recent communications). Beyond Craftax-Coop, \textsc{alem} explicitly embeds communication messages, coordination states, including soft/hard requirements, handover countdowns, and multi-agent mob targets.

\paragraph{Action space.} Agents act in a 60-dimensional discrete action space encompassing environment manipulation, crafting, exploration, and combat. To facilitate teamwork, agents can issue persistent resource requests (active for up to 10 timesteps), execute teammate-targeted giving, and broadcast discrete messages. The primary \textsc{Do} action is contextual, it supports both environment interaction and direct agent-to-agent mechanics, such as reviving downed teammates or triggering friendly fire. We apply conservative action masking to filter strictly infeasible choices.

\paragraph{Achievements and rewards.} 

Following prior environments~\citep{matthews2024craftax,omari2025multi}, \textsc{alem} pairs a dense health reward with sparse, first-time rewards for completing a hierarchy of achievements. We introduce 27 novel coordination achievements spanning synchronised mining, temporal handovers, cooperative construction, and coordinated combat. Rewards are individual by default; coordination bonuses are awarded only to participating agents by default (details in App.~\ref{app:rewards_achievements}).

\subsubsection{Text-based Language Interface}\label{subsec:text_interface}

To evaluate language agents, we expose a structured text interface mapped directly to the underlying environment (details in App.~\ref{app:text_interface}). At each timestep, models receive a textual state containing recent observation-action history, teammate messages, a private scratchpad, and visible coordination constraints, together with explicit action affordances. Models must return a structured response containing exactly one valid action and, optionally, reasoning, communication, and scratchpad fields. This parsed action is then mapped to the corresponding discrete environment step. A sample prompt is provided in App.~\ref{app:prompt_examples}.

\subsection{MARL Baselines}\label{sec:marl_baselines}

We train IPPO~\citep{de2020independent}, MAPPO~\citep{yu2022surprising}, PQN-VDN~\citep{gallicisimplifying}, and HyperMARL-IPPO~\citep{tessera2025hypermarl} to provide reference points for zero-shot LLM performance and to verify that the difficulty settings meaningfully affect learned policies. Figure~\ref{fig:difficulty-and-rl} reports 1B-step performance (hyperparameter sweeps detailed in App.~\ref{app:hyperparameters}). Performance sharply degrades as coordination difficulty increases; even after 3B environment steps, the strongest baseline reaches only 18\% of the maximum total reward on the Hard setting (App.~\ref{app:rl-3b-long}). These results show that \textsc{alem} remains an unsaturated challenge for current MARL methods.

\section{Experiments}

We evaluate 13 modern LLMs in zero-shot homogeneous teams, with 4 trained MARL baselines as reference points. Our goal is to measure coordination in long-horizon, open-ended environments, while separating coordination difficulty from the difficulty of the underlying world. To calibrate this distinction, we compare \textsc{alem} against a single-agent baseline and find that its multi-agent dynamics result in a drop in performance, showing that coordination introduces a distinct layer of difficulty (App.~\ref{app:env_calibration}).

Our experiments ask three questions:
\textbf{Q1:} How well do LLM agents coordinate zero-shot within homogeneous teams?
\textbf{Q2:} How do communication, memory, and reasoning affect coordination?
\textbf{Q3:} How well do agents coordinate in heterogeneous teams composed of different models?

\subsection{Experimental Setup}

\paragraph{Evaluation protocol.}
We follow a zero-shot evaluation protocol similar to BALROG~\citep{paglieri2025balrog}. At each timestep, each agent receives the game rules, environment details, local observation, and legal actions in text form through the interface described in Sec.~\ref{subsec:text_interface}. By default, agents have scratchpad memory, broadcast communication, and a text history containing the last eight observations and actions. Success in \textsc{alem} requires spatial reasoning, long-horizon planning, tracking crafting dependencies, memory, effective communication, and coordination under partial observability.

We provide the rules so that agents can plan and coordinate without prior knowledge of Craftax. Models may nevertheless recognise crafting recipes or progression steps from pretraining. To test how far base-game knowledge alone can take them, we compare three disclosure settings in App.~\ref{app:prompt-disclosure}. On Hard, Gemma-4-31B-it scores only 0.7 Coord.\% when given the full Craftax rules, compared with 8.8 when it also receives \textsc{alem}'s coordination rules and target-specific cues during play. Coordination performance therefore depends on \textsc{alem}-specific information that pretraining on Craftax cannot provide.

\paragraph{Metrics.}\label{subsec:metrics}
Our primary metric is normalised team achievement reward, reported as a percentage of the maximum achievable reward. Each achievement contributes its reward once per episode if any agent unlocks it. Following previous Craftax-based environments~\citep{omari2025multi,matthews2024craftax}, we normalise by the maximum reward for the relevant category and average scores across episodes. We report Base\%, Coord.\%, and Total\%, each normalised independently, so Total\% is not the sum of Base\% and Coord.\%. Reward bounds are given in App.~\ref{app:additional_metrics}.

\paragraph{Models.}

We evaluate 13 modern language models. Closed-source models are evaluated via their respective APIs (GPT-5.4~\citep{gpt54_2026} and Gemini 3.1 Pro~\citep{gemini31pro_2026}). Open-weight models are served with vLLM~\citep{kwon2023efficient} and include Gemma-4-E4B-it/26B-A4B-it/31B-it~\citep{gemma4_2026}, Llama-3.1-8B-Instruct and Llama-3.3-70B-Instruct~\citep{grattafiori2024llama}, Qwen-3.5-9B/27B/35B-A3B/122B-A10B~\citep{qwen35blog}, and Qwen-3.6-27B/35B-A3B~\citep{qwen36plus}. Where applicable, reasoning is enabled, with proprietary models configured to high-reasoning settings. We run 20 seeds per difficulty for open-weight models and 10 for closed-source models (due to API costs), reporting cross-seed means with 95\% stratified bootstrap confidence intervals using the rliable library~\citep{agarwal2021deep}.

\paragraph{Agent harness.}\label{subsec:harness}
Language models are evaluated with the agent harness described in Sec.~\ref{subsec:text_interface} and App.~\ref{app:text_interface}. The harness follows a reason-then-act paradigm similar to ReAct~\citep{yao2023react}, but generates a new prompt at each step focusing the evaluation on the reasoning and coordination capabilities of the model instead of confounding factors such as context length or compaction strategy.

\paragraph{Reference points.} We compare the zero-shot LLM teams with trained MARL agents, a privileged heuristic, and a human team. For each difficulty, we report the best-performing MARL algorithm from Section~\ref{sec:marl_baselines} at each of two training budgets, 100M and 1B environment steps. The privileged heuristic is a hand-designed policy with access to the full environment state and the coordination requirements, run on five seeds per difficulty to see how far a scripted team could get with this information (App.~\ref{app:oracle}). For the human pilot, three authors played as a team for one session on Easy (one-hour limit) and one on Hard (two-hour limit), with no session on Medium (App.~\ref{app:human_play}).

\setcounter{topnumber}{1}
\subsection{Results}

\subsubsection{\textbf{Q1:} How well do LLM agents coordinate zero-shot within homogeneous teams?}\label{subsec:q1}

\begin{figure*}[t]
    \centering
    \captionsetup[subfigure]{font=small,labelfont=bf,skip=2pt}
    \begin{subfigure}[t]{0.30\textwidth}
        \centering
        \includegraphics[height=4.5cm, width=\linewidth, keepaspectratio]{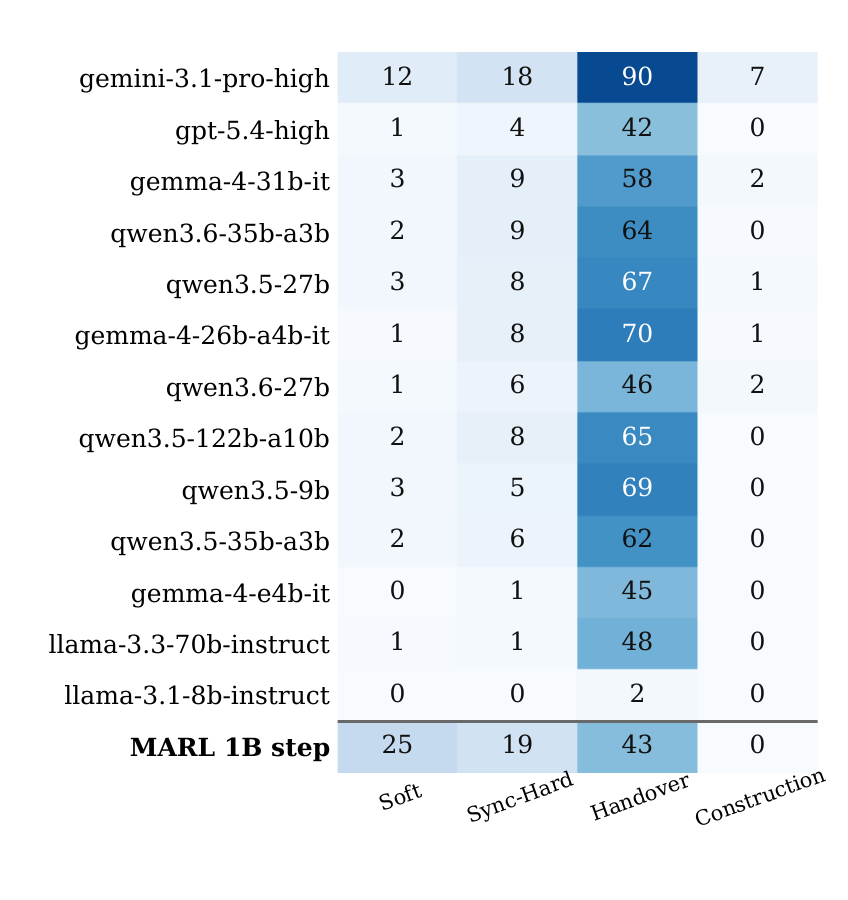}
        \caption{Coordination breakdown}
        \label{fig:main-diagnostic-coord}
    \end{subfigure}%
    \hspace{0.00\textwidth}%
    \begin{subfigure}[t]{0.20\textwidth}
        \raggedright
        \includegraphics[height=4.46cm, width=\linewidth, keepaspectratio]{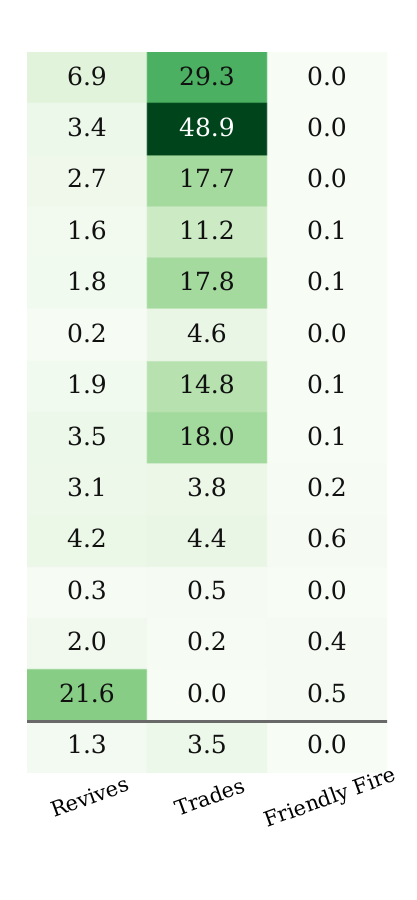}
        \caption{Cooperative events}
        \label{fig:main-diagnostic-coop}
    \end{subfigure}\hfill
    \begin{subfigure}[t]{0.5\textwidth}
        \centering
        \includegraphics[width=\linewidth]{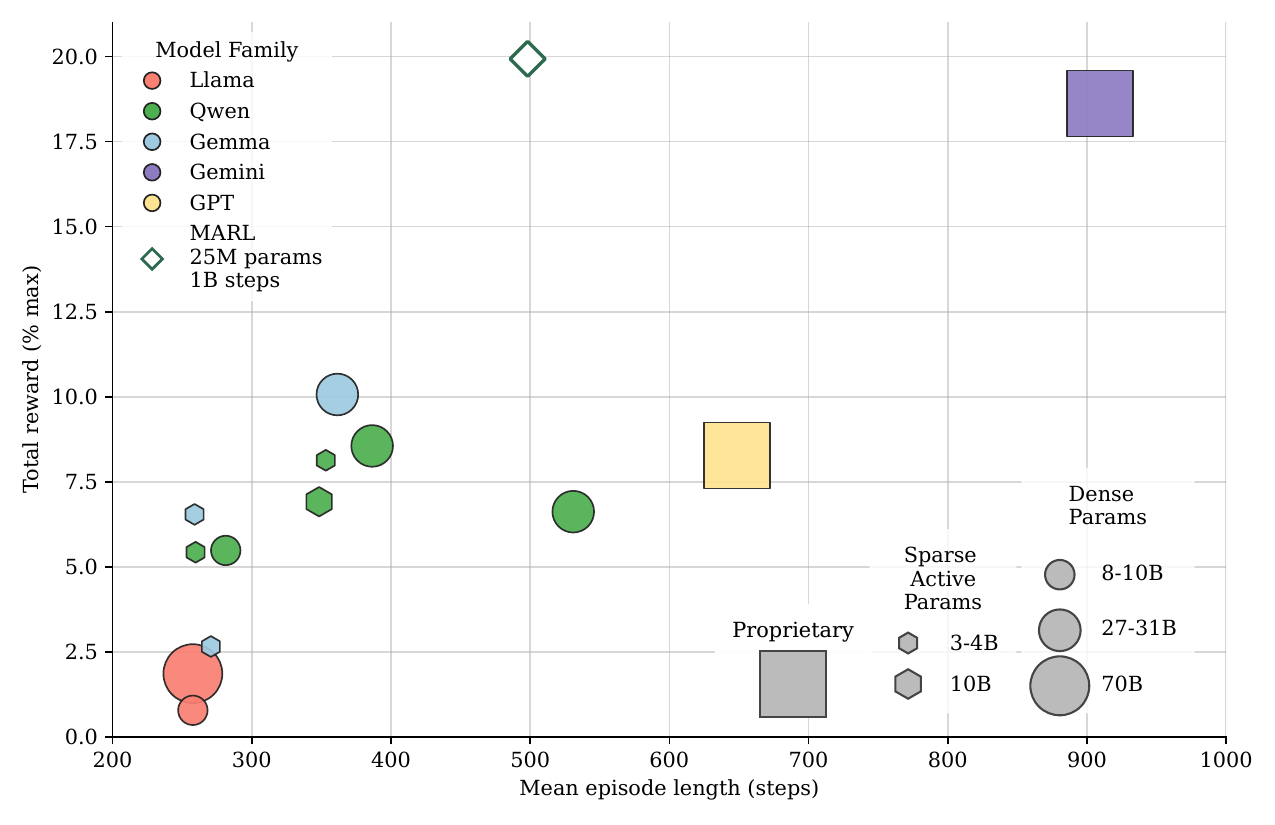}
        \caption{Survival vs. reward}
        \label{fig:main-diagnostic-survival}
    \end{subfigure}
    \caption{
        \textbf{Diagnostic breakdown of zero-shot homogeneous team failures in \textsc{alem}.}
        (\textbf{A}) Coordination reward coverage by coordination type, averaged across difficulties. Cell values report the percentage of the maximum attainable reward within each coordination category. 
        (\textbf{B}) Local cooperative event counts per evaluation across all difficulties.
        (\textbf{C}) Mean episode length versus Total\% reward on Medium difficulty, showing that longer survival does not necessarily translate into progression or coordinated achievements. Marker size and shape denote model parameter count and architecture type, respectively. MARL 1B is shown as a separated reference row in (A,B) and as a diamond in (C).
    }
    \label{fig:main-diagnostic-panel}
\end{figure*}

\paragraph{\textsc{alem} is unsolved but separates models.}
Table~\ref{tab:alem_results} shows that \textsc{alem} remains far from saturated by both zero-shot LLMs and trained MARL baselines, while separating models across a broad performance range. On Easy, Total\% spans from $0.8\%$ for Llama-3.1-8B-Instruct to $18.1\%$ for Gemini-3.1-Pro-High; on Hard, LLMs average only $6.2\%$ Total\%. Yet the top model, Gemini-3.1-Pro-High, reaches $17.5\%$ Coord.\% on Hard, comparable to the $17.6\%$ Coord.\% of MARL agents trained for one billion environment steps. The human pilot scored $22.1\%$ Total\% on Hard after only 455 of 10{,}000 steps and with no deaths, so the performance ceiling sits above every agent. Many models also survive for long episodes despite low returns (Fig.~\ref{fig:main-diagnostic-survival}), suggesting that the main bottleneck is not basic survival or interface parsing (parsing rates in Fig.~\ref{fig:action_parse}, App.~\ref{app:additional_metrics}), but converting interaction time into coordinated progress.

\begin{table}[t]
\centering
\captionsetup{position=top,skip=10pt}
\caption{\textbf{Zero-shot coordination of homogeneous teams in \textsc{alem}.}
Performance across Easy, Medium, and Hard difficulties, with trained MARL baselines, a privileged scripted policy, and a human pilot for comparison. LLMs are ordered by Total\% on Hard. We report mean team achievement reward with 95\% stratified bootstrap confidence intervals. Base\%, Coord.\%, and Total\% are normalised independently by the maximum achievable reward in their categories (Total\% is not a direct sum). MARL 100M and 1B are environment training steps, not model parameters; for each compute budget, we report the best-performing MARL algorithm. $^{\dagger}$ Hand-designed policy with full state access, five seeds per difficulty; a reference point, not an upper bound (App.~\ref{app:oracle}). $^{\ddagger}$ One session each on Easy and Hard by three authors, so no confidence intervals and no Medium entry (App.~\ref{app:human_play}).}
\label{tab:alem_results}
\begingroup
\definecolor{rowalt}{HTML}{F4F4F5}
\setlength{\tabcolsep}{3.6pt}
\renewcommand{\arraystretch}{1.03}
\footnotesize
\begin{adjustbox}{max width=\linewidth}
\begin{tabular}{@{}p{3.0cm} ccc ccc ccc@{}}
\toprule
 & \multicolumn{3}{c}{Easy} & \multicolumn{3}{c}{Medium} & \multicolumn{3}{c}{Hard} \\
\cmidrule(lr){2-4} \cmidrule(lr){5-7} \cmidrule(lr){8-10}
Method & Base\% & Coord.\% & Total\% & Base\% & Coord.\% & Total\% & Base\% & Coord.\% & Total\% \\
\midrule
gemini-3.1-pro-high & 18.1\,{\tiny [14.2,21.8]} & 18.0\,{\tiny [12.5,23.9]} & 18.1\,{\tiny [14.8,21.0]} & 15.1\,{\tiny [12.3,18.0]} & 23.4\,{\tiny [20.3,27.0]} & 18.6\,{\tiny [16.3,21.3]} & 13.8\,{\tiny [11.2,16.6]} & 17.5\,{\tiny [13.5,21.5]} & 15.4\,{\tiny [13.2,17.8]} \\
\rowcolor{rowalt}gpt-5.4-high & 14.1\,{\tiny [11.5,16.7]} & 7.0\,{\tiny [3.6,10.8]} & 11.1\,{\tiny [9.2,13.0]} & 10.0\,{\tiny [6.5,13.5]} & 4.0\,{\tiny [2.1,5.7]} & 7.4\,{\tiny [4.9,9.8]} & 11.8\,{\tiny [8.0,15.3]} & 4.2\,{\tiny [2.0,6.6]} & 8.6\,{\tiny [5.9,10.8]} \\
gemma-4-31b-it & 9.6\,{\tiny [7.4,11.9]} & 8.6\,{\tiny [6.4,11.0]} & 9.2\,{\tiny [7.6,10.8]} & 9.7\,{\tiny [8.0,11.3]} & 10.6\,{\tiny [8.3,13.0]} & 10.1\,{\tiny [9.0,11.1]} & 7.3\,{\tiny [5.9,8.9]} & 8.8\,{\tiny [6.9,10.9]} & 8.0\,{\tiny [6.7,9.1]} \\
\rowcolor{rowalt}gemma-4-26b-a4b-it & 4.3\,{\tiny [2.6,6.2]} & 7.3\,{\tiny [5.0,9.6]} & 5.6\,{\tiny [3.8,7.4]} & 4.1\,{\tiny [2.8,5.3]} & 6.1\,{\tiny [3.8,8.3]} & 4.9\,{\tiny [3.3,6.4]} & 5.9\,{\tiny [5.3,6.6]} & 9.5\,{\tiny [7.7,11.3]} & 7.4\,{\tiny [6.6,8.2]} \\
qwen3.6-27b & 5.1\,{\tiny [3.4,6.9]} & 4.7\,{\tiny [2.9,6.7]} & 4.9\,{\tiny [3.4,6.5]} & 7.0\,{\tiny [5.4,8.8]} & 6.1\,{\tiny [4.3,7.9]} & 6.6\,{\tiny [5.4,7.9]} & 7.5\,{\tiny [5.7,9.5]} & 7.2\,{\tiny [5.4,8.9]} & 7.4\,{\tiny [6.1,8.8]} \\
\rowcolor{rowalt}qwen3.6-35b-a3b & 2.8\,{\tiny [1.4,4.5]} & 4.4\,{\tiny [2.3,6.7]} & 3.5\,{\tiny [1.8,5.3]} & 6.3\,{\tiny [5.0,7.9]} & 9.7\,{\tiny [7.7,11.6]} & 7.7\,{\tiny [6.3,9.1]} & 6.3\,{\tiny [4.8,8.1]} & 8.7\,{\tiny [6.7,10.7]} & 7.3\,{\tiny [6.0,8.6]} \\
qwen3.5-27b & 4.8\,{\tiny [3.1,6.9]} & 7.0\,{\tiny [4.9,9.2]} & 5.7\,{\tiny [4.1,7.4]} & 4.1\,{\tiny [2.4,6.0]} & 6.5\,{\tiny [3.8,9.3]} & 5.1\,{\tiny [3.1,7.2]} & 5.1\,{\tiny [3.5,7.1]} & 8.1\,{\tiny [5.8,10.4]} & 6.4\,{\tiny [4.8,7.9]} \\
\rowcolor{rowalt}qwen3.5-122b-a10b & 5.4\,{\tiny [4.3,6.5]} & 8.4\,{\tiny [6.6,10.0]} & 6.6\,{\tiny [5.5,7.7]} & 5.0\,{\tiny [4.6,5.3]} & 9.6\,{\tiny [7.5,11.5]} & 6.9\,{\tiny [6.0,7.9]} & 4.7\,{\tiny [3.9,5.5]} & 8.3\,{\tiny [6.2,10.6]} & 6.2\,{\tiny [5.0,7.4]} \\
qwen3.5-35b-a3b & 1.9\,{\tiny [1.0,2.9]} & 4.2\,{\tiny [2.1,6.3]} & 2.9\,{\tiny [1.6,4.2]} & 2.2\,{\tiny [1.4,3.1]} & 4.7\,{\tiny [2.7,6.8]} & 3.3\,{\tiny [2.0,4.6]} & 2.9\,{\tiny [2.2,3.7]} & 5.3\,{\tiny [3.4,7.5]} & 4.0\,{\tiny [2.7,5.2]} \\
\rowcolor{rowalt}qwen3.5-9b & 2.7\,{\tiny [2.0,3.4]} & 7.0\,{\tiny [4.6,9.5]} & 4.6\,{\tiny [3.2,5.9]} & 2.7\,{\tiny [1.6,4.0]} & 4.7\,{\tiny [2.7,7.1]} & 3.6\,{\tiny [2.2,4.9]} & 2.6\,{\tiny [1.7,3.4]} & 5.0\,{\tiny [2.9,7.1]} & 3.6\,{\tiny [2.3,4.8]} \\
gemma-4-e4b-it & 2.4\,{\tiny [1.9,2.9]} & 4.6\,{\tiny [3.0,6.3]} & 3.3\,{\tiny [2.6,4.1]} & 2.4\,{\tiny [1.9,2.8]} & 3.1\,{\tiny [1.7,4.5]} & 2.7\,{\tiny [2.0,3.3]} & 2.3\,{\tiny [1.9,2.7]} & 3.2\,{\tiny [1.9,4.5]} & 2.7\,{\tiny [2.1,3.3]} \\
\rowcolor{rowalt}llama-3.3-70b-instruct & 1.6\,{\tiny [1.1,2.3]} & 5.0\,{\tiny [3.2,6.8]} & 3.1\,{\tiny [2.2,4.0]} & 1.1\,{\tiny [0.8,1.5]} & 2.9\,{\tiny [1.4,4.4]} & 1.9\,{\tiny [1.3,2.5]} & 1.4\,{\tiny [1.1,1.7]} & 4.2\,{\tiny [2.7,5.8]} & 2.6\,{\tiny [1.8,3.3]} \\
llama-3.1-8b-instruct & 1.4\,{\tiny [1.1,1.7]} & 0.0\,{\tiny [0.0,0.0]} & 0.8\,{\tiny [0.6,1.0]} & 1.4\,{\tiny [1.1,1.6]} & 0.0\,{\tiny [0.0,0.0]} & 0.8\,{\tiny [0.6,0.9]} & 1.3\,{\tiny [1.0,1.5]} & 0.3\,{\tiny [0.0,0.9]} & 0.9\,{\tiny [0.7,1.2]} \\
\midrule
Across LLM agents & 5.7 & 6.6 & 6.1 & 5.5 & 7.0 & 6.1 & 5.6 & 6.9 & 6.2 \\
\midrule
MARL 100M & 11.6\,{\tiny [11.1,12.1]} & 20.0\,{\tiny [19.0,21.4]} & 15.1\,{\tiny [14.6,15.6]} & 12.1\,{\tiny [11.5,12.8]} & 17.0\,{\tiny [16.2,18.0]} & 14.2\,{\tiny [13.5,15.0]} & 12.2\,{\tiny [11.6,12.8]} & 5.6\,{\tiny [5.1,6.2]} & 9.4\,{\tiny [8.9,9.9]} \\
\rowcolor{rowalt}MARL 1B & 21.0\,{\tiny [19.4,22.6]} & 22.0\,{\tiny [20.8,23.1]} & 21.4\,{\tiny [20.9,21.9]} & 20.1\,{\tiny [19.6,20.7]} & 19.8\,{\tiny [19.4,20.2]} & 19.9\,{\tiny [19.5,20.5]} & 13.6\,{\tiny [12.8,14.2]} & 17.6\,{\tiny [16.9,18.2]} & 15.3\,{\tiny [14.9,15.7]} \\
Privileged heuristic$^{\dagger}$ & 27.1\,{\tiny [17.0,35.7]} & 36.5\,{\tiny [33.2,39.2]} & 31.1\,{\tiny [25.0,36.9]} & 25.7\,{\tiny [16.7,33.0]} & 31.8\,{\tiny [30.1,34.2]} & 28.3\,{\tiny [22.9,32.3]} & 7.3\,{\tiny [6.4,8.6]} & 17.0\,{\tiny [13.6,20.8]} & 11.4\,{\tiny [9.4,13.7]} \\
\rowcolor{rowalt}Human pilot$^{\ddagger}$ & 25.3 & 28.3 & 26.6 & -- & -- & -- & 24.4 & 18.9 & 22.1 \\
\bottomrule
\end{tabular}
\end{adjustbox}
\endgroup
\end{table}

\paragraph{Base competence does not guarantee coordination.} 
Decomposing performance reveals failures hidden by aggregate scores. GPT-5.4-High illustrates this separation: despite achieving the second-highest Base\% across difficulties ($14.1\%, 10.0\%, 11.8\%$), its Coord.\% ($7.0\%, 4.0\%, 4.2\%$) remains much lower, lagging behind smaller models such as Gemma-4-26B-A4B-it on Hard with non-overlapping $95\%$ CIs. Fig.~\ref{fig:main-diagnostic-coop} shows that GPT-5.4-High has the highest trade count ($48.9$) but relatively few revives ($3.4$), whereas Gemini-3.1-Pro-High is more balanced ($29.3$ trades, $6.9$ revives). Thus, local cooperative actions and base-task progress do not necessarily translate into sustained coordination. Conversely, Qwen3.5-122B-A10B and Gemma-4-26B-A4B-it achieve higher Coord.\% than Base\%, indicating that coordination can be exploited even when general competence is limited.

\paragraph{Most LLM agents do not exploit the difficulty axis.}
Trained MARL agents and the privileged heuristic both degrade as coordination difficulty increases (Sec.~\ref{subsec:difficulty}, Tbl.~\ref{tab:alem_results}). By contrast, aggregate LLM Total\% remains nearly flat at $\approx6\%$ across Easy, Medium, and Hard. This suggests that most LLM agents do not coordinate reliably enough for the difficulty parameter to strongly affect aggregate performance. Achievement-tier coverage supports this interpretation (Fig.~\ref{fig:achievement-tier-coverage}), as most LLMs fail to progress beyond Basic achievements regardless of difficulty, whereas stronger agents, such as MARL 1B and Gemini-3.1-Pro-High, show clearer degradation at intermediate and advanced tiers.

\paragraph{Coordination failures differ across task structure.}
The task-level breakdown in Fig.~\ref{fig:main-diagnostic-coord} shows that coordination is not a single axis. \textit{Handover} tasks are comparatively more accessible because they allow temporal slack, as one agent can initiate an opportunity and another can complete it within a bounded time window. By contrast, \textit{Synchronous-Hard} tasks require agents to identify a shared target, navigate to compatible positions, communicate intent, and act in the same timestep. This strict synchronisation compounds the spatial reasoning challenges already observed in single-agent LLMs~\citep{paglieri2025balrog}. The results reflect this difficulty, as Gemini-3.1-Pro-High leads LLMs on synchronous-hard coordination with only $18\%$ coverage, while the next-best tested LLMs, Gemma-4-31B-it and Qwen3.6-35B-A3B, reach $9\%$.

\paragraph{Pretrained priors offer advantages in sparse, multi-stage coordination.}
\textit{Construction} tasks require long-horizon planning in addition to coordinated execution. Agents must collect resources, progress through a technology chain, converge on a shared site, and execute a synchronised build. As shown in Fig.~\ref{fig:main-diagnostic-coord}, construction coverage is at most $2\%$ for every LLM except Gemini-3.1-Pro-High ($7\%$), and $0\%$ for the MARL 1B reference. This suggests that sparse, long-horizon coordination is difficult to discover through trial-and-error at this MARL compute budget, while pretrained priors may give stronger LLMs a limited advantage.
\paragraph{Scale alone does not explain coordination performance.}
Across open-weight models, performance is not monotonic in parameter count. Even within model families such as Gemma-4 and Qwen-3.5/3.6, coordination performance does not scale linearly across dense and mixture-of-experts variants (Figs.~\ref{fig:main-diagnostic-coord},~\ref{fig:main-diagnostic-survival}). This suggests that coordination depends on factors beyond raw scale, including possibly post-training, reasoning, and how effectively agents use communication.

\newlength{\panelheight}
\setlength{\panelheight}{0.18\textheight} %

\begin{figure}[t]
  \centering
  \captionsetup[subfigure]{font=small,labelfont=bf,skip=2pt}
  \includegraphics[width=0.36\textwidth]{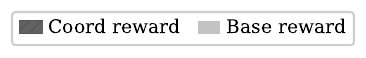}\\[0.25em]
  \begin{subfigure}[t]{0.32\textwidth}
    \centering
    \includegraphics[width=\linewidth]{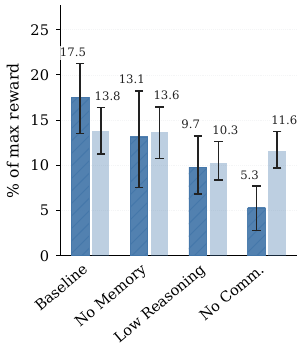}
    \caption{gemini-3.1-pro-high}
  \end{subfigure}
  \hspace{2em}
  \begin{subfigure}[t]{0.32\textwidth}
    \centering
    \includegraphics[width=\linewidth]{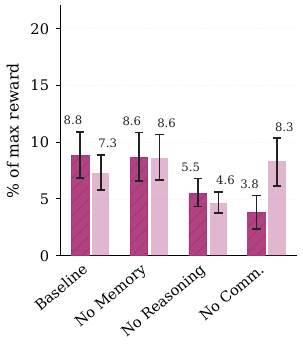}
    \caption{gemma-4-31b-it}
  \end{subfigure}
  \caption{\textbf{Harness ablations on Hard.} Communication, scratchpad memory, and reasoning ablations, reporting Coord.\% and Base\% with 95\% CIs. Communication is critical for coordination, while memory helps when agents use it for planning.}
  \label{fig:ablations}
\end{figure}

\subsubsection{\textbf{Q2:} How do communication, memory, and reasoning affect coordination?}\label{subsec:q2_ablate_mem_coms_reason}

We next ablate three components of the LLM agent harness: scratchpad memory, communication, and reasoning (Sec.~\ref{subsec:harness}). We run these ablations on Hard for the strongest closed-source and open-weight models from Sec.~\ref{subsec:q1}, Gemini-3.1-Pro-High and Gemma-4-31B-it. To understand how the agents use each component, we also analyse their messages, scratchpad entries, and reasoning traces. We use Gemini~3~Flash to annotate $7{,}167$ items from $2{,}396$ controlled agent steps sampled across both models and all three difficulties, with model identity and reward values masked (App.~\ref{app:functional-traces}).

\paragraph{Communication is critical for coordination.}
Removing communication produces the largest drop in coordination performance for both models (Fig.~\ref{fig:ablations}). Gemini-3.1-Pro-High falls from $17.5$ to $5.3$ Coord.\%, while Gemma-4-31B-it falls from $8.8$ to $3.8$ Coord.\%. The corresponding changes in Base\% are smaller and less consistent (Gemini $13.8 \to 11.6$, Gemma $7.3 \to 8.3$), suggesting that communication primarily supports coordination rather than general environment progress. The messages suggest how: agents use them to establish what to do together, who should act, and when (App.~\ref{app:comm_style}). Across all three difficulties, Gemini more often proposes joint targets ($39.1\%$ of messages versus $30.6\%$ for Gemma) and assigns actions to teammates ($41.1\%$ versus $31.1\%$). The largest gap is in coordinating timing ($35.7\%$ versus $21.7\%$; App.~\ref{app:functional-traces}). This is particularly important in \textsc{alem}'s synchronous and handover tasks, where teammates must act in the same step or complete successive actions within a short window.

\paragraph{Scratchpad memory helps when used for planning.}

The memory ablation is model-dependent. Removing scratchpad memory lowers Gemini-3.1-Pro-High's mean Coord.\% ($17.5 \to 13.1$), leaving Base\% nearly unchanged ($13.8 \to 13.6$). For Gemma-4-31B-it, Coord.\% changes little ($8.8 \to 8.6$), while Base\% rises ($7.3 \to 8.6$), with overlapping confidence intervals for each metric.

The scratchpads suggest a possible explanation for this difference. On Hard, Gemini uses its scratchpad as a \emph{forward-looking planner}. Most entries span several lines ($71.5\%$), and some contain turn-indexed sequences such as \texttt{T9:Do}, \texttt{T10:Move West}, and \texttt{T11:Give wood to Agent~1}. Gemma's scratchpads are mostly one-line state summaries (average $1.08$ lines), often listing coordinates and status information already available in the next observation (App.~\ref{app:memory_scratchpad}). Across all three difficulties, Gemini also records plans more often: $68.7\%$ of its scratchpad entries contain an ordered plan, compared with $32.5\%$ for Gemma, while $17.9\%$ contain a contingency plan, compared with $3.5\%$ (App.~\ref{app:functional-traces}). Keeping track of what to do next may help explain why Gemini benefits more from scratchpad memory.
\paragraph{Reasoning supports both coordination and base progress.}
Reasoning is the only ablation that lowers both Base\% and Coord.\% for both models (Fig.~\ref{fig:ablations}). For Gemini-3.1-Pro-High, we reduce reasoning to the lowest setting available through the API. Even so, Coord.\% falls from $17.5$ to $9.7$ and Base\% from $13.8$ to $10.3$. For Gemma-4-31B-it, disabling reasoning reduces Coord.\% from $8.8$ to $5.5$ and Base\% from $7.3$ to $4.6$.

With full reasoning, both models plan more often in their reasoning than anywhere else -- $83.6\%$ of Gemini's reasoning traces and $66.3\%$ of Gemma's contain a multi-step plan. Gemini, however, writes these plans down far more often. On steps where it plans in its reasoning, it also records a plan in its scratchpad or messages $88.8\%$ of the time, while Gemma does so $60.3\%$ of the time (App.~\ref{app:functional-traces}). 

Reducing reasoning leads to different failure modes in the two models (App.~\ref{app:reasoning}). For Gemini, it removes much of this planner-like behaviour: scratchpad length falls from $207$ to $69$ characters, turn-indexed plans nearly disappear ($12.4\% \to 0.1\%$), and messages become shorter ($132 \to 53$ characters). Gemma instead writes longer scratchpads without reasoning ($153 \to 228$ characters), and more of its entries contain future-oriented words such as \emph{will}, \emph{next}, or \emph{then} ($36.6\% \to 73.9\%$). This suggests that Gemma may use its scratchpad to compensate for the removed reasoning, although its scores still fall on both base tasks and coordination.

\begin{table}[tb]
  \centering
  \caption{\textbf{Heterogeneous teams on Hard.} $\Delta$ is the mixed-team score minus the mean of the same three models in homogeneous teams (percentage points; 10 seeds per team, 20 for the all-Gemma team). Coordination falls for every team, while base reward mostly rises. The last two rows use the same models with roles reassigned.}
  \label{tab:heterogeneous-main}
  \vspace{3.5pt}
  \footnotesize
  \renewcommand{\arraystretch}{0.9}
  \setlength{\tabcolsep}{3pt}
  \begin{tabular}{@{}lllrrr@{}}
    \toprule
    Warrior & Forager & Miner & $\Delta$Base & $\Delta$Coord. & $\Delta$Total \\
    \midrule
    gemma-4-e4b-it      & gemma-4-26b-a4b-it & gemma-4-31b-it      & $+1.1$ & $-1.3$ & $+0.1$ \\
    gemini-3.1-pro-high & gpt-5.4-high       & gemma-4-31b-it      & $+2.0$ & $-2.9$ & $-0.1$ \\
    gemini-3.1-pro-high & gpt-5.4-high       & qwen3.6-27b         & $+0.5$ & $-2.2$ & $-0.7$ \\
    gpt-5.4-high        & gemma-4-31b-it     & qwen3.6-27b         & $+1.8$ & $-3.0$ & $-0.3$ \\
    \addlinespace[3pt]
    \multicolumn{6}{@{}l}{\textit{Same three models, different roles}} \\
    \cmidrule(r){1-3}
    gemini-3.1-pro-high & gemma-4-26b-a4b-it & gemma-4-31b-it      & $-0.1$ & $-6.9$ & $-3.0$ \\
    gemma-4-26b-a4b-it  & gemma-4-31b-it     & gemini-3.1-pro-high & $+1.0$ & $-3.3$ & $-0.8$ \\
    \midrule
    Mean & & & $+1.1$ & $-3.3$ & $-0.8$ \\
    \bottomrule
  \end{tabular}
\end{table}

\subsubsection{\textbf{Q3:} How well do agents coordinate in teams composed of different models?}

We evaluate six heterogeneous teams on Hard. One uses three Gemma models of different sizes, and the other five combine models from different families. For each mixed team, we take as our baseline the mean score of its three models when each plays with two other copies of itself (Table~\ref{tab:heterogeneous-main}).

\paragraph{Mixed teams score lower on coordination but typically higher on base tasks.} All six teams score below their homogeneous baseline on Coord.\%, by 3.3 percentage points (pp) on average. Five of the six score higher on Base\%, with a mean gain of 1.1~pp across all six. These gains partly offset the coordination losses. Five teams finish within 0.8~pp of their baseline on Total\%.

Role assignment also makes a difference for the team with Gemini and two Gemma models. With Gemini as warrior, the team scores 5.0 on Coord.\%. Rotating all three roles so that Gemini plays the miner brings the score to 8.6, still below the homogeneous baseline of 11.9.

The results could indicate that mixed teams may face additional coordination problems, as agents may have to align their communication conventions, planning horizons, or actions.

\paragraph{Summary.}
Overall, \textsc{alem} reveals coordination as a bottleneck distinct from base-task performance. Communication matters most among the ablated harness components. Initial heterogeneous-team experiments suggest that a strong model does not improve its teammates' coordination: mixed teams score lower on coordination reward but higher on base reward, and role assignment matters.

\section{Related Work}
\paragraph{Long-horizon and agentic evaluation.} LLMs are increasingly evaluated as agents in long-horizon, open-ended tasks~\citep{jansen2024discoveryworld,kwa2025measuring,yamada2025ai,chan2025mle}. Recent benchmarks test models in complex environments requiring spatial reasoning, exploration, planning, and tool use~\citep{paglieri2025balrog,zhang2026deepplanning,zhao2026ama,he2026texttt}. Procedural generation is especially useful because it reduces memorisation, mitigates contamination, and supports scalable variation across tasks~\citep{paglieri2025balrog,shojaee2025illusion}. \textsc{alem} brings these principles to multi-agent evaluation, where agents must also infer and satisfy diverse coordination demands under partial observability.

\paragraph{Multi-Agent LLM collaboration.} Multi-agent LLMs have shown promise in code generation, debate, and collaborative reasoning~\citep{hong2023metagpt,du2023improving}. Recent benchmarks evaluate LLM collaboration in interactive tasks~\citep{zhu2025multiagentbench,sun2025collab,grotschla2025agentsnet,davidson2025collaboration} and embodied household and Minecraft environments~\citep{chang2025partnr,long2024teamcraft,white2025collaborating}. COOP2 studies cooperation through tasks requiring simultaneous resource collection or block pushing~\citep{yang2026emcoop}. \textsc{alem} varies both when agents must act and how many must participate, sampling coordination requirements across tasks and episodes. This tests whether agents can adapt to different forms of coordination as they progress through an open-ended, long-horizon world.

\paragraph{MARL and coordination.} MARL benchmarks study coordination through joint combat, task handovers, and resource interdependence~\citep{samvelyan2019starcraft,carroll2019utility,agapiou2022melting}, including in open-world settings~\citep{suarez2023neural,ye2025efficient}. \textsc{alem} builds on Craftax-Coop~\citep{omari2025multi}, but makes the coordination structure itself procedural: each episode samples when agents must act and how many must participate at individual targets, spanning a coordination spectrum (Fig.~\ref{fig:coordination_spectrum}). It combines these requirements with soft rather than hard specialisation, explicit communication, and controllable coordination difficulty. We further use information-theoretic diagnostics~\citep{tessera2026probing} to provide evidence that these mechanics induce multi-agent interdependence (App.~\ref{app:env_probing}).

\section{Conclusions}

We introduce \textsc{alem}, a JAX-based benchmark for long-horizon, open-ended multi-agent coordination. Our zero-shot evaluations of 13 LLMs alongside trained MARL baselines show that coordination is a distinct bottleneck for current agents, base-task competence does not guarantee coordinated progress, communication has the largest effect on coordinated behaviour, and mixed-model teams coordinate worse than their homogeneous counterparts. By exposing failure modes across communication, memory, reasoning, and team composition, \textsc{alem} makes coordination measurable and provides a controlled testbed for developing agents that coordinate effectively.

\paragraph{Limitations and Future Work.} \label{par:limitations} 

Our results come from a single environment, so findings may not transfer to settings with different dynamics or forms of interdependence. We also evaluate only text-based LLM agents, leaving Vision-Language Models (VLMs) and Claude-family models for future work, and API costs limit proprietary models (GPT and Gemini) to 10 seeds per difficulty. Our heterogeneous-team results cover only six teams, so we treat them as initial evidence. Beyond these, \textsc{alem} can support work on stronger agent harnesses, cross-episode memory, and lifelong learning in interactive multi-agent worlds.

\begin{ack}
We thank the Gemini Academic Program, the Edinburgh ELM Platform (through EDINA), and the Edinburgh International Data Facility (EIDF) for their support. We also thank Bassel Al Omari, Samuel Garcin, Raul Steleac, Benjamin Sanati, and Leonard Hinckeldey for insightful discussions and feedback on this work. Aidan Scannell was supported by the UK Engineering and Physical Sciences Research Council (EPSRC, grant EP/W002876/1). Andras Szecsenyi and Cameron Barker were funded by EPSRC through PhD studentships within the CDT in Machine Learning Systems, hosted by the School of Informatics, University of Edinburgh (EP/Y03516X/1). Henry Gouk was supported by the Royal Academy of Engineering under the Research Fellowship programme.
\end{ack}
\newpage

\bibliographystyle{plainnat}
\bibliography{ref}

\begin{thebibliography}{77}
\providecommand{\natexlab}[1]{#1}
\providecommand{\url}[1]{\texttt{#1}}
\expandafter\ifx\csname urlstyle\endcsname\relax
  \providecommand{\doi}[1]{doi: #1}\else
  \providecommand{\doi}{doi: \begingroup \urlstyle{rm}\Url}\fi

\bibitem[Agapiou et~al.(2022)Agapiou, Vezhnevets, Du{\'e}{\~n}ez-Guzm{\'a}n,
  Matyas, Mao, Sunehag, K{\"o}ster, Madhushani, Kopparapu, Comanescu,
  et~al.]{agapiou2022melting}
John~P Agapiou, Alexander~Sasha Vezhnevets, Edgar~A Du{\'e}{\~n}ez-Guzm{\'a}n,
  Jayd Matyas, Yiran Mao, Peter Sunehag, Raphael K{\"o}ster, Udari Madhushani,
  Kavya Kopparapu, Ramona Comanescu, et~al.
\newblock Melting pot 2.0.
\newblock \emph{arXiv preprint arXiv:2211.13746}, 2022.

\bibitem[Agarwal et~al.(2021)Agarwal, Schwarzer, Castro, Courville, and
  Bellemare]{agarwal2021deep}
Rishabh Agarwal, Max Schwarzer, Pablo~Samuel Castro, Aaron Courville, and
  Marc~G Bellemare.
\newblock Deep reinforcement learning at the edge of the statistical precipice.
\newblock \emph{Advances in Neural Information Processing Systems}, 2021.

\bibitem[Agashe et~al.(2025)Agashe, Fan, Reyna, and Wang]{agashe2025llm}
Saaket Agashe, Yue Fan, Anthony Reyna, and Xin~Eric Wang.
\newblock Llm-coordination: evaluating and analyzing multi-agent coordination
  abilities in large language models.
\newblock In \emph{Findings of the Association for Computational Linguistics:
  NAACL 2025}, pages 8038--8057, 2025.

\bibitem[Bard et~al.(2020)Bard, Foerster, Chandar, Burch, Lanctot, Song,
  Parisotto, Dumoulin, Moitra, Hughes, et~al.]{bard2020hanabi}
Nolan Bard, Jakob~N Foerster, Sarath Chandar, Neil Burch, Marc Lanctot,
  H~Francis Song, Emilio Parisotto, Vincent Dumoulin, Subhodeep Moitra, Edward
  Hughes, et~al.
\newblock The hanabi challenge: A new frontier for ai research.
\newblock \emph{Artificial Intelligence}, 280:\penalty0 103216, 2020.

\bibitem[Bellemare et~al.(2013)Bellemare, Naddaf, Veness, and
  Bowling]{bellemare2013arcade}
Marc~G Bellemare, Yavar Naddaf, Joel Veness, and Michael Bowling.
\newblock The arcade learning environment: An evaluation platform for general
  agents.
\newblock \emph{Journal of artificial intelligence research}, 47:\penalty0
  253--279, 2013.

\bibitem[Bradbury et~al.(2018)Bradbury, Frostig, Hawkins, Johnson, Katariya,
  Leary, Maclaurin, Necula, Paszke, Vander{P}las, Wanderman-{M}ilne, and
  Zhang]{jax2018github}
James Bradbury, Roy Frostig, Peter Hawkins, Matthew~James Johnson, Yash
  Katariya, Chris Leary, Dougal Maclaurin, George Necula, Adam Paszke, Jake
  Vander{P}las, Skye Wanderman-{M}ilne, and Qiao Zhang.
\newblock {JAX}: composable transformations of {P}ython+{N}um{P}y programs,
  2018.
\newblock URL \url{http://github.com/jax-ml/jax}.

\bibitem[Brown and Sandholm(2018)]{brown2018superhuman}
Noam Brown and Tuomas Sandholm.
\newblock Superhuman ai for heads-up no-limit poker: Libratus beats top
  professionals.
\newblock \emph{Science}, 359\penalty0 (6374):\penalty0 418--424, 2018.

\bibitem[Brown et~al.(2020)Brown, Mann, Ryder, Subbiah, Kaplan, Dhariwal,
  Neelakantan, Shyam, Sastry, Askell, et~al.]{brown2020language}
Tom Brown, Benjamin Mann, Nick Ryder, Melanie Subbiah, Jared~D Kaplan, Prafulla
  Dhariwal, Arvind Neelakantan, Pranav Shyam, Girish Sastry, Amanda Askell,
  et~al.
\newblock Language models are few-shot learners.
\newblock \emph{Advances in neural information processing systems},
  33:\penalty0 1877--1901, 2020.

\bibitem[Campbell et~al.(2002)Campbell, Hoane~Jr, and Hsu]{campbell2002deep}
Murray Campbell, A~Joseph Hoane~Jr, and Feng-hsiung Hsu.
\newblock Deep blue.
\newblock \emph{Artificial intelligence}, 134\penalty0 (1-2):\penalty0 57--83,
  2002.

\bibitem[Carroll et~al.(2019)Carroll, Shah, Ho, Griffiths, Seshia, Abbeel, and
  Dragan]{carroll2019utility}
Micah Carroll, Rohin Shah, Mark~K Ho, Tom Griffiths, Sanjit Seshia, Pieter
  Abbeel, and Anca Dragan.
\newblock On the utility of learning about humans for human-ai coordination.
\newblock \emph{Advances in neural information processing systems}, 32, 2019.

\bibitem[Chan et~al.(2025)Chan, Chowdhury, Jaffe, Aung, Sherburn, Mays,
  Starace, Liu, Maksin, Patwardhan, Madry, and Weng]{chan2025mle}
Jun~Shern Chan, Neil Chowdhury, Oliver Jaffe, James Aung, Dane Sherburn, Evan
  Mays, Giulio Starace, Kevin Liu, Leon Maksin, Tejal Patwardhan, Aleksander
  Madry, and Lilian Weng.
\newblock {MLE}-bench: Evaluating machine learning agents on machine learning
  engineering.
\newblock In \emph{The Thirteenth International Conference on Learning
  Representations}, 2025.
\newblock URL \url{https://openreview.net/forum?id=6s5uXNWGIh}.

\bibitem[Chang et~al.(2025)Chang, Chhablani, Clegg, Dallaire~Cote, Desai,
  Hlavac, Karashchuk, Krantz, Mottaghi, Parashar, et~al.]{chang2025partnr}
Matthew Chang, Gunjan Chhablani, Alexander Clegg, Mikael Dallaire~Cote, Ruta
  Desai, Michal Hlavac, Vladimir Karashchuk, Jacob Krantz, Roozbeh Mottaghi,
  Priyam Parashar, et~al.
\newblock Partnr: A benchmark for planning and reasoning in embodied
  multi-agent tasks.
\newblock In \emph{International Conference on Learning Representations},
  volume 2025, pages 65205--65268, 2025.

\bibitem[Davidson et~al.(2025)Davidson, Fourney, Amershi, West, Horvitz, and
  Kamar]{davidson2025collaboration}
Tim~R Davidson, Adam Fourney, Saleema Amershi, Robert West, Eric Horvitz, and
  Ece Kamar.
\newblock The collaboration gap: Exploration and benchmarking of open-world
  agentic cooperation.
\newblock \emph{arXiv preprint arXiv:2511.02687}, 2025.

\bibitem[De~Witt et~al.(2020)De~Witt, Gupta, Makoviichuk, Makoviychuk, Torr,
  Sun, and Whiteson]{de2020independent}
Christian~Schroeder De~Witt, Tarun Gupta, Denys Makoviichuk, Viktor
  Makoviychuk, Philip~HS Torr, Mingfei Sun, and Shimon Whiteson.
\newblock Is independent learning all you need in the starcraft multi-agent
  challenge?
\newblock \emph{arXiv preprint arXiv:2011.09533}, 2020.

\bibitem[Devlin et~al.(2019)Devlin, Chang, Lee, and Toutanova]{devlin2019bert}
Jacob Devlin, Ming-Wei Chang, Kenton Lee, and Kristina Toutanova.
\newblock Bert: Pre-training of deep bidirectional transformers for language
  understanding.
\newblock In \emph{Proceedings of the 2019 conference of the North American
  chapter of the association for computational linguistics: human language
  technologies, volume 1 (long and short papers)}, pages 4171--4186, 2019.

\bibitem[Du et~al.(2023)Du, Li, Torralba, Tenenbaum, and
  Mordatch]{du2023improving}
Yilun Du, Shuang Li, Antonio Torralba, Joshua~B Tenenbaum, and Igor Mordatch.
\newblock Improving factuality and reasoning in language models through
  multiagent debate.
\newblock In \emph{Forty-first International Conference on Machine Learning},
  2023.

\bibitem[Ellis et~al.(2023)Ellis, Cook, Moalla, Samvelyan, Sun, Mahajan,
  Foerster, and Whiteson]{ellis2023smacv2}
Benjamin Ellis, Jonathan Cook, Skander Moalla, Mikayel Samvelyan, Mingfei Sun,
  Anuj Mahajan, Jakob Foerster, and Shimon Whiteson.
\newblock Smacv2: An improved benchmark for cooperative multi-agent
  reinforcement learning.
\newblock \emph{Advances in Neural Information Processing Systems},
  36:\penalty0 37567--37593, 2023.

\bibitem[Gallici et~al.(2025)Gallici, Fellows, Ellis, Pou, Masmitja, Foerster,
  and Martin]{gallicisimplifying}
Matteo Gallici, Mattie Fellows, Benjamin Ellis, Bartomeu Pou, Ivan Masmitja,
  Jakob~Nicolaus Foerster, and Mario Martin.
\newblock Simplifying deep temporal difference learning.
\newblock In \emph{The Thirteenth International Conference on Learning
  Representations}, 2025.

\bibitem[Gessler et~al.(2025)Gessler, Dizdarevic, Calinescu, Ellis, Lupu, and
  Foerster]{gessler2025overcookedv}
Tobias Gessler, Tin Dizdarevic, Ani Calinescu, Benjamin Ellis, Andrei Lupu, and
  Jakob~Nicolaus Foerster.
\newblock Overcookedv2: Rethinking overcooked for zero-shot coordination.
\newblock In \emph{The Thirteenth International Conference on Learning
  Representations}, 2025.
\newblock URL \url{https://openreview.net/forum?id=hlvLM3GX8R}.

\bibitem[{Google DeepMind}(2026{\natexlab{a}})]{gemini31pro_2026}
{Google DeepMind}.
\newblock Gemini 3.1 pro model card.
\newblock \url{https://deepmind.google/models/model-cards/gemini-3-1-pro/},
  February 2026{\natexlab{a}}.

\bibitem[{Google DeepMind}(2026{\natexlab{b}})]{gemma4_2026}
{Google DeepMind}.
\newblock Gemma 4 model card.
\newblock \url{https://ai.google.dev/gemma/docs/core/model_card_4}, April
  2026{\natexlab{b}}.
\newblock Accessed: 2026-04-25.

\bibitem[Grady et~al.(2026)Grady, Parker, Zarov, Course, Taylor, and
  Taylor]{grady2026kellybench}
Thomas Grady, Kip Parker, Iliyan Zarov, Henry Course, Chengxi Taylor, and Ross
  Taylor.
\newblock Kellybench: A benchmark for long-horizon sequential decision making.
\newblock \emph{arXiv preprint arXiv:2604.27865}, 2026.

\bibitem[Grattafiori et~al.(2024)Grattafiori, Dubey, Jauhri, Pandey, Kadian,
  Al-Dahle, Letman, Mathur, Schelten, Vaughan, et~al.]{grattafiori2024llama}
Aaron Grattafiori, Abhimanyu Dubey, Abhinav Jauhri, Abhinav Pandey, Abhishek
  Kadian, Ahmad Al-Dahle, Aiesha Letman, Akhil Mathur, Alan Schelten, Alex
  Vaughan, et~al.
\newblock The llama 3 herd of models.
\newblock \emph{arXiv preprint arXiv:2407.21783}, 2024.

\bibitem[Gr{\"o}tschla et~al.(2025)Gr{\"o}tschla, M{\"u}ller, T{\"o}nshoff,
  Galkin, and Perozzi]{grotschla2025agentsnet}
Florian Gr{\"o}tschla, Luis M{\"u}ller, Jan T{\"o}nshoff, Mikhail Galkin, and
  Bryan Perozzi.
\newblock Agentsnet: Coordination and collaborative reasoning in multi-agent
  llms.
\newblock \emph{arXiv preprint arXiv:2507.08616}, 2025.

\bibitem[Guo et~al.(2024)Guo, Chen, Wang, Chang, Pei, Chawla, Wiest, and
  Zhang]{guo2024large}
Taicheng Guo, Xiuying Chen, Yaqi Wang, Ruidi Chang, Shichao Pei, Nitesh~V
  Chawla, Olaf Wiest, and Xiangliang Zhang.
\newblock Large language model based multi-agents: a survey of progress and
  challenges.
\newblock In \emph{Proceedings of the Thirty-Third International Joint
  Conference on Artificial Intelligence}, pages 8048--8057, 2024.

\bibitem[Hafner(2022)]{hafner2022benchmarking}
Danijar Hafner.
\newblock Benchmarking the spectrum of agent capabilities.
\newblock In \emph{International Conference on Learning Representations}, 2022.
\newblock URL \url{https://openreview.net/forum?id=1W0z96MFEoH}.

\bibitem[Hammond et~al.(2025)Hammond, Chan, Clifton, Hoelscher-Obermaier, Khan,
  McLean, Smith, Barfuss, Foerster, Gaven{\v{c}}iak, et~al.]{hammond2025multi}
Lewis Hammond, Alan Chan, Jesse Clifton, Jason Hoelscher-Obermaier, Akbir Khan,
  Euan McLean, Chandler Smith, Wolfram Barfuss, Jakob Foerster, Tom{\'a}{\v{s}}
  Gaven{\v{c}}iak, et~al.
\newblock Multi-agent risks from advanced ai.
\newblock \emph{arXiv preprint arXiv:2502.14143}, 2025.

\bibitem[Hansen et~al.(2004)Hansen, Bernstein, and
  Zilberstein]{hansen2004dynamic}
Eric~A Hansen, Daniel~S Bernstein, and Shlomo Zilberstein.
\newblock Dynamic programming for partially observable stochastic games.
\newblock In \emph{AAAI}, volume~4, pages 709--715, 2004.

\bibitem[He et~al.(2026)He, Jain, Kumar, Tu, Bakshi, Patro, and
  Rajani]{he2026texttt}
Muyu He, Adit Jain, Anand Kumar, Vincent Tu, Soumyadeep Bakshi, Sachin Patro,
  and Nazneen Rajani.
\newblock $yc-bench$: Benchmarking ai agents for long-term planning and
  consistent execution.
\newblock \emph{arXiv preprint arXiv:2604.01212}, 2026.

\bibitem[Hong et~al.(2023)Hong, Zhuge, Chen, Zheng, Cheng, Wang, Zhang, Wang,
  Yau, Lin, et~al.]{hong2023metagpt}
Sirui Hong, Mingchen Zhuge, Jonathan Chen, Xiawu Zheng, Yuheng Cheng, Jinlin
  Wang, Ceyao Zhang, Zili Wang, Steven Ka~Shing Yau, Zijuan Lin, et~al.
\newblock Metagpt: Meta programming for a multi-agent collaborative framework.
\newblock In \emph{The twelfth international conference on learning
  representations}, 2023.

\bibitem[Hughes et~al.(2024)Hughes, Dennis, Parker-Holder, Behbahani,
  Mavalankar, Shi, Schaul, and Rockt\"{a}schel]{pmlr-v235-hughes24a}
Edward Hughes, Michael~D Dennis, Jack Parker-Holder, Feryal Behbahani, Aditi
  Mavalankar, Yuge Shi, Tom Schaul, and Tim Rockt\"{a}schel.
\newblock Position: Open-endedness is essential for artificial superhuman
  intelligence.
\newblock In \emph{Proceedings of the 41st International Conference on Machine
  Learning}, volume 235 of \emph{Proceedings of Machine Learning Research},
  pages 20597--20616. PMLR, 21--27 Jul 2024.
\newblock URL \url{https://proceedings.mlr.press/v235/hughes24a.html}.

\bibitem[Jansen et~al.(2024)Jansen, C{\^o}t{\'e}, Khot, Bransom, Dalvi~Mishra,
  Majumder, Tafjord, and Clark]{jansen2024discoveryworld}
Peter Jansen, Marc-Alexandre C{\^o}t{\'e}, Tushar Khot, Erin Bransom, Bhavana
  Dalvi~Mishra, Bodhisattwa~Prasad Majumder, Oyvind Tafjord, and Peter Clark.
\newblock Discoveryworld: A virtual environment for developing and evaluating
  automated scientific discovery agents.
\newblock \emph{Advances in neural information processing systems},
  37:\penalty0 10088--10116, 2024.

\bibitem[Kazemkhani et~al.(2025)Kazemkhani, Pandya, Cornelisse, Shacklett, and
  Vinitsky]{kazemkhani2025gpudrive}
Saman Kazemkhani, Aarav Pandya, Daphne Cornelisse, Brennan Shacklett, and
  Eugene Vinitsky.
\newblock Gpudrive: Data-driven, multi-agent driving simulation at 1 million
  fps.
\newblock In \emph{Proceedings of the International Conference on Learning
  Representations (ICLR)}, 2025.
\newblock URL \url{https://arxiv.org/abs/2408.01584}.

\bibitem[Kwa et~al.(2025)Kwa, West, Becker, Deng, Garcia, Hasin, Jawhar,
  Kinniment, Rush, Von~Arx, et~al.]{kwa2025measuring}
Thomas Kwa, Ben West, Joel Becker, Amy Deng, Katharyn Garcia, Max Hasin, Sami
  Jawhar, Megan Kinniment, Nate Rush, Sydney Von~Arx, et~al.
\newblock Measuring ai ability to complete long tasks.
\newblock \emph{arXiv preprint arXiv:2503.14499}, 352, 2025.

\bibitem[Kwon et~al.(2023)Kwon, Li, Zhuang, Sheng, Zheng, Yu, Gonzalez, Zhang,
  and Stoica]{kwon2023efficient}
Woosuk Kwon, Zhuohan Li, Siyuan Zhuang, Ying Sheng, Lianmin Zheng, Cody~Hao Yu,
  Joseph Gonzalez, Hao Zhang, and Ion Stoica.
\newblock Efficient memory management for large language model serving with
  pagedattention.
\newblock In \emph{Proceedings of the 29th symposium on operating systems
  principles}, pages 611--626, 2023.

\bibitem[Leibo et~al.(2021)Leibo, Due{\~n}ez-Guzman, Vezhnevets, Agapiou,
  Sunehag, Koster, Matyas, Beattie, Mordatch, and Graepel]{leibo2021scalable}
Joel~Z Leibo, Edgar~A Due{\~n}ez-Guzman, Alexander Vezhnevets, John~P Agapiou,
  Peter Sunehag, Raphael Koster, Jayd Matyas, Charlie Beattie, Igor Mordatch,
  and Thore Graepel.
\newblock Scalable evaluation of multi-agent reinforcement learning with
  melting pot.
\newblock In \emph{International conference on machine learning}, pages
  6187--6199. PMLR, 2021.

\bibitem[Liu et~al.(2023)Liu, Shah, Boussif, Meo, Goyal, Shu, Mozer, Heess, and
  Bengio]{liu2023stateful}
Dianbo Liu, Vedant Shah, Oussama Boussif, Cristian Meo, Anirudh Goyal, Tianmin
  Shu, Michael~Curtis Mozer, Nicolas Heess, and Yoshua Bengio.
\newblock Stateful active facilitator: Coordination and environmental
  heterogeneity in cooperative multi-agent reinforcement learning.
\newblock In \emph{The Eleventh International Conference on Learning
  Representations}, 2023.
\newblock URL \url{https://openreview.net/forum?id=B4maZQLLW0_}.

\bibitem[Long et~al.(2024)Long, Li, Gong, Wu, Terzopoulos, and
  Gao]{long2024teamcraft}
Qian Long, Zhi Li, Ran Gong, Ying~Nian Wu, Demetri Terzopoulos, and Xiaofeng
  Gao.
\newblock Teamcraft: A benchmark for multi-modal multi-agent systems in
  minecraft.
\newblock \emph{arXiv preprint arXiv:2412.05255}, 2024.

\bibitem[Lu et~al.(2026)Lu, Lu, Lange, Yamada, Hu, Foerster, Ha, and
  Clune]{lu2026towards}
Chris Lu, Cong Lu, Robert~Tjarko Lange, Yutaro Yamada, Shengran Hu, Jakob
  Foerster, David Ha, and Jeff Clune.
\newblock Towards end-to-end automation of ai research.
\newblock \emph{Nature}, 651\penalty0 (8107):\penalty0 914--919, 2026.

\bibitem[Malone and Crowston(1994)]{malone1994interdisciplinary}
Thomas~W Malone and Kevin Crowston.
\newblock The interdisciplinary study of coordination.
\newblock \emph{ACM Computing Surveys (CSUR)}, 26\penalty0 (1):\penalty0
  87--119, 1994.

\bibitem[Matthews et~al.(2024)Matthews, Beukman, Ellis, Samvelyan, Jackson,
  Coward, and Foerster]{matthews2024craftax}
Michael Matthews, Michael Beukman, Benjamin Ellis, Mikayel Samvelyan, Matthew
  Jackson, Samuel Coward, and Jakob Foerster.
\newblock Craftax: A lightning-fast benchmark for open-ended reinforcement
  learning.
\newblock In \emph{International Conference on Machine Learning ({ICML})},
  2024.

\bibitem[Meireles et~al.(2025)Meireles, Lauffer, Bhati, and
  Allen]{meireles2025the}
Mariana Meireles, Niklas Lauffer, Rupali Bhati, and Cameron Allen.
\newblock The influence of scaffolds on coordination scaling laws in {LLM}
  agents.
\newblock In \emph{Workshop on Scaling Environments for Agents}, 2025.
\newblock URL \url{https://openreview.net/forum?id=E9whrbtgUA}.

\bibitem[Nakano et~al.(2021)Nakano, Hilton, Balaji, Wu, Ouyang, Kim, Hesse,
  Jain, Kosaraju, Saunders, et~al.]{nakano2021webgpt}
Reiichiro Nakano, Jacob Hilton, Suchir Balaji, Jeff Wu, Long Ouyang, Christina
  Kim, Christopher Hesse, Shantanu Jain, Vineet Kosaraju, William Saunders,
  et~al.
\newblock Webgpt: Browser-assisted question-answering with human feedback.
\newblock \emph{arXiv preprint arXiv:2112.09332}, 2021.

\bibitem[Omari et~al.(2025)Omari, Matthews, Rutherford, and
  Foerster]{omari2025multi}
Bassel~Al Omari, Michael Matthews, Alexander Rutherford, and Jakob~Nicolaus
  Foerster.
\newblock Multi-agent craftax: Benchmarking open-ended multi-agent
  reinforcement learning at the hyperscale.
\newblock \emph{arXiv preprint arXiv:2511.04904}, 2025.

\bibitem[{OpenAI}(2026)]{gpt54_2026}
{OpenAI}.
\newblock Introducing gpt-5.4.
\newblock \url{https://openai.com/index/introducing-gpt-5-4/}, March 2026.

\bibitem[Paglieri et~al.(2025)Paglieri, Cupia{\l}, Coward, Piterbarg, Wolczyk,
  Khan, Pignatelli, Kuci{\'n}ski, Pinto, Fergus, Foerster, Parker-Holder, and
  Rockt{\"a}schel]{paglieri2025balrog}
Davide Paglieri, Bart{\l}omiej Cupia{\l}, Samuel Coward, Ulyana Piterbarg,
  Maciej Wolczyk, Akbir Khan, Eduardo Pignatelli, {\L}ukasz Kuci{\'n}ski,
  Lerrel Pinto, Rob Fergus, Jakob~Nicolaus Foerster, Jack Parker-Holder, and
  Tim Rockt{\"a}schel.
\newblock {BALROG}: Benchmarking agentic {LLM} and {VLM} reasoning on games.
\newblock In \emph{The Thirteenth International Conference on Learning
  Representations}, 2025.
\newblock URL \url{https://openreview.net/forum?id=fp6t3F669F}.

\bibitem[{Qwen Team}(2026{\natexlab{a}})]{qwen35blog}
{Qwen Team}.
\newblock Qwen3.5: Accelerating productivity with native multimodal agents,
  February 2026{\natexlab{a}}.
\newblock URL \url{https://qwen.ai/blog?id=qwen3.5}.

\bibitem[{Qwen Team}(2026{\natexlab{b}})]{qwen36plus}
{Qwen Team}.
\newblock {Qwen3.6-Plus}: Towards real world agents, April 2026{\natexlab{b}}.
\newblock URL \url{https://qwen.ai/blog?id=qwen3.6}.

\bibitem[Rutherford et~al.(2024)Rutherford, Ellis, Gallici, Cook, Lupu,
  Ingvarsson, Willi, Hammond, Khan, de~Witt, et~al.]{rutherford2024jaxmarl}
Alexander Rutherford, Benjamin Ellis, Matteo Gallici, Jonathan Cook, Andrei
  Lupu, Gar{\dh}ar Ingvarsson, Timon Willi, Ravi Hammond, Akbir Khan,
  Christian~S de~Witt, et~al.
\newblock Jaxmarl: Multi-agent rl environments and algorithms in jax.
\newblock \emph{Advances in Neural Information Processing Systems},
  37:\penalty0 50925--50951, 2024.

\bibitem[Samvelyan et~al.(2019)Samvelyan, Rashid, Schroeder~de Witt, Farquhar,
  Nardelli, Rudner, Hung, Torr, Foerster, and Whiteson]{samvelyan2019starcraft}
Mikayel Samvelyan, Tabish Rashid, Christian Schroeder~de Witt, Gregory
  Farquhar, Nantas Nardelli, Tim G.~J. Rudner, Chia-Man Hung, Philip H.~S.
  Torr, Jakob Foerster, and Shimon Whiteson.
\newblock The starcraft multi-agent challenge.
\newblock In \emph{Proceedings of the 18th International Conference on
  Autonomous Agents and MultiAgent Systems}, AAMAS '19, page 2186–2188,
  Richland, SC, 2019. International Foundation for Autonomous Agents and
  Multiagent Systems.

\bibitem[Schelling(1980)]{schelling1980strategy}
Thomas~C Schelling.
\newblock \emph{The Strategy of Conflict: with a new Preface by the Author}.
\newblock Harvard university press, 1980.

\bibitem[Schrittwieser et~al.(2020)Schrittwieser, Antonoglou, Hubert, Simonyan,
  Sifre, Schmitt, Guez, Lockhart, Hassabis, Graepel,
  et~al.]{schrittwieser2020mastering}
Julian Schrittwieser, Ioannis Antonoglou, Thomas Hubert, Karen Simonyan,
  Laurent Sifre, Simon Schmitt, Arthur Guez, Edward Lockhart, Demis Hassabis,
  Thore Graepel, et~al.
\newblock Mastering atari, go, chess and shogi by planning with a learned
  model.
\newblock \emph{Nature}, 588\penalty0 (7839):\penalty0 604--609, 2020.

\bibitem[Shojaee et~al.(2025)Shojaee, Mirzadeh, Alizadeh, Horton, Bengio, and
  Farajtabar]{shojaee2025illusion}
Parshin Shojaee, Seyed~Iman Mirzadeh, Keivan Alizadeh, Maxwell Horton, Samy
  Bengio, and Mehrdad Farajtabar.
\newblock The illusion of thinking: Understanding the strengths and limitations
  of reasoning models via the lens of problem complexity.
\newblock In \emph{The Thirty-ninth Annual Conference on Neural Information
  Processing Systems}, 2025.
\newblock URL \url{https://openreview.net/forum?id=YghiOusmvw}.

\bibitem[Sinha et~al.(2026)Sinha, Arun, Goel, Staab, and Geiping]{sinha2026the}
Akshit Sinha, Arvindh Arun, Shashwat Goel, Steffen Staab, and Jonas Geiping.
\newblock The illusion of diminishing returns: Measuring long horizon execution
  in {LLM}s.
\newblock In \emph{The Fourteenth International Conference on Learning
  Representations}, 2026.
\newblock URL \url{https://openreview.net/forum?id=3lm8lWYxiq}.

\bibitem[Skyrms(2004)]{skyrms2004stag}
Brian Skyrms.
\newblock \emph{The stag hunt and the evolution of social structure}.
\newblock Cambridge University Press, 2004.

\bibitem[Smith et~al.(2025)Smith, Abdulhai, Diaz, Tesic, Trivedi, Vezhnevets,
  Hammond, Clifton, Chang, Du{\'e}{\~n}ez-Guzm{\'a}n,
  et~al.]{smith2025evaluating}
Chandler Smith, Marwa Abdulhai, Manfred Diaz, Marko Tesic, Rakshit Trivedi,
  Sasha Vezhnevets, Lewis Hammond, Jesse Clifton, Minsuk Chang, Edgar~A
  Du{\'e}{\~n}ez-Guzm{\'a}n, et~al.
\newblock Evaluating generalization capabilities of llm-based agents in
  mixed-motive scenarios using concordia.
\newblock In \emph{The Thirty-ninth Annual Conference on Neural Information
  Processing Systems Datasets and Benchmarks Track}, 2025.

\bibitem[Suarez et~al.(2019)Suarez, Du, Isola, and Mordatch]{suarez2019neural}
Joseph Suarez, Yilun Du, Phillip Isola, and Igor Mordatch.
\newblock Neural mmo: A massively multiagent game environment for training and
  evaluating intelligent agents.
\newblock \emph{arXiv preprint arXiv:1903.00784}, 2019.

\bibitem[Suarez et~al.(2023)Suarez, Bloomin, Choe, Li, Sullivan, Kanna, Scott,
  Shuman, Bradley, Castricato, et~al.]{suarez2023neural}
Joseph Suarez, David Bloomin, Kyoung~Whan Choe, Hao~Xiang Li, Ryan Sullivan,
  Nishaanth Kanna, Daniel Scott, Rose Shuman, Herbie Bradley, Louis Castricato,
  et~al.
\newblock Neural mmo 2.0: a massively multi-task addition to massively
  multi-agent learning.
\newblock \emph{Advances in Neural Information Processing Systems},
  36:\penalty0 50094--50104, 2023.

\bibitem[Sun et~al.(2025)Sun, Zhang, Niu, Ren, Xu, Fu, Zhao, Yuan, and
  Wang]{sun2025collab}
Haochen Sun, Shuwen Zhang, Lujie Niu, Lei Ren, Hao Xu, Hao Fu, Fangkun Zhao,
  Caixia Yuan, and Xiaojie Wang.
\newblock Collab-overcooked: Benchmarking and evaluating large language models
  as collaborative agents.
\newblock In \emph{Proceedings of the 2025 Conference on Empirical Methods in
  Natural Language Processing}, pages 4922--4951, 2025.

\bibitem[Tessera et~al.(2025)Tessera, Rahman, Storkey, and
  Albrecht]{tessera2025hypermarl}
{Kale-ab}~Abebe Tessera, Arrasy Rahman, Amos Storkey, and Stefano~V Albrecht.
\newblock Hypermarl: Adaptive hypernetworks for multi-agent rl.
\newblock In \emph{The Thirty-ninth Annual Conference on Neural Information
  Processing Systems}, 2025.
\newblock URL \url{https://openreview.net/forum?id=56CgYnf9Dr}.

\bibitem[Tessera et~al.(2026)Tessera, Hinckeldey, Zamboni, Abel, and
  Storkey]{tessera2026probing}
Kale-ab~Abebe Tessera, Leonard Hinckeldey, Riccardo Zamboni, David Abel, and
  Amos Storkey.
\newblock Probing dec-{POMDP} reasoning in cooperative {MARL}.
\newblock In \emph{The 25th International Conference on Autonomous Agents and
  Multi-Agent Systems (AAMAS), Oral}, 2026.

\bibitem[Vaswani et~al.(2017)Vaswani, Shazeer, Parmar, Uszkoreit, Jones, Gomez,
  Kaiser, and Polosukhin]{vaswani2017attention}
Ashish Vaswani, Noam Shazeer, Niki Parmar, Jakob Uszkoreit, Llion Jones,
  Aidan~N Gomez, {\L}ukasz Kaiser, and Illia Polosukhin.
\newblock Attention is all you need.
\newblock \emph{Advances in neural information processing systems}, 30, 2017.

\bibitem[Vezhnevets et~al.(2023)Vezhnevets, Agapiou, Aharon, Ziv, Matyas,
  Du{\'e}{\~n}ez-Guzm{\'a}n, Cunningham, Osindero, Karmon, and
  Leibo]{vezhnevets2023generative}
Alexander~Sasha Vezhnevets, John~P Agapiou, Avia Aharon, Ron Ziv, Jayd Matyas,
  Edgar~A Du{\'e}{\~n}ez-Guzm{\'a}n, William~A Cunningham, Simon Osindero,
  Danny Karmon, and Joel~Z Leibo.
\newblock Generative agent-based modeling with actions grounded in physical,
  social, or digital space using concordia.
\newblock \emph{arXiv preprint arXiv:2312.03664}, 2023.

\bibitem[Vinyals et~al.(2019)Vinyals, Babuschkin, Czarnecki, Mathieu, Dudzik,
  Chung, Choi, Powell, Ewalds, Georgiev, et~al.]{vinyals2019grandmaster}
Oriol Vinyals, Igor Babuschkin, Wojciech~M Czarnecki, Micha{\"e}l Mathieu,
  Andrew Dudzik, Junyoung Chung, David~H Choi, Richard Powell, Timo Ewalds,
  Petko Georgiev, et~al.
\newblock Grandmaster level in starcraft ii using multi-agent reinforcement
  learning.
\newblock \emph{nature}, 575\penalty0 (7782):\penalty0 350--354, 2019.

\bibitem[Wang et~al.(2024{\natexlab{a}})Wang, Xie, Jiang, Mandlekar, Xiao, Zhu,
  Fan, and Anandkumar]{wang2023voyager}
Guanzhi Wang, Yuqi Xie, Yunfan Jiang, Ajay Mandlekar, Chaowei Xiao, Yuke Zhu,
  Linxi Fan, and Anima Anandkumar.
\newblock Voyager: An open-ended embodied agent with large language models.
\newblock \emph{Transactions on Machine Learning Research}, 2024{\natexlab{a}}.
\newblock ISSN 2835-8856.
\newblock URL \url{https://openreview.net/forum?id=ehfRiF0R3a}.

\bibitem[Wang et~al.(2024{\natexlab{b}})Wang, Zhang, Feng, Wang, and
  Tang]{wang2024battleagentbench}
Wei Wang, Dan Zhang, Tao Feng, Boyan Wang, and Jie Tang.
\newblock Battleagentbench: A benchmark for evaluating cooperation and
  competition capabilities of language models in multi-agent systems.
\newblock \emph{arXiv preprint arXiv:2408.15971}, 2024{\natexlab{b}}.

\bibitem[Wang et~al.(2025)Wang, Han, Diaz, Xu, R{\"u}hle, and
  Rajmohan]{wang2025odysseybench}
Weixuan Wang, Dongge Han, Daniel~Madrigal Diaz, Jin Xu, Victor R{\"u}hle, and
  Saravan Rajmohan.
\newblock Odysseybench: Evaluating llm agents on long-horizon complex office
  application workflows.
\newblock \emph{arXiv preprint arXiv:2508.09124}, 2025.

\bibitem[White et~al.(2025)White, Nottingham, Maniar, Robinson, Lillemark,
  Maheshwari, Qin, and Ammanabrolu]{white2025collaborating}
Isadora White, Kolby Nottingham, Ayush Maniar, Max Robinson, Hansen Lillemark,
  Mehul Maheshwari, Lianhui Qin, and Prithviraj Ammanabrolu.
\newblock Collaborating action by action: A multi-agent llm framework for
  embodied reasoning.
\newblock \emph{arXiv preprint arXiv:2504.17950}, 2025.

\bibitem[Yamada et~al.(2025)Yamada, Lange, Lu, Hu, Lu, Foerster, Clune, and
  Ha]{yamada2025ai}
Yutaro Yamada, Robert~Tjarko Lange, Cong Lu, Shengran Hu, Chris Lu, Jakob
  Foerster, Jeff Clune, and David Ha.
\newblock The ai scientist-v2: Workshop-level automated scientific discovery
  via agentic tree search.
\newblock \emph{arXiv preprint arXiv:2504.08066}, 2025.

\bibitem[Yang et~al.(2025)Yang, Chen, Siew, Botran, and
  Joe-Wong]{yang2025llmpowered}
Hanqing Yang, Jingdi Chen, Marie Siew, Tania~Lorido Botran, and Carlee
  Joe-Wong.
\newblock {LLM}-powered decentralized generative agents with adaptive
  hierarchical knowledge graph for cooperative planning.
\newblock In \emph{The First MARW: Multi-Agent AI in the Real World Workshop at
  AAAI 2025}, 2025.
\newblock URL \url{https://openreview.net/forum?id=l9QUw0oUTa}.

\bibitem[Yang et~al.(2026)Yang, Nourzad, Chen, Siew, Chen, and
  Joe-Wong]{yang2026emcoop}
Hanqing Yang, Narjes Nourzad, Shiyu Chen, Marie Siew, Jingdi Chen, and Carlee
  Joe-Wong.
\newblock {{COOP}$^2$}: Defining, observing, and repairing cooperation in {LLM}
  multi-agent systems.
\newblock \emph{arXiv preprint arXiv:2603.00349}, 2026.

\bibitem[Yao et~al.(2023)Yao, Zhao, Yu, Du, Shafran, Narasimhan, and
  Cao]{yao2023react}
Shunyu Yao, Jeffrey Zhao, Dian Yu, Nan Du, Izhak Shafran, Karthik~R Narasimhan,
  and Yuan Cao.
\newblock React: Synergizing reasoning and acting in language models.
\newblock In \emph{The Eleventh International Conference on Learning
  Representations}, 2023.
\newblock URL \url{https://openreview.net/forum?id=WE_vluYUL-X}.

\bibitem[Ye et~al.(2025)Ye, Tao, and Jaques]{ye2025efficient}
Eric Ye, Ren Tao, and Natasha Jaques.
\newblock An efficient open world environment for multi-agent social learning.
\newblock \emph{arXiv preprint arXiv:2508.15679}, 2025.

\bibitem[Yu et~al.(2022)Yu, Velu, Vinitsky, Gao, Wang, Bayen, and
  Wu]{yu2022surprising}
Chao Yu, Akash Velu, Eugene Vinitsky, Jiaxuan Gao, Yu~Wang, Alexandre Bayen,
  and Yi~Wu.
\newblock The surprising effectiveness of ppo in cooperative multi-agent games.
\newblock \emph{Advances in neural information processing systems},
  35:\penalty0 24611--24624, 2022.

\bibitem[Zhang et~al.(2026)Zhang, Jiang, Li, Tu, Su, Deng, Guo, Lv, and
  Lin]{zhang2026deepplanning}
Yinger Zhang, Shutong Jiang, Renhao Li, Jianhong Tu, Yang Su, Lianghao Deng,
  Xudong Guo, Chenxu Lv, and Junyang Lin.
\newblock Deepplanning: Benchmarking long-horizon agentic planning with
  verifiable constraints.
\newblock \emph{arXiv preprint arXiv:2601.18137}, 2026.

\bibitem[Zhao et~al.(2026)Zhao, Yuan, Huang, Yuan, Yu, Xu, Hu, Shankarampeta,
  Huang, Ni, et~al.]{zhao2026ama}
Yujie Zhao, Boqin Yuan, Junbo Huang, Haocheng Yuan, Zhongming Yu, Haozhou Xu,
  Lanxiang Hu, Abhilash Shankarampeta, Zimeng Huang, Wentao Ni, et~al.
\newblock Ama-bench: Evaluating long-horizon memory for agentic applications.
\newblock \emph{arXiv preprint arXiv:2602.22769}, 2026.

\bibitem[Zhu et~al.(2025)Zhu, Du, Hong, Yang, Guo, Wang, Wang, Qian, Tang, Ji,
  et~al.]{zhu2025multiagentbench}
Kunlun Zhu, Hongyi Du, Zhaochen Hong, Xiaocheng Yang, Shuyi Guo, Daisy~Zhe
  Wang, Zhenhailong Wang, Cheng Qian, Robert Tang, Heng Ji, et~al.
\newblock Multiagentbench: Evaluating the collaboration and competition of llm
  agents.
\newblock In \emph{Proceedings of the 63rd Annual Meeting of the Association
  for Computational Linguistics (Volume 1: Long Papers)}, pages 8580--8622,
  2025.

\end{thebibliography}

\setcounter{topnumber}{2}
\appendix

\newpage

\definecolor{cmarkgreen}{RGB}{46,139,87}
\definecolor{xmarkred}{RGB}{178,34,34}
\definecolor{pmarkorange}{RGB}{210,120,20}
\definecolor{headerblue}{RGB}{48,89,160}

\newcommand{\cmark}{\textcolor{cmarkgreen}{\ding{51}}}
\newcommand{\xmark}{\textcolor{xmarkred}{\ding{55}}}
\newcommand{\pmark}{\textcolor{pmarkorange}{$\sim$}}

\section{\textsc{alem}'s Place Among Existing Environments}

Table~\ref{tab:environment_comparison} compares \textsc{alem} with representative interactive multi-agent environments. We compare environments using four criteria. \emph{Long horizon} means that success requires maintaining progress over extended trajectories, with rewards delayed by $T$ timesteps where $T$ is typically in the hundreds or thousands. \emph{Bounded open-endedness} means that the environment supports a diverse but finite set of achievable goals $G=\{g_1,\dots,g_n\}$. This differs from fully open-ended systems, where $n$ can grow over time through the generation of novel and learnable goals~\citep{pmlr-v235-hughes24a}. \emph{Explicit coordination requirements} means that the environment contains identifiable multi-agent task structures, such as synchronous actions, handovers, role allocation, or long-range interdependence. \emph{Coordination difficulty control} means that the benchmark exposes an explicit parameter or protocol for scaling coordination difficulty.

\begin{table}[b]
\centering
\captionsetup{position=top,skip=10pt}
\caption{\textbf{Comparison with representative multi-agent interactive environments.}
\cmark indicates that the property is a central design feature, \xmark indicates that it is absent or not a primary design goal, and \pmark indicates restricted support. For \emph{long horizon}, restricted support means that only some scenarios require extended trajectories or delayed outcomes. For \emph{explicit coordination requirements}, restricted support means that coordination-relevant interactions exist, but are fixed, implicit, narrow in type, or not procedurally generated as explicit task requirements. \emph{Coordination difficulty control} requires an explicit parameter or protocol for scaling coordination difficulty, rather than only changing maps, scenarios, opponents, or task instances.}

\small
\begin{adjustbox}{max width=\linewidth}
\begin{tabular}{lcccc}
\toprule
\textcolor{headerblue}{\textbf{Environment}} &
\textbf{Long} &
\textbf{Bounded} &
\textbf{Explicit coord.} &
\textbf{Coord. difficulty} \\
&
\textbf{horizon} &
\textbf{open-endedness} &
\textbf{requirements} &
\textbf{control} \\
\midrule
SMAC / SMACv2/ SMAX~\citep{samvelyan2019starcraft,ellis2023smacv2,rutherford2024jaxmarl}
  & \xmark & \xmark & \pmark & \xmark \\
Overcooked / Overcookedv2~\citep{carroll2019utility,gessler2025overcookedv}
  & \xmark & \xmark & \pmark & \xmark \\
Hanabi~\citep{bard2020hanabi}
  & \xmark & \xmark & \pmark & \xmark \\
Melting Pot~\citep{agapiou2022melting,leibo2021scalable}
  & \pmark & \xmark & \pmark & \xmark \\
Concordia~\citep{smith2025evaluating,vezhnevets2023generative}
  & \xmark & \xmark & \pmark & \xmark \\
HECOGrid~\citep{liu2023stateful}
  & \xmark & \xmark & \cmark & \cmark \\
GPUDrive~\citep{kazemkhani2025gpudrive}
  & \xmark & \xmark & \xmark & \xmark \\
Neural MMO~\citep{suarez2019neural,suarez2023neural}
  & \cmark & \cmark & \pmark & \xmark \\
Multi-Agent Craftax v1~\citep{ye2025efficient}
  & \cmark & \cmark & \pmark & \xmark \\
Craftax-Coop~\citep{omari2025multi}
  & \cmark & \cmark & \pmark & \xmark \\
\textsc{alem}
  & \cmark & \cmark & \cmark & \cmark \\
\bottomrule
\end{tabular}
\end{adjustbox}

\label{tab:environment_comparison}
\end{table}

Existing environments cover important parts of this space, but not the full combination targeted by \textsc{alem}. Classic MARL benchmarks such as SMAC, Overcooked, and Hanabi isolate useful coordination motifs, but are short-horizon and task-specific. Open-world environments such as Neural MMO and Craftax-Coop provide longer horizons and bounded open-endedness, but do not expose explicit controls for coordination difficulty. \textsc{alem} combines these ingredients by adding procedurally generated coordination tasks, soft specialisation, communication, and a difficulty parameter that scales coordination demands while preserving the underlying world structure.

The closest environments are Multi-Agent Craftax v1~\citep{ye2025efficient} and Craftax-Coop~\citep{omari2025multi}. Both inherit Craftax's long-horizon, bounded open-ended structure, but their coordination demands are mostly implicit. Multi-Agent Craftax v1 places multiple agents in the same open world, but does not introduce explicit coordination tasks, a controllable coordination-difficulty axis, or the full nine-level progression used in Craftax and \textsc{alem} (it is implemented only on the overworld). Craftax-Coop adds roles, creating long-range interdependence through specialisation. However, it does not procedurally generate coordination requirements over entities, require synchronous or handover coordination, provide an explicit communication channel, or expose a parameter that scales coordination difficulty while preserving the same world. \textsc{alem} adds these mechanisms directly: coordination tasks are sampled each episode, span synchronous and handover structures, vary the required number of agents, and are controlled by a single difficulty parameter.

\clearpage

\section{Environment Details}\label{app:env_details}

 \begin{figure}[h]
    \centering
    \includegraphics[width=0.42\linewidth]{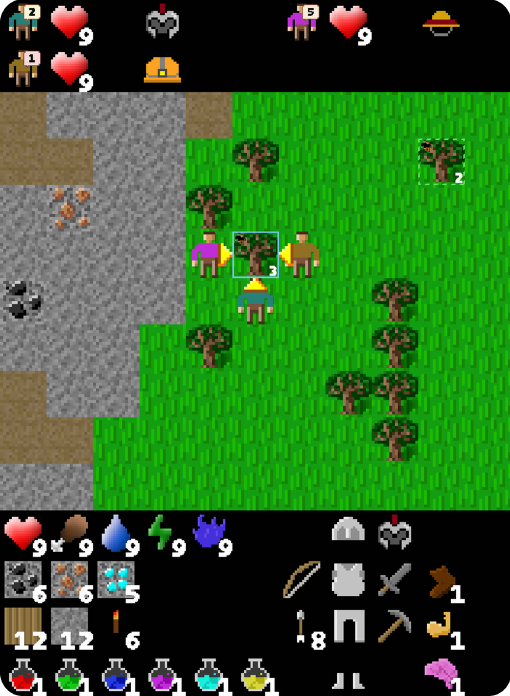}
    \captionsetup{justification=raggedright,singlelinecheck=false}
    \caption{Example pixel-based observation in \textsc{Alem}. \textbf{Top bar:} teammate status: role icon (forager/miner/warrior), health, and a coloured channel badge showing the discrete broadcast message each teammate sent at this step (badge absent when the agent did not communicate). \textbf{Centre:} the agent's local world view: coordination overlays mark blocks and sites that require joint action, a solid border with an agent-count label denotes a \emph{hard} synchronisation, a dashed border denotes a \emph{soft} requirement, and a timer badge marks an active \emph{handover} window within which a second agent must act. \textbf{Bottom:} the agent's inventory and attributes (resources, tools, armour, potions, enchantments). \textsc{Alem} also provides a symbolic observation (a structured array of nearby blocks, teammate states, and inventory tensors) and a text-based observation used by the LLM-agent baselines.}
  \label{fig:alem_pixel_obs}
\end{figure}

\begin{figure}[t]
  \centering

  \newlength{\mapw}
  \setlength{\mapw}{0.295\linewidth}   %
  \newlength{\lblgap}
  \setlength{\lblgap}{2pt}             %

  \begin{minipage}[c]{0.012\linewidth}
    \rotatebox{90}{\footnotesize\textit{Diversity}}
  \end{minipage}\hspace{\lblgap}%
  \begin{minipage}[c]{\mapw}
    \includegraphics[width=\linewidth]{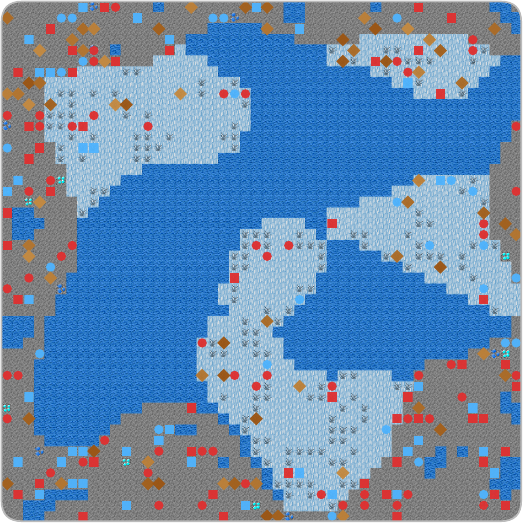}\\[-1pt]
    \centering{\scriptsize Seed 0}
  \end{minipage}\hfill
  \begin{minipage}[c]{\mapw}
    \includegraphics[width=\linewidth]{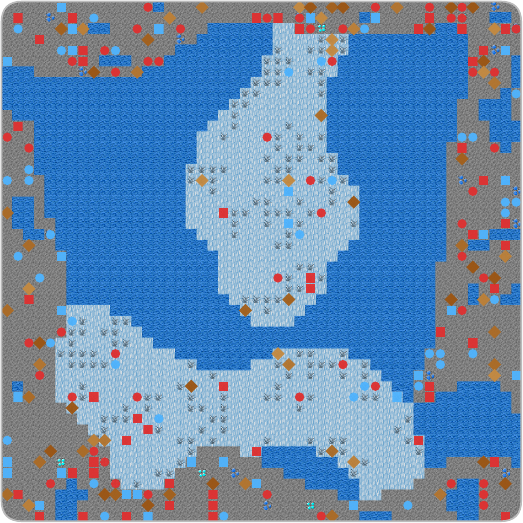}\\[-1pt]
    \centering{\scriptsize Seed 1}
  \end{minipage}\hfill
  \begin{minipage}[c]{\mapw}
    \includegraphics[width=\linewidth]{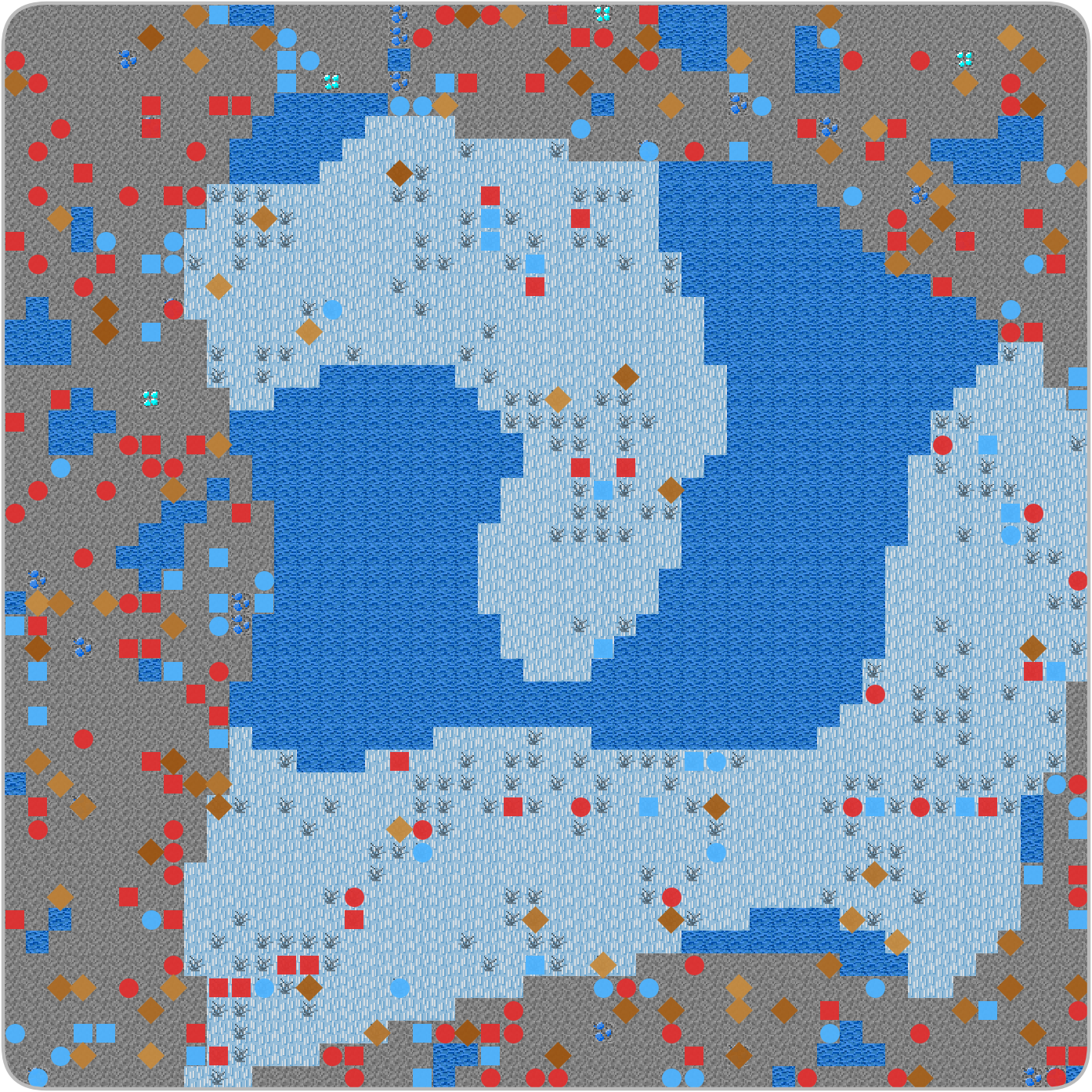}\\[-1pt]
    \centering{\scriptsize Seed 2}
  \end{minipage}

  \vspace{5pt}

  \includegraphics[width=0.82\linewidth]{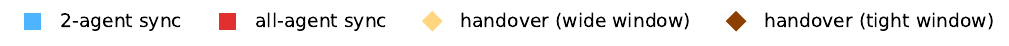}

  \caption{%
    \textbf{\textsc{alem} procedurally generates diverse worlds and coordination layouts.}
    Three independently sampled episodes at the same medium difficulty show variation in both terrain layout and the spatial placement of coordination tasks (\textcolor[HTML]{4FB4FF}{$\blacksquare$}~2-agent sync,
     \textcolor[HTML]{E03030}{$\blacksquare$}~all-agent sync,
     \textcolor[HTML]{FFD580}{$\blacklozenge$}~handover).
  }
  \label{fig:world-diversity}
\end{figure}

\subsection{Probing Environment Changes}
\label{app:env_probing}

We use the information-theoretic diagnostics proposed in \citet{tessera2026probing} to check whether the design changes in \textsc{alem} introduce the intended behavioural demands. We roll out trained RL policies under the same four settings used in the environment calibration: a single-agent setting, a multi-agent setting without role pressure or coordination tasks, a multi-agent setting with soft specialisation, and full \textsc{alem} (\textsc{Easy}). The diagnostics are reported as behavioural questions in Table~\ref{tab:behavioral_metrics}, with the corresponding metric acronym shown in parentheses.

All four diagnostic signals increase as multi-agent structure is added. This suggests that full \textsc{alem} induces stronger demands for memory use, hidden teammate information, synchronous coordination, and temporal coordination, rather than simply making the base game harder. We report diagnostic magnitudes rather than Wilcoxon tests because this is a single-environment calibration across settings, not a suite-level comparison across independent scenarios.

\begin{table*}[tb]
\centering
\caption{\textbf{Behavioural diagnostics across environment settings.}
We use the diagnostics of \citep{tessera2026probing} on the same four settings used in the environment calibration: single-agent, multi-agent without role pressure or coordination tasks, multi-agent with soft specialisation, and full \textsc{alem} (\textsc{Easy}). Signals for memory use, hidden teammate information, synchronous coordination, and temporal coordination increase as these components are added. Higher values indicate stronger statistical dependence between agents' actions and the corresponding information source or coordination relationship, not higher return. Values are mean $\pm$ standard deviation over seeds.}
\label{tab:behavioral_metrics}
\vspace{2pt}
\begingroup
\setlength{\tabcolsep}{5pt}
\footnotesize
\begin{tabular}{@{}p{5.3cm}cccc@{}}
\toprule
Behavioural question &
Single-agent &
Multiple agent &
MA + Soft spec. &
\textsc{alem} \\
\midrule
Do agents benefit from memory? (HAR) &
$0.355 \pm 0.142$ &
$0.372 \pm 0.022$ &
$0.437 \pm 0.123$ &
$0.641 \pm 0.064$ \\

Does teammate info. help predict actions? (PIF) &
-- &
$0.314 \pm 0.036$ &
$0.384 \pm 0.125$ &
$0.609 \pm 0.148$ \\

Does synchronous coordination emerge? (AA) &
-- &
$0.017 \pm 0.004$ &
$0.032 \pm 0.011$ &
$0.061 \pm 0.022$ \\

Does temporal coordination emerge? (DAI) &
-- &
$0.319 \pm 0.057$ &
$0.406 \pm 0.128$ &
$0.613 \pm 0.165$ \\
\bottomrule
\end{tabular}
\endgroup
\end{table*}

\subsection{Coordination Tasks}\label{app:coord_details}

\subsubsection{Coordination Task Categories}

We instantiate coordination across four activity categories:
\begin{itemize}[leftmargin=*]
    \item \textbf{Mining.} Resource blocks (including trees, ores, gems) are sampled for coordination during world generation. Synchronous mining requires the assigned number of agents to act on the same block in the same step. Handover mining requires a second agent to mine the block within $\Delta t \in [\Delta t_{\min}, \Delta t_{\max}]$ steps of the first agent.

    \item \textbf{Construction.} Construction sites for shelters, forges, and beacons are placed on valid overworld and mining-floor tiles, independently of mining blocks. Each site receives the same synchronous or handover structure as mining, together with a soft or hard synchronous variant. Building also requires the appropriate resources for the chosen structure. In handover construction, one agent initiates a specific structure, and a second agent must complete that structure type before the deadline, otherwise the site reverts and materials are refunded to the initiator.

    \item \textbf{Combat.} Mob slots are pre-labelled during world generation as standard or coordinated mobs. Coordinated mobs appear as either elite melee mobs, elite ranged mobs, or large passive mobs,  and each is marked as soft or hard and assigned the required agent count.

    \item \textbf{Crafting.} Diamond pickaxes, swords, and armour require synchronous coordination, i.e. the craft succeeds only when the required number of agents attempt it at the same Epic Forge in the same step. This is a deliberately restricted variant, it is hard coordination only and tied to a specific tool station, and tests whether agents can converge on a shared location for a one-shot joint action.
\end{itemize}

\subsection{Coordination Difficulty}\label{app:coordination_difficulty}

Table \ref{tab:coordination_difficulty_scaled} and \ref{tab:coordination_difficulty_fixed} describe what exactly changes across different difficulty levels.

\begin{table}[t]
\centering
\captionsetup{position=top,skip=6pt}
\caption{Coordination difficulty parameters controlled by the scalar $\alpha$. Easy, Medium, and Hard correspond to $\alpha \in \{0.3,0.6,0.9\}$. Larger $\alpha$ makes coordination harder to execute, but does not change how many coordination opportunities are sampled. $^\dagger$ Only used when \texttt{scale\_base\_difficulty=True}. In the main benchmark, coordination difficulty can be scaled independently of base survival difficulty.}
\label{tab:coordination_difficulty_scaled}
\small
\setlength{\tabcolsep}{5pt}
\begin{tabular}{lcccc}
\toprule
Parameter & Formula & Easy & Medium & Hard \\
\midrule
$p_{\max}$ & $\alpha$ & 0.30 & 0.60 & 0.90 \\
handover\_window\_min & $\max(3,\lceil 12(1-\alpha)\rceil)$ & 9 & 5 & 3 \\
handover\_window\_max & $\max(6,\lceil 24(1-\alpha)\rceil)$ & 17 & 10 & 6 \\
soft\_solo\_fail\_prob & $\alpha$ & 0.30 & 0.60 & 0.90 \\
non\_specialist\_efficiency & $1-\alpha$ & 0.70 & 0.40 & 0.10 \\
\midrule
mob\_health\_multiplier$^\dagger$ & $1+\alpha/3$ & 1.10 & 1.20 & 1.30 \\
starting\_resource\_multiplier$^\dagger$ & $1-\alpha/3$ & 0.90 & 0.80 & 0.70 \\
\bottomrule
\end{tabular}

\end{table}

\begin{table}[t]
\centering
\captionsetup{position=top,skip=6pt}
\caption{Coordination opportunity parameters held fixed across difficulty levels. These control how much coordination exists in the world, rather than how hard it is to execute.}
\label{tab:coordination_difficulty_fixed}
\small
\setlength{\tabcolsep}{5pt}
\begin{tabular}{lc}
\toprule
Parameter & Value \\
\midrule
coordination\_enabled & True \\
coordination\_probability & 0.25 \\
soft\_coordination\_ratio & 0.50 \\
handover\_ratio & 0.25 \\
hard\_mob\_probability & 0.50 \\
elite\_mob\_probability & 0.15 \\
large\_passive\_probability & 0.20 \\
construction\_enabled & True \\
num\_construction\_sites & 8 \\
num\_mining\_construction\_sites & 4 \\
soft\_construction\_ratio & 0.50 \\
crafting\_coordination\_enabled & True \\
diamond\_crafting\_agents\_required & 2 \\
soft\_specialization & True \\
\bottomrule
\end{tabular}
\end{table}

\newpage

\section{Detailed Experimental Results}

\begin{figure}[H]
    \centering
    \begin{minipage}[t]{0.32\textwidth}
        \centering
        \includegraphics[width=\textwidth]{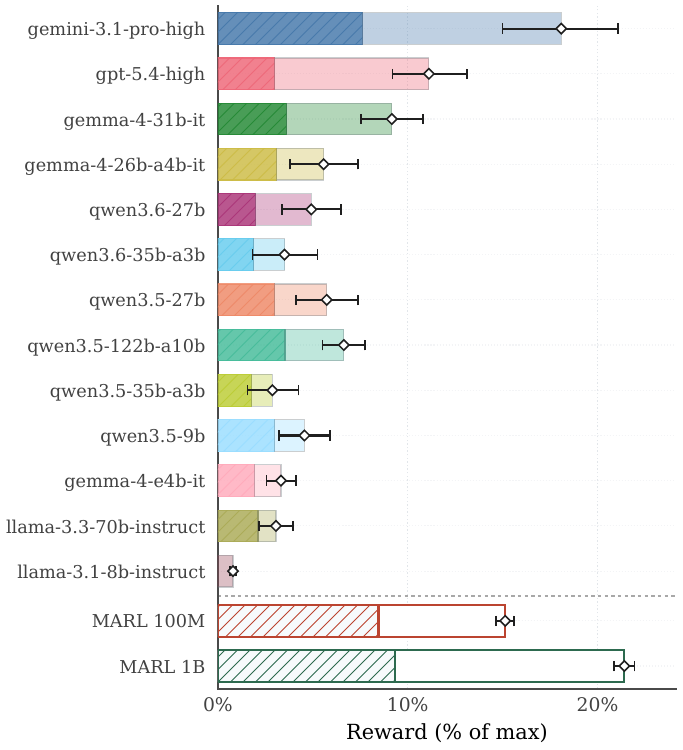}
        \vspace{-0.5em}
        \subcaption{Easy}
    \end{minipage}\hfill
    \begin{minipage}[t]{0.32\textwidth}
        \centering
        \includegraphics[width=\textwidth]{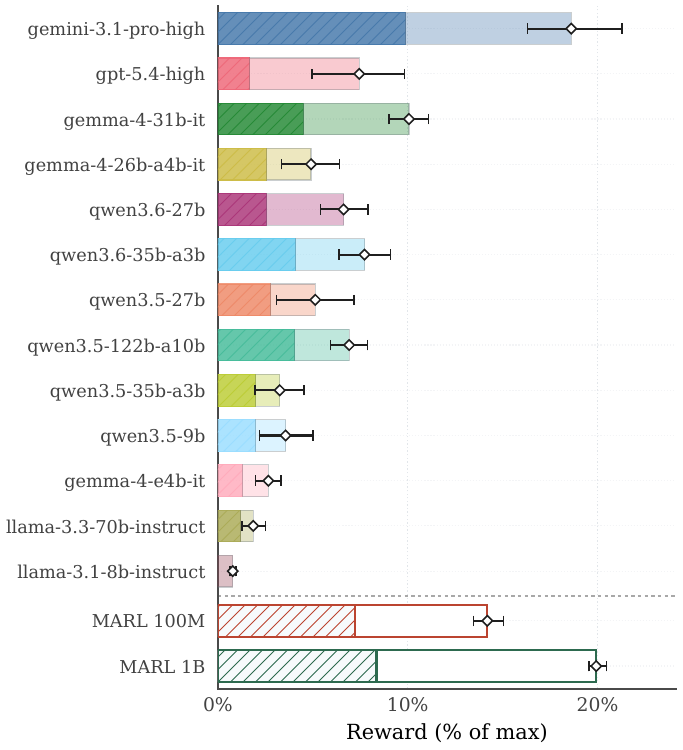}
        \vspace{-0.5em}
        \subcaption{Medium}
    \end{minipage}\hfill
    \begin{minipage}[t]{0.32\textwidth}
        \centering
        \includegraphics[width=\textwidth]{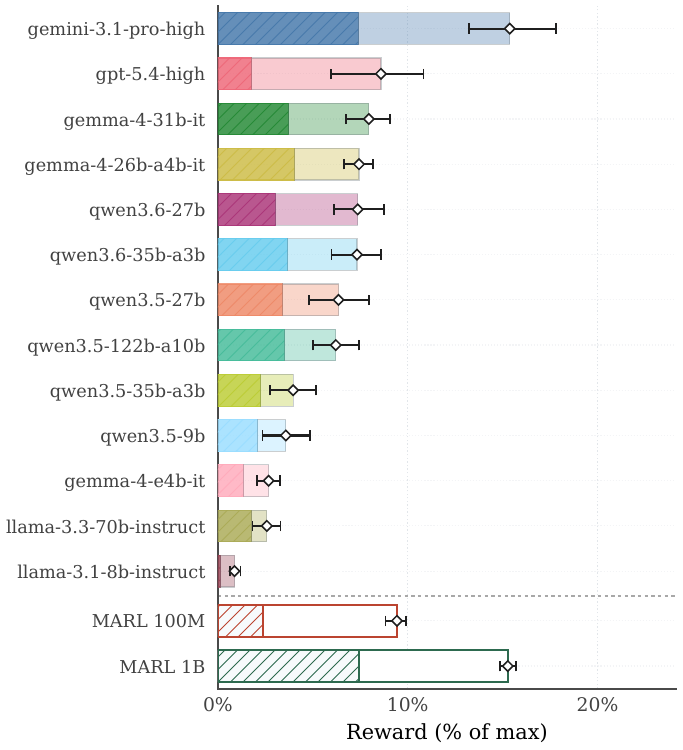}
        \vspace{-0.5em}
        \subcaption{Hard}
    \end{minipage}

    \vspace{0.3em}
    
    \includegraphics[width=0.85\textwidth]{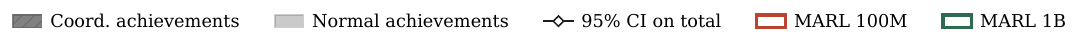}
   
    \caption{
Zero-shot homogeneous coordination in ALEM. We evaluate modern LLMs on Easy, Medium, and Hard difficulties, and compare them against RL baselines at 100M and 1B environment steps. Solid bars denote base reward, dashed bars denote coordination reward, and diamonds denote total reward. Values are reported as percentages of the maximum achievable total reward, with 95\% confidence intervals on total reward. The stacked base and coordination contributions sum to Total\%.
}
    \label{fig:main_results}
\end{figure}

\clearpage

\subsection{Algorithm Hyperparameters}
\label{app:hyperparameters}

The hyperparameters for the IPPO, MAPPO, HyperMARL, and PQN-VDN algorithms are summarised in Table~\ref{tab:hyperparameters}. These parameters were standardised where applicable to ensure comparable evaluation across multi-agent reinforcement learning baselines.

\begin{table}[htbp]
    \centering
    \captionsetup{position=top,skip=6pt}
    \caption{Hyperparameters for Multi-Agent Reinforcement Learning Algorithms.}
    \label{tab:hyperparameters}
    \begin{tabular}{@{}l c c c c@{}}
        \toprule
        \textbf{Parameter} & \textbf{IPPO} & \textbf{MAPPO} & \textbf{HyperMARL} & \textbf{PQN-VDN} \\
        \midrule
        \multicolumn{5}{c}{\textit{Training \& Environment}} \\
        \midrule
        Total Timesteps & $10^9$ & $10^9$ & $10^9$ & $10^9$ \\
        Number of Environments & $1024$ & $512$ & $1024$ & $512$ \\
        Steps per Environment & $64$ & $64$ & $64$ & $128$ \\
        Number of Epochs & $4$ & $4$ & $4$ & $4$ \\
        Number of Minibatches & $8$ & $8$ & $8$ & $16$ ($4^\dagger$) \\
        Learning Rate & $2 \times 10^{-4}$ & $2 \times 10^{-4}$ & $1 \times 10^{-4}$ & $5 \times 10^{-5}$ \\
        Max Gradient Norm & $1.0$ & $1.0$ & $1.0$ & $0.5$ \\
        Discount Factor ($\gamma$) & $0.99$ & $0.99$ & $0.99$ & $0.99$ \\
        \midrule
        \multicolumn{5}{c}{\textit{Algorithm-Specific Optimization}} \\
        \midrule
        GAE / Eligibility $\lambda$ & $0.8$ & $0.8$ & $0.8$ & $0.9$ \\
        PPO Clip Ratio ($\epsilon$) & $0.2$ & $0.2$ & $0.2$ & -- \\
        Entropy Coefficient & $0.01$ & $0.01$ & $0.01$ & -- \\
        Value Function Coefficient & $0.5$ & $0.5$ & $0.5$ & -- \\
        $\epsilon$-greedy Start & -- & -- & -- & $1.0$ \\
        $\epsilon$-greedy Finish & -- & -- & -- & $0.01$ ($0.005^\dagger$) \\
        $\epsilon$-greedy Decay & -- & -- & -- & $0.1$ \\
        \midrule
        \multicolumn{5}{c}{\textit{Network Architecture}} \\
        \midrule
        RNN Hidden Dimension & $512$ & $512$ & $512$ & $1024$ \\
        FC Layer Dimension & $128$ & $128$ & $128$ & -- \\
        Activation Function & \texttt{tanh} & \texttt{tanh} & \texttt{relu} & -- \\
        Normalization Type & -- & -- & -- & Layer Norm \\
        \midrule
        \multicolumn{5}{c}{\textit{HyperMARL-Specific}} \\
        \midrule
        Hypernet Embedding Dim & -- & -- & $64$ & -- \\
        Hypernet Hidden Dims & -- & -- & $[64]$ & -- \\
        Hypernet Init Scale & -- & -- & $\sqrt{2}$ & -- \\
        \bottomrule
        \multicolumn{5}{l}{\footnotesize $^\dagger$ Denotes alternative configurations used in specific cooperative PQN variants.}
    \end{tabular}
\end{table}

\paragraph{Hyperparameter sweeps.}
To ensure a fair comparison, IPPO, MAPPO, and HyperMARL were tuned over an identical grid (Table~\ref{tab:sweeps}, top), with $5$ seeds per configuration. PQN-VDN was tuned separately over a normalisation, eligibility-trace, and learning-rate grid with $3$ seeds per configuration (Table~\ref{tab:sweeps}, bottom). All sweeps were run on \textsc{alem-lite} (the overworld-only variant, App.~\ref{app:sps}) through the symbolic interface, with non-shared rewards, soft specialisation, and action masking; the PPO-family grid was swept at Hard difficulty and the PQN-VDN grid at Easy. The configurations selected (in bold) are those used for the main results in Table~\ref{tab:hyperparameters}.

  \begin{table}[htbp]
      \centering
      \captionsetup{position=top,skip=6pt}
      \caption{Hyperparameter sweep grids and selected values (\textbf{bold}). The PPO-family grid is shared across IPPO, MAPPO, and HyperMARL.}
      \label{tab:sweeps}                                                                                                                                                       
      \begin{tabular}{@{}l l@{}}                                                                           \toprule
\textbf{Hyperparameter} & \textbf{Sweep range} \\                                                         \midrule                                                                                                  \multicolumn{2}{c}{\textit{PPO-family (IPPO / MAPPO / HyperMARL), $5$ seeds per config}} \\
 \midrule                                                                                                 Learning rate & $\{4\!\times\!10^{-3},\ 1\!\times\!10^{-3},\ 3\!\times\!10^{-4}, 2\!\times\!10^{-4},\ 1\!\times\!10^{-4}\}$ \\
 IPPO/MAPPO - $\mathbf{2\!\times\!10^{-4}}$; HyperMARL - $\mathbf{1\!\times\!10^{-4}}$ \\   PPO clip ratio $\epsilon$ & $\{0.1,\ \mathbf{0.2}\}$ \\                                                  Update epochs & $\{2,\ \mathbf{4}\}$ \\                                                                  Max gradient norm & $\{0.5,\ \mathbf{1.0}\}$ \\                                                                                                           \midrule                                                                                        \multicolumn{2}{c}{\textit{PQN-VDN, $3$ seeds per config}} \\                                   \midrule                                                                                        Learning rate & $\{3\!\times\!10^{-4},\ 1\!\times\!10^{-4},\ \mathbf{5\!\times\!10^{-5}},\ 3\!\times\!10^{-5},\ 1\!\times\!10^{-5}\}$ \\
          Number of minibatches & $\{4,\ 8,\ \mathbf{16}\}$ \\                                            Eligibility trace $\lambda$ & $\{0.5,\ 0.7,\ \mathbf{0.9}\}$ \\                                 Normalization type & $\{\mathbf{\text{layer\_norm}},\ \text{batch\_norm}\}$ \\                  Normalize inputs & $\{$true,\ \textbf{false}$\}$ \\                                             $\epsilon$-greedy finish & $\{\mathbf{0.01},\ 0.005\}$ \\                                       Number of epochs & $\{2,\mathbf{4}\}$ \\                                          Recurrent cell & $\{\textbf{LSTM},\ \text{GRU}\}$ \\                                            \bottomrule                                                                                                                                                          
      \end{tabular}                                                                                                                                                            
  \end{table}                                                                                                                                                           
  \noindent The PQN-VDN sweep additionally revealed two non-default choices worth flagging: (i) although the original PureJaxQL Craftax recipe enables input normalization,    
  this consistently degraded performance in our setting and was disabled; and (ii) GRU was substantially worse than LSTM as the recurrent cell, despite GRU being used by all
  PPO-family baselines. The cooperative PQN variant ($^\dagger$ in Table~\ref{tab:hyperparameters}) used $4$ minibatches and $\epsilon$-finish $=0.005$.  
  \clearpage

\begin{figure*}[t]
  \centering
  \includegraphics[width=0.98\textwidth]{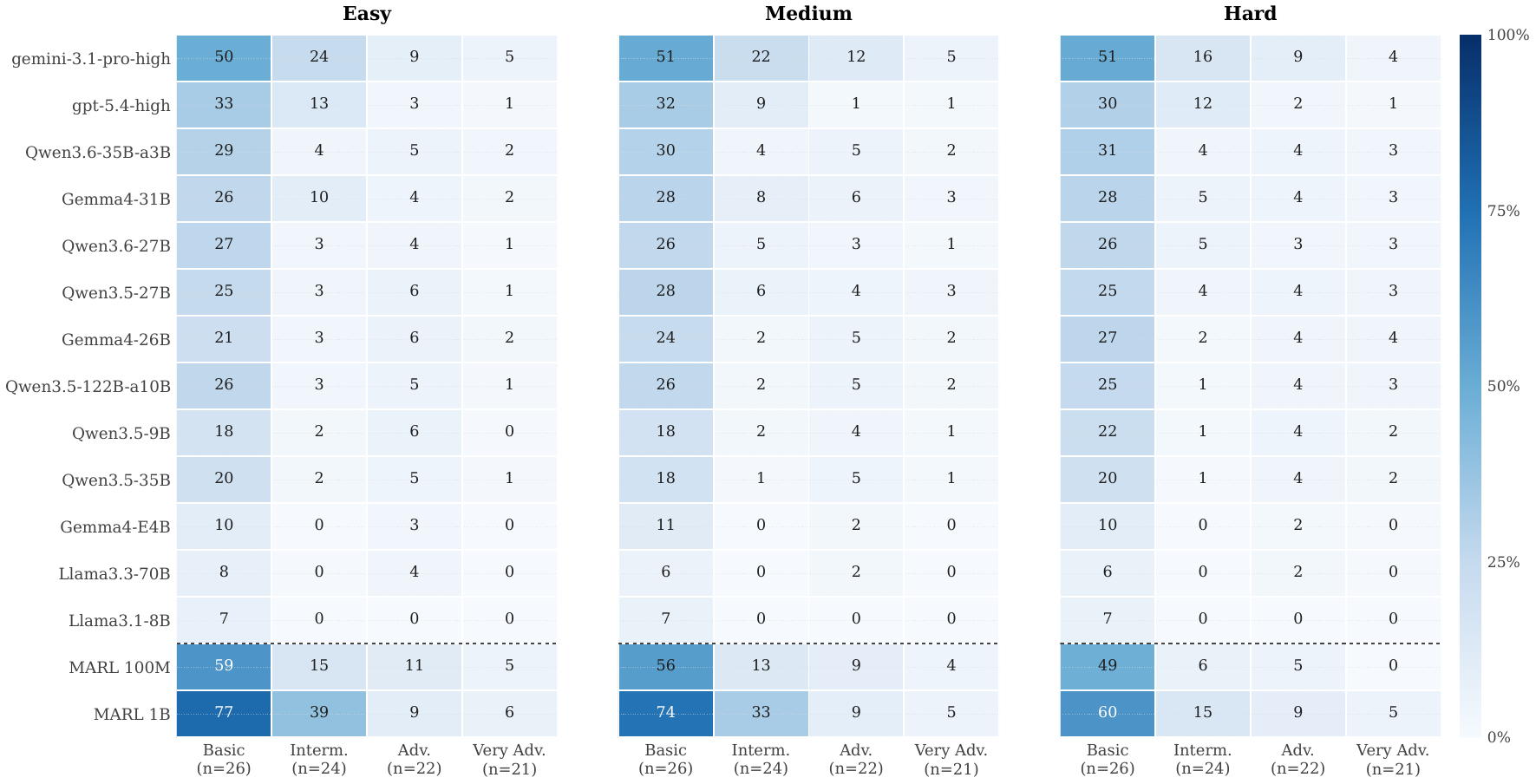}
  \caption{Achievement-tier coverage across coordination difficulty. Each cell reports the mean fraction of achievements completed within a tier by a method; columns group achievement tiers and panels separate coordination settings. Cell values are percentages and colour encodes coverage from 0\% to 100\%. MARL baselines are separated by a dashed rule, and methods are ordered by their total achievement count on the hard coordination setting.}
  \label{fig:achievement-tier-coverage}
\end{figure*}

\subsection{Additional Metrics}\label{app:additional_metrics}

Our primary metric is normalised team achievement reward. Each achievement contributes its reward once per episode if any agent unlocks it, we divide the episode total by the maximum reward available in the relevant category and average across episodes. We report coordination reward normalised by $R_{\max}^{\mathrm{coord}}=159$, base reward by $R_{\max}^{\mathrm{base}}=217$, and total reward by $R_{\max}^{\mathrm{total}}=376$.

\begin{figure}
    \centering
    \includegraphics[width=0.6\linewidth]{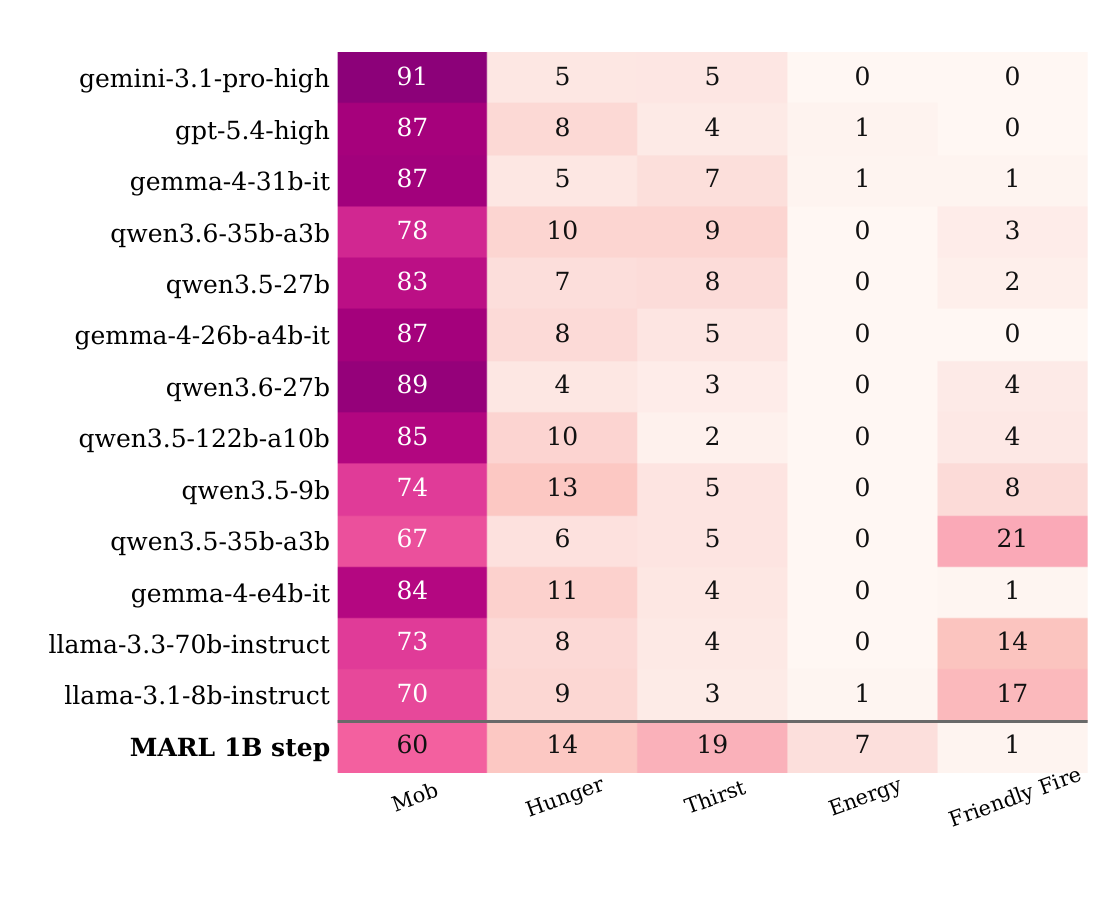}
    \caption{Different causes of death for different agents, averaged across easy, medium and hard.}
    \label{fig:death_causes}
\end{figure}

\begin{figure}
    \centering
    \includegraphics[width=0.7\linewidth]{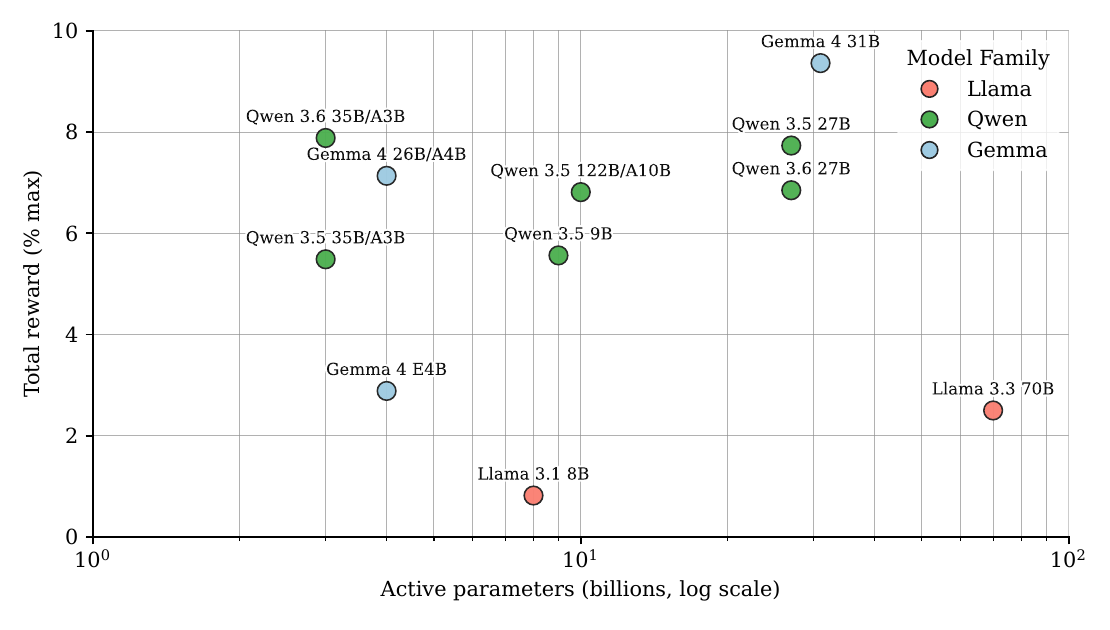}
    \caption{Performance vs active model size.}
    \label{fig:perf_vs_size}
\end{figure}
\begin{figure}[htbp]
    \centering
    \begin{minipage}{0.48\textwidth}
        \centering
        \includegraphics[width=\linewidth]{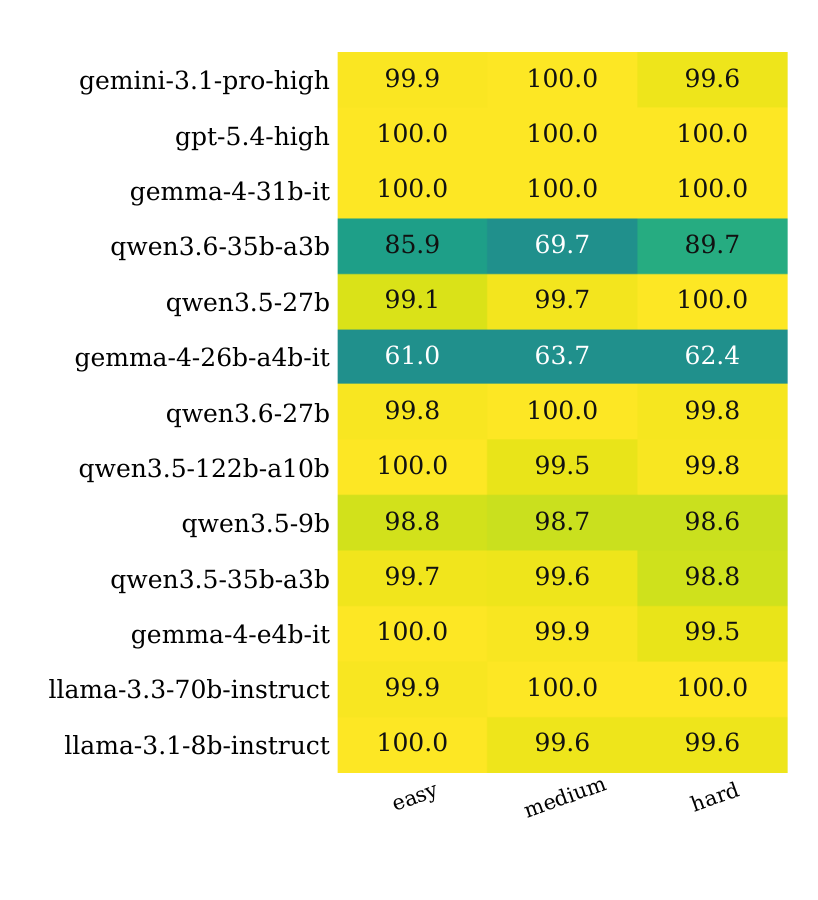}
        \caption{Percentage success rate for parsing actions.}
        \label{fig:action_parse}
    \end{minipage}
    \hfill %
    \begin{minipage}{0.48\textwidth}
        \centering
        \includegraphics[width=\linewidth]{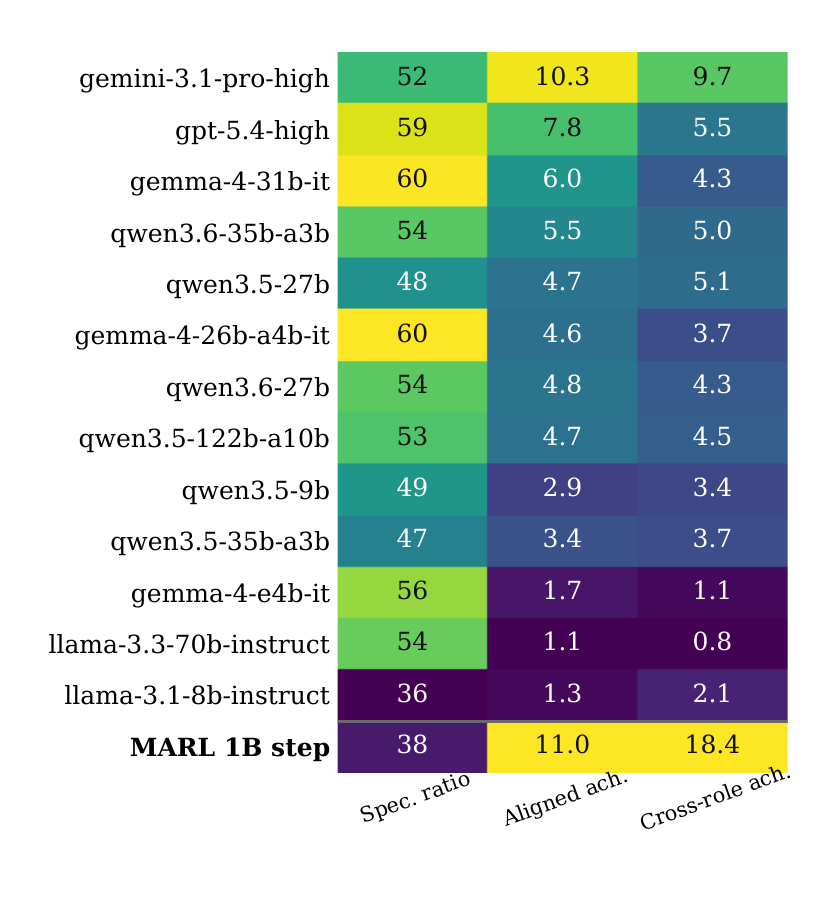}
        \caption{Specialisation metrics, averaged across easy, medium and hard.}
        \label{fig:specialisation_metrics}
    \end{minipage}
\end{figure}

\clearpage
\subsection{Environment Speed}
\label{app:sps}

We benchmark \textsc{alem}'s simulation speed against two baselines: Craftax-Coop~\citep{omari2025multi}, the multi-agent environment we extend, and Craftax~\citep{matthews2024craftax}, the original single-agent JAX environment. Measurements report steps per second (SPS) for a full IPPO training iteration (rollouts and updates, excluding logging) on a single NVIDIA L40S (48\,GB) GPU using JAX 0.4.38. \textsc{alem} is evaluated under its default training configuration (Easy difficulty, four communication channels).

\begin{figure}[h]
    \centering
    \begin{subfigure}[b]{0.49\linewidth}
        \centering
        \includegraphics[width=0.90\linewidth]{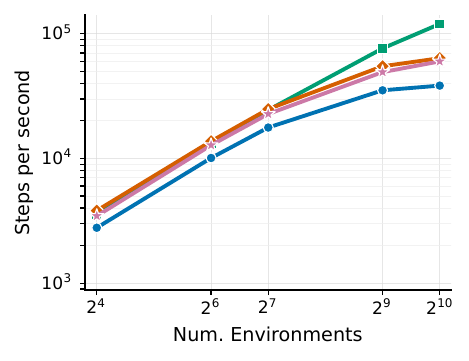}
        \caption{1 Agent}
    \end{subfigure}
    \hfill
    \begin{subfigure}[b]{0.49\linewidth}
        \centering
        \includegraphics[width=0.90\linewidth]{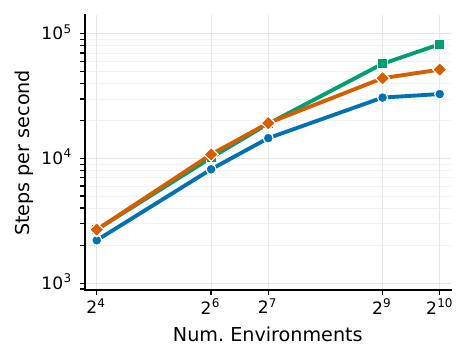}
        \caption{2 Agents}
    \end{subfigure}

    \vspace{0.5em}

    \begin{subfigure}[b]{0.49\linewidth}
        \centering
        \includegraphics[width=0.90\linewidth]{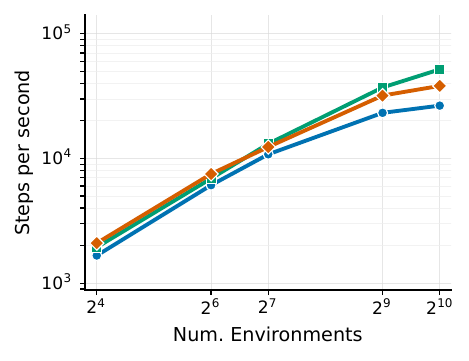}
        \caption{4 Agents}
    \end{subfigure}
    \hfill
    \begin{subfigure}[b]{0.49\linewidth}
        \centering
        \includegraphics[width=0.90\linewidth]{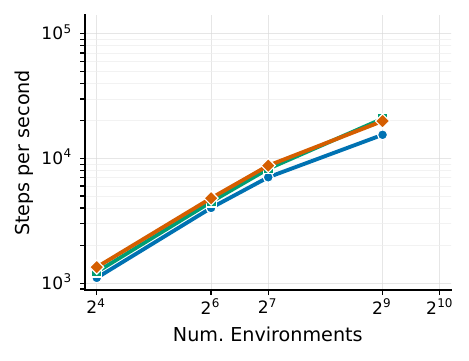}
        \caption{8 Agents}
    \end{subfigure}

    \vspace{0.5em}

    \begin{subfigure}[b]{\linewidth}
        \centering
        \includegraphics[width=0.76\linewidth]{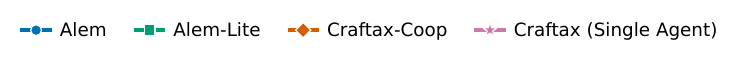}
    \end{subfigure}
    \caption{
        Simulation throughput (SPS) during a full IPPO training step across 1--8 agents. The single-agent Craftax baseline is included as a global reference.
    }
    \label{fig:sps}
\end{figure}

Figure~\ref{fig:sps} shows throughput scaling efficiently across 1 to 8 agents. The throughput reduction compared to Craftax-Coop directly reflects \textsc{alem}'s expanded mechanics: a richer observation space (coordination tracking and messages), coordination dynamics, additional tasks (building shelter, new passive mobs), soft specialisation, larger action space, and coordination reward logic. Crucially, the environment remains fully JIT-compiled; a 1B-timestep multi-agent training run completes in under two days on a single GPU, ensuring that large-scale experimentation remains highly accessible. 

For researchers prioritising execution speed, we also provide \textsc{alem-lite}. This provides the exact same coordination logic, but limits the environment to one level (the overworld), as opposed to the full nine levels. This offers the highest speed across all tested environments.

Our LLM experiments were carried out on 3 NVIDIA A100 80GB devices and took between 10 and 30 hours per 20 seeds per difficulty level for each model.

\clearpage
\subsection{Extended MARL Training}
\label{app:rl-3b-long}

\begin{figure*}[t]
  \centering
  \includegraphics[width=0.2\textwidth]{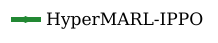}\\
  \begin{minipage}[t]{0.327\textwidth}
    \centering
    \includegraphics[width=\linewidth]{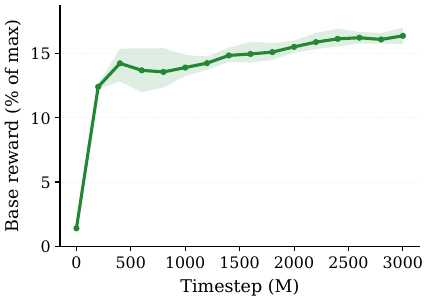}
    \subcaption{Base reward}
  \end{minipage}\hfill
  \begin{minipage}[t]{0.327\textwidth}
    \centering
    \includegraphics[width=\linewidth]{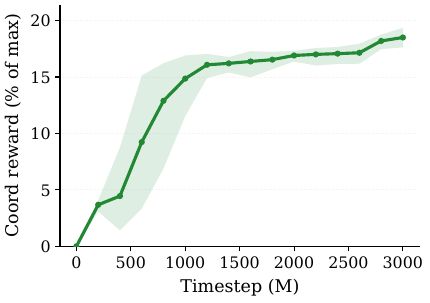}
    \subcaption{Coordination reward}
  \end{minipage}\hfill
  \begin{minipage}[t]{0.327\textwidth}
    \centering
    \includegraphics[width=\linewidth]{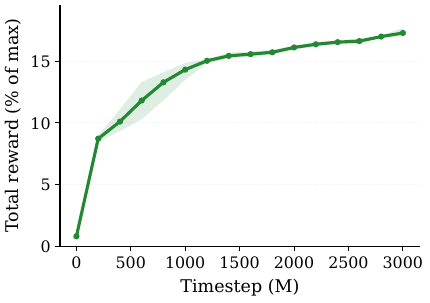}
    \subcaption{Total reward}
  \end{minipage}
  \caption{
    Extended MARL ablation on hard \textsc{alem}. HyperMARL-IPPO is trained
    for 3B environment steps across five seeds. Curves show seed means and
    shaded bands show 95\% bootstrap confidence intervals. Performance does
    not saturate: coordination and total reward continue improving late in
    training, and total reward remains well below the maximum achievable score.
  }
  \label{fig:appendix-rl-3b-long-run}
\end{figure*}

To test whether \textsc{alem}'s hard coordination setting is solved simply by
running MARL for longer, we extend the strongest RL baseline,
HyperMARL-IPPO, to 3B environment steps with five seeds. Figure~\ref{fig:appendix-rl-3b-long-run}
shows that the environment is not saturated: base reward improves early and
then plateaus, while coordination and total reward continue to climb steadily through
the end of training. Even after 3B steps, total reward remains far from the
maximum achievable score, indicating that the hard setting retains substantial
headroom for stronger learning algorithms.

\subsection{Text-based Language Interface Details}
\label{app:text_interface}

\paragraph{Prompt Structure.}
At each timestep $t$, the prompt for agent $i$ is assembled from three components: a system prompt, a history of recent observations and actions, and the current observation with instructions.

\begin{enumerate}
    \item \textbf{System prompt:}
        The system prompt is shared across all time steps for an agent and defines the agent identity and role, the team size, the objective, and the rules needed to act in the environment.
        By default it also optionally includes the full action catalogue with natural-language descriptions, game and coordination mechanics.
    \item \textbf{Observation and Action History:}
        The observation and action history is a rolling sequence of recent messages from the environment and the agent itself.
        The observation messages contain previous observations, labelled as ``Observation from $k$ step(s) ago'', and the action messages contain the action previously taken by the agent.
        When memory and communication is enabled, the observation history also includes scratchpad notes and teammate messages, each labelled with the corresponding timestep.
    \item \textbf{Current Observation and Action Space:}
        The current observation message contains textual descriptions of: achievement progress, current level, nearby terrain, items and enemies, visible coordination opportunities and requirements, teammate status, agent stats, vitals and inventory. The agent is then prompted to respond with an action.
        By default, the call to action prompt also includes current legal action affordances, instructions to think step-by-step before acting, to optionally broadcast a message, and to optionally write a private scratchpad note.
\end{enumerate}

The length of observation, communication, and memory history is independently configurable, and the system prompt can be ablated to control how much information about the game and coordination mechanics is disclosed to the model. The default setting includes all information in the system prompt and current observation with communication and scratchpad history enabled, but we ablate these components to probe their effects on LLM performance in \ref{subsec:q2_ablate_mem_coms_reason}.

\paragraph{Prompt Disclosure Levels.}
In order to reduce the amount of information given to the model, we only reveal specific gameplay information as the team reaches the relevant level.

\paragraph{Coordination Information.}
The current observation includes explicit natural language descriptions of the current coordination targets and requirements.
For synchronous targets, the prompt states whether the target works solo with a bonus or requires $N$ agents simultaneously, what the required action is, and which side of the target the acting agent currently occupies.
The location and vitals of teammates are included as well as their previous messages.

\paragraph{Action Parsing and Execution.}
The model response is instructed to respond in a structured format containing exactly one action tag, optionally preceded by a think tag for reasoning, communication tags for teammate messages, and scratchpad tags for private notes.
The required action is parsed from the response and must match one of the legal actions.
If parsing fails, the agent is re-prompted with a format error.
If the model still fails to produce a valid action after the configured retry budget is exhausted, the action defaults to \textsc{Noop}.
The final parsed action is mapped back to the underlying discrete action index before stepping the environment.

\subsection{Rewards and Achievements}\label{app:rewards_achievements}

\begin{table}[htbp]
  \centering
  \caption{Complete list of all 93 achievements available in the environment, sorted alphabetically.}
  \label{tab:achievements}
  {\small
  \begin{tabular}{@{}llll@{}}
    \toprule
    \textbf{Achievement} & \textbf{Achievement} & \textbf{Achievement} & \textbf{Achievement} \\
    \midrule
    Cast Spell & Coord Mine Coal Hard & Defeat Necromancer & Handover Complete \\
    Collect Coal & Coord Mine Coal Soft & Defeat Orc Mage & Learn Spell \\
    Collect Diamond & Coord Mine Diamond Hard & Defeat Orc Soldier & Make Arrow \\
    Collect Drink & Coord Mine Diamond Soft & Defeat Pigman & Make Diamond Armour \\
    Collect Food & Coord Mine Handover & Defeat Skeleton & Make Diamond Pickaxe \\
    Collect Iron & Coord Mine Iron Hard & Defeat Troll & Make Diamond Sword \\
    Collect Ruby & Coord Mine Iron Soft & Defeat Zombie & Make Iron Armour \\
    Collect Sapling & Coord Mine Ruby Hard & Drink Potion & Make Iron Pickaxe \\
    Collect Sapphire & Coord Mine Ruby Soft & Eat Bat & Make Iron Sword \\
    Collect Stone & Coord Mine Sapphire Hard & Eat Cow & Make Stone Pickaxe \\
    Collect Wood & Coord Mine Sapphire Soft & Eat Plant & Make Stone Sword \\
    Coord 2 Agents Hard & Coord Mine Stone Hard & Eat Snail & Make Torch \\
    Coord 2 Agents Soft & Coord Mine Stone Soft & Enchant Armour & Make Wood Pickaxe \\
    Coord 3 Agents Hard & Damage Necromancer & Enchant Sword & Make Wood Sword \\
    Coord 3 Agents Soft & Defeat Archer & Enter Dungeon & Open Chest \\
    Coord Build Beacon & Defeat Deep Thing & Enter Fire Realm & Place Furnace \\
    Coord Build Forge & Defeat Fire Elemental & Enter Gnomish Mines & Place Plant \\
    Coord Build Shelter & Defeat Frost Troll & Enter Graveyard & Place Stone \\
    Coord Diamond Armour & Defeat Gnome Archer & Enter Ice Realm & Place Table \\
    Coord Diamond Pickaxe & Defeat Gnome Warrior & Enter Sewers & Place Torch \\
    Coord Diamond Sword & Defeat Ice Elemental & Enter Troll Mines & Wake Up \\
    Coord Elite Melee Kill & Defeat Knight & Enter Vault & \\
    Coord Elite Ranged Kill & Defeat Kobold & Find Bow & \\
    Coord Large Passive Kill & Defeat Lizard & Fire Bow & \\
    \bottomrule
  \end{tabular}}
\end{table}

\clearpage
\subsection{Prompt Examples}\label{app:prompt_examples}

\definecolor{promptblue}{HTML}{1F6FEB}
\definecolor{promptbluebg}{HTML}{F3F8FF}
\definecolor{obsgreen}{HTML}{1A7F37}
\definecolor{obsgreenbg}{HTML}{F1FAF4}

\lstdefinestyle{promptstyle}{
  basicstyle=\ttfamily\scriptsize,
  breaklines=true,
  breakatwhitespace=false,
  columns=fullflexible,
  keepspaces=true,
  showstringspaces=false,
  frame=none,
  tabsize=2,
  upquote=true
}

\newtcblisting{llmpromptbox}[3][]{%
  enhanced,
  breakable,
  listing only,
  listing engine=listings,
  listing options={style=promptstyle},
  colback=promptbluebg,
  colframe=promptblue,
  coltitle=white,
  colbacktitle=promptblue,
  title={#3},
  fonttitle=\small\bfseries,
  arc=2pt,
  boxrule=0.6pt,
  left=6pt,
  right=6pt,
  top=5pt,
  bottom=5pt,
  phantom={\refstepcounter{lstlisting}\label{#2}},
  #1
}

\newtcblisting{llmobservationbox}[3][]{%
  enhanced,
  breakable,
  listing only,
  listing engine=listings,
  listing options={style=promptstyle},
  colback=obsgreenbg,
  colframe=obsgreen,
  coltitle=white,
  colbacktitle=obsgreen,
  title={#3},
  fonttitle=\small\bfseries,
  arc=2pt,
  boxrule=0.6pt,
  left=6pt,
  right=6pt,
  top=5pt,
  bottom=5pt,
  phantom={\refstepcounter{lstlisting}\label{#2}},
  #1
}

\begin{llmpromptbox}{lst:llm-prompt}{System prompt given to the LLM}
system

You are Agent 1 (forager) in a 3-agent cooperative survival game. Your goal is to gather resources, craft gear, fight monsters, and descend through 9 dungeon levels, while coordinating with teammates. You must survive -- if your health reaches zero, you die, and if all agents die the game ends. Maximize the number of achievements while staying alive.

<game_rules>

## How to play
- Each turn, choose exactly one action.
- **Movement** uses absolute directions: north, south, east, and west. Any move attempt changes your facing to that direction, even if the move is blocked and you stay in place. A move is blocked if the target tile is solid, including trees, stone, ore veins, walls, crafting stations, chests, and plants, or if it contains water, lava, a mob, or another player. If repeated move attempts do not change your position, that direction is blocked. You can also use a blocked move to turn in place, for example to face an adjacent tree.
- **Facing**: your facing direction is set by your last movement action and persists until you move again. **Do** always targets the tile in your current facing direction.
- **Do** is your main interaction: face a tile and use the **Do** action on exactly that tile to chop trees, mine ore, attack creatures, drink water, open chests, or revive a downed teammate. If the faced tile contains a downed teammate, Do revives them. If the faced tile contains a living teammate, Do targets that teammate instead and can cause friendly fire. For synchronous-style coordination, all required agents must stand next to the same target tile, face it, and act together.
- **Crafting**: stand next to (including diagonally) the required station and use the craft action; you do NOT need to face it. Diamond items always require an adjacent epic forge, not a table.
- **Placing**: face the target tile, then use the place action. Tables and furnaces need an empty non-solid tile that is not water or lava; stone can also be placed into water (costs 1 stone). Place Plant puts a sapling on the faced tile. Place Torch lights dark areas.
- **Ranged combat**: use Shoot Arrow while facing a creature (requires a bow + arrows). Bows are found in dungeon chests.
- **Elite mobs** are tougher and deal more damage; coordinating with teammates (multiple agents attacking together) makes them much easier to defeat.
- **Request/Give**: use Request [Resource] to broadcast a resource request to teammates for 10 turns; teammates can use Give to Agent X to transfer one unit of the requested resource directly -- no adjacency required, works at any distance. Give only appears as an available action when a teammate has an active Request.

## Survival stats
Food, drink, and energy deplete gradually over time -- roughly every 20-30 steps you lose 1 point of each (dexterity slows this rate). When food or drink reaches 0, your health starts dropping. When energy reaches 0, you automatically fall asleep and cannot act until energy is full. While sleeping, you take 2.5x damage from all sources. Mana does NOT decay -- it is only spent by casting spells or enchanting. Mana slowly regenerates over time (faster while sleeping).
- **Sleep**: choose this voluntarily to recover energy at 2x the passive rate. Ends automatically when energy is full.
- **Rest**: choose this to recover health gradually. Requires food, drink, and energy all > 0; ends when health is full or a stat runs out.

## Roles
Role-restricted actions succeed with reduced probability for non-specialists. Depending on the difficulty configuration, non-specialist success rates are 10%, 40%, or 70%. Specialist success rate is always 100%.

- **Forager**: collecting water, saplings, eating passive mobs (e.g. cows/bats/snails). Also has 3x base food and drink capacity.
- **Miner**: crafting pickaxes/torches, placing stone.
- **Warrior**: crafting swords and arrows. Also deals 2x melee damage, and specializes in enchanting swords and bows.
- No role restriction: Place Table, Place Furnace, Wood Sword, Iron Armour, Diamond Armour

## Coordination
Some tasks, tiles, creatures, and structures require multiple agents. When collaboration is possible the observation will include a short coordination hint. Follow the rules below to coordinate safely and efficiently.

- **Sync**: N agents must each stand on a different tile adjacent to the shared target, each facing it, and all choose the **Do** action on the same turn. Approach from different sides so every agent targets the shared tile directly. If a teammate is between you and the target, your **Do** will hit the teammate instead and can cause friendly fire.
- **Handover**: one agent starts the task with **Do** (pays the resources required), and another agent finishes it with **Do** within a small time window shown in the observation. If no agent completes it in time, the site resets and materials are refunded to the initiator. Follow the exact handover timing when specified.
- **Construction**: Construction sites (shelters, forges, beacons) may require either sync or handover. Always follow the coordination rule shown in the observation for that site.
- **Elite mobs**: stronger enemies may benefit from or require coordinated attacks. Attack from different sides, avoid standing between a teammate and the mob, and avoid blocking another agent's attack.
- **Revive**: Face a downed teammate and use **Do** to revive them.
- **Epic forge / Diamond crafting**: Diamond-tier items require multiple agents to craft simultaneously at an adjacent epic forge. All required agents must choose the crafting action for the same item on the same turn while adjacent to the forge.

## Resource chain
Trees -> wood (no tool required) -> Stone/Coal (needs wood pickaxe) -> Iron (needs stone pickaxe) -> Diamond (iron pickaxe) -> Ruby/Sapphire (diamond pickaxe)

## Crafting recipes
All recipes consume the listed materials.
Stations: Table (2 wood), Furnace (1 stone)
- Wood pickaxe/sword: table + 1 wood
- Stone pickaxe/sword: table + 1 wood + 1 stone
- Iron pickaxe/sword: table + furnace + 1 wood + 1 stone + 1 iron + 1 coal
- Iron armour: table + furnace + 3 iron + 3 coal
- Diamond pickaxe: epic forge + 1 wood + 3 diamond + enough agents crafting the same item there on the same turn
- Diamond sword: epic forge + 1 wood + 2 diamond + enough agents crafting the same item there on the same turn
- Diamond armour: epic forge + 3 diamond + enough agents crafting the same item there on the same turn
- Arrows: table + 1 wood + 1 stone (yields 2)
- Torch: table + 1 wood + 1 coal (yields 4)

Construction (at a construction site, face it and use Build action):
- Build Shelter: needs 10 wood + 5 stone. Shelters result in +50% energy regeneration while resting (all agents).
- Build Forge: needs 10 stone + 3 iron + 2 coal. Creates an epic forge, which enables diamond gear crafting.
- Build Beacon: needs 3 iron + 2 coal. Expands the lit area on this level.

## Attributes
Gain 1 XP each time you descend to a new floor. Spend XP with Level Up actions.
- **Strength**: max health = 8 + strength
- **Dexterity**: max food = 7 + 2*dexterity (+2 extra for foragers); max drink = same; max energy = 7 + 2*dexterity
- **Intelligence**: max mana = 6 + 3*intelligence; enchantment damage +5% per point above 1.

## Progression
1. Gather wood -> place a table -> craft a wood pickaxe; craft a wood sword early if combat is likely.
2. Mine stone and coal -> place a furnace -> craft iron tools and iron armour.
3. To descend: stand on the `ladder_down` tile (visible in your observation when close) and use the Descend action. The ladder only becomes usable after enough monsters on that level have been killed. Only one agent needs to use Descend/Ascend -- all teammates are teleported with them.
</game_rules>

<achievements>
## Achievements
Collect Wood
Place Table
Eat Cow
Collect Sapling
Collect Drink
Collect Food
Make Wood Pickaxe
Make Wood Sword
Place Plant
Defeat Zombie
Collect Stone
Place Stone
Eat Plant
Defeat Skeleton
Make Stone Pickaxe
Make Stone Sword
Wake Up
Place Furnace
Collect Coal
Collect Iron
Collect Diamond
Make Iron Pickaxe
Make Iron Sword
Make Arrow
Make Torch
Place Torch
Make Diamond Pickaxe
Make Diamond Sword
Make Iron Armour
Make Diamond Armour
Enter Gnomish Mines
Coord 2 Agents Soft
Coord Large Passive Kill
Coord Mine Stone Soft
Coord Mine Stone Hard
Coord Mine Coal Soft
Coord Mine Coal Hard
Coord 2 Agents Hard
Coord 3 Agents Soft
Handover Complete
Coord Mine Handover
Coord Mine Iron Soft
Coord Mine Iron Hard
Coord Mine Diamond Soft
Coord Mine Diamond Hard
Coord Build Shelter
Coord 3 Agents Hard
Coord Build Forge
Coord Build Beacon
Coord Diamond Pickaxe
Coord Diamond Sword
Coord Diamond Armour
Coord Elite Melee Kill
Coord Elite Ranged Kill
</achievements>

<output_format>
Each turn you receive an observation showing what you see, your inventory, teammates, and available actions.
Think first, then output strictly in the following format:
1. (Required) Exactly one action from the available action list:
<action>YOUR_CHOSEN_ACTION</action>
2. (Optional) Broadcast to teammates, up to 400 chars. Teammates can only act on what you tell them. Be specific (e.g. 'Dig on tree next turn', 'Ladder at 5NE', 'Need 2 wood'). Reply to teammates' requests.
<communication>YOUR_MESSAGE</communication>
3. (Optional) Private notes, up to 1000 chars -- not shared with teammates. Your context resets each turn -- this is your only memory. Don't repeat what's already in your observation; store what you'll need later. Record teammates' plans and any facts you'll need after they scroll out of view.
<scratchpad>YOUR_NOTES</scratchpad>
Important: every tag you open must be closed (e.g. <communication>...</communication>).
Token budget: 8192 tokens for your full response (including reasoning). Keep reasoning concise and stop thinking early enough to emit every required tag -- if you exhaust the budget mid-reasoning, no action is produced and your turn fails.
</output_format>
\end{llmpromptbox}

\begin{llmobservationbox}{lst:llm-prompt-obs}{Example observation given to the LLM}
The LLM receives the last 8 observation-action turns as context. Below we show the current observation from one turn.

Step: 0/10000 (10000 remaining, ends early if all agents die)
Current Observation:
Position: (x=25, y=24)
Role: forager
Location: Overworld (surface)
Achievements: 0/93 (39 unlock later)

Level info:
- Light: bright (0.80)
- Level: cleared -- you can find the ladder down tile and use Descend to go deeper.

You see:
- tree 3 steps north (x=25, y=21)
- construction_site 1 step north (x=25, y=23)
- water 5 steps east (x=30, y=24)

Facing: north.
Do target: construction_site (x=25, y=23).

Coordination:
- construction_site 1 step north (x=25, y=23): requires 2 agents to use a Build action simultaneously (fails alone). You are on the south side, adjacent (facing target).
- construction_site 1 step west (x=24, y=24): works solo but grants a bonus when 3 agents use a Build action simultaneously. You are on the east side, adjacent (facing north).
- tree 1 step south and 2 steps west (x=23, y=25): works solo but grants a bonus when 2 agents select Do simultaneously. You are on the north-east side, 3 steps away.
- tree 2 steps north and 2 steps east (x=27, y=22): requires 3 agents to select Do simultaneously (fails alone). You are on the south-west side, 4 steps away.

Teammates:
Agent 0 (warrior): 1 step west (x=24, y=24), health=9
Agent 2 (miner): 1 step south and 1 step west (x=24, y=25), health=9

Your status:
- health: 9
- food: 9
- drink: 9
- energy: 9
- mana: 9
- xp: 0

Attributes:
- dexterity: 1
- strength: 1
- intelligence: 1

You have nothing in your inventory.

Available actions:
 - Noop
 - Move West
 - Move East
 - Move North
 - Move South
 - Do
 - Sleep
 - Rest
 - Request Food
 - Request Drink
 - Request Wood
 - Request Stone
 - Request Iron
 - Request Coal
 - Request Diamond
 - Request Ruby
 - Request Sapphire

---
Response format (in this order): <action> , <communication> , <scratchpad>. Close every tag you open.
\end{llmobservationbox}

\clearpage
\section{MARL Baselines}\label{app:marl_baselines}

\begin{figure}[h]
    \centering
    \includegraphics[width=0.5\linewidth]{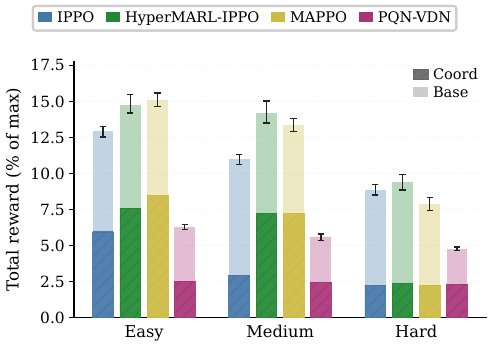}
    \caption{\textbf{MARL 100m Results.} MARL baseline performance over 5 training seeds, reported as the mean percentage of max achievable reward with 95\% CIs and decomposed into coordination (dark) and base (light) reward.}
    \label{fig:100m_results}
\end{figure}

\begin{figure*}[h]
\centering
\includegraphics[width=0.62\textwidth]{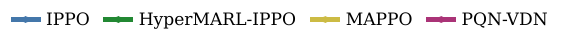}
\\[1.5em]

\begin{subfigure}[b]{0.32\textwidth}
    \centering
    \includegraphics[width=\textwidth]{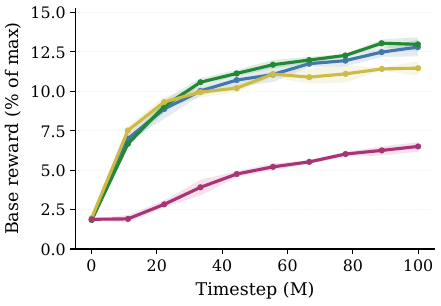}
    \caption{Base Reward -- Easy}
    \label{fig:base_easy_100m}
\end{subfigure}\hfill
\begin{subfigure}[b]{0.32\textwidth}
    \centering
    \includegraphics[width=\textwidth]{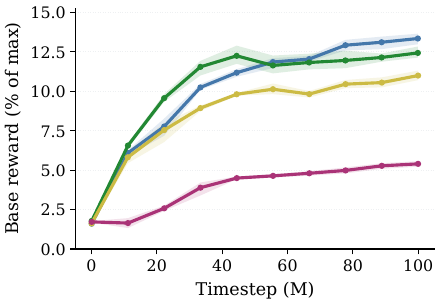}
    \caption{Base Reward -- Medium}
    \label{fig:base_med_100m}
\end{subfigure}\hfill
\begin{subfigure}[b]{0.32\textwidth}
    \centering
    \includegraphics[width=\textwidth]{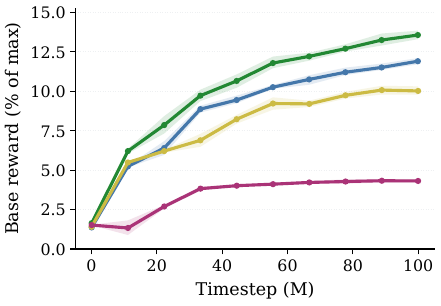}
    \caption{Base Reward -- Hard}
    \label{fig:base_hard_100m}
\end{subfigure}
\\[1em] %

\begin{subfigure}[b]{0.32\textwidth}
    \centering
    \includegraphics[width=\textwidth]{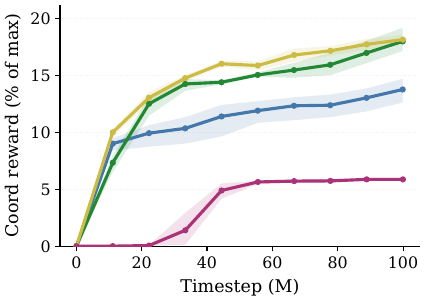}
    \caption{Coord Reward -- Easy}
    \label{fig:coord_easy_100m}
\end{subfigure}\hfill
\begin{subfigure}[b]{0.32\textwidth}
    \centering
    \includegraphics[width=\textwidth]{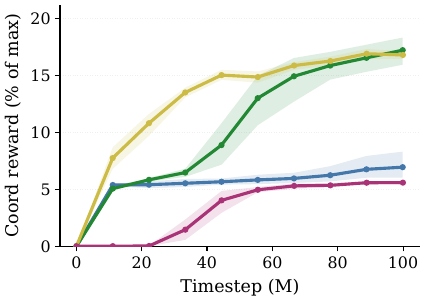}
    \caption{Coord Reward -- Medium}
    \label{fig:coord_med_100m}
\end{subfigure}\hfill
\begin{subfigure}[b]{0.32\textwidth}
    \centering
    \includegraphics[width=\textwidth]{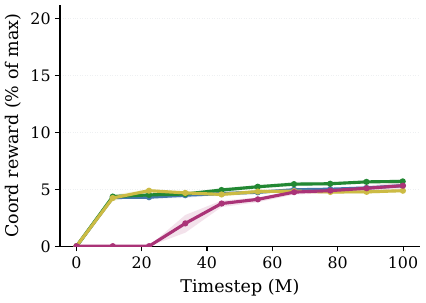}
    \caption{Coord Reward -- Hard}
    \label{fig:coord_hard_100m}
\end{subfigure}
\\[1em]

\begin{subfigure}[b]{0.32\textwidth}
    \centering
    \includegraphics[width=\textwidth]{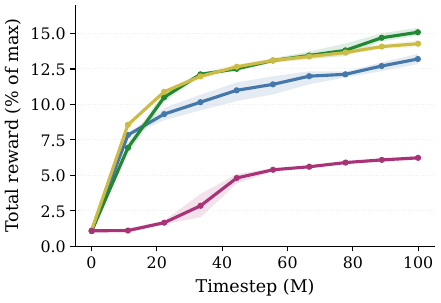}
    \caption{Total Reward -- Easy}
    \label{fig:total_easy_100m}
\end{subfigure}\hfill
\begin{subfigure}[b]{0.32\textwidth}
    \centering
    \includegraphics[width=\textwidth]{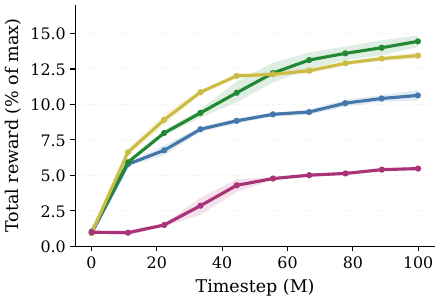}
    \caption{Total Reward -- Medium}
    \label{fig:total_med_100m}
\end{subfigure}\hfill
\begin{subfigure}[b]{0.32\textwidth}
    \centering
    \includegraphics[width=\textwidth]{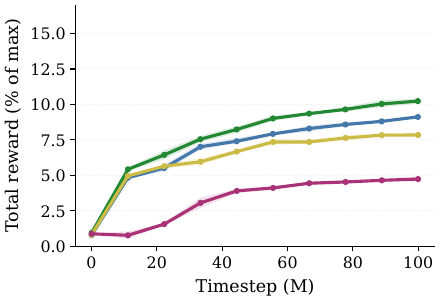}
    \caption{Total Reward -- Hard}
    \label{fig:total_hard_100m}
\end{subfigure}

\caption{
\textbf{Learning curves during training for MARL baselines across coordination difficulty, trained for 100m steps.} 
Individual panels (a)-(i) break down performance by environment difficulty (Easy, Medium, Hard) and reward type (Base, Coordination, Total). Curves show the mean across 5 independent training seeds; shaded regions are 95\% bootstrap confidence intervals. Base and coordination scores are normalised by their category-specific achievable maxima, while total reward is normalised by the full achievable reward.
}
\label{fig:rl_learning_curves_100m}
\end{figure*}

\begin{figure*}[h]
\centering
\includegraphics[width=0.62\textwidth]{images/rl_results_curves_legend.pdf}
\\[1.5em]

\begin{subfigure}[b]{0.32\textwidth}
    \centering
    \includegraphics[width=\textwidth]{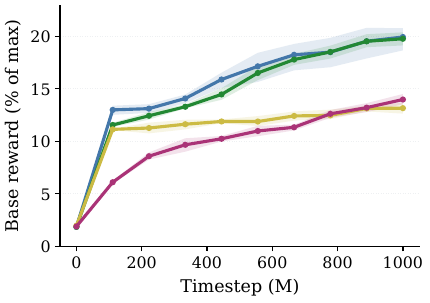}
    \caption{Base Reward -- Easy}
    \label{fig:base_easy}
\end{subfigure}\hfill
\begin{subfigure}[b]{0.32\textwidth}
    \centering
    \includegraphics[width=\textwidth]{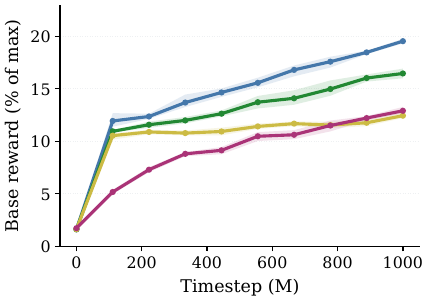}
    \caption{Base Reward -- Medium}
    \label{fig:base_med}
\end{subfigure}\hfill
\begin{subfigure}[b]{0.32\textwidth}
    \centering
    \includegraphics[width=\textwidth]{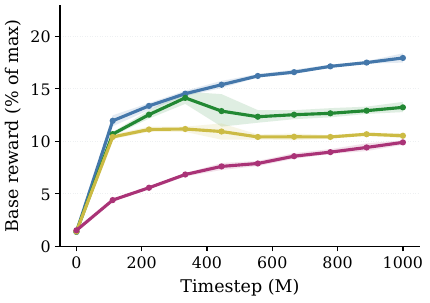}
    \caption{Base Reward -- Hard}
    \label{fig:base_hard}
\end{subfigure}
\\[1em] %

\begin{subfigure}[b]{0.32\textwidth}
    \centering
    \includegraphics[width=\textwidth]{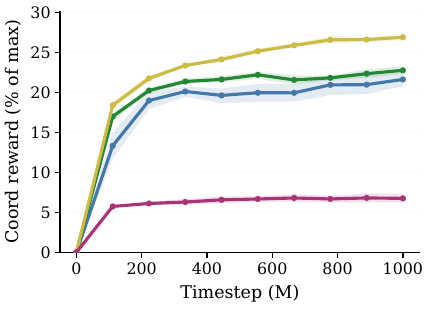}
    \caption{Coord Reward -- Easy}
    \label{fig:coord_easy}
\end{subfigure}\hfill
\begin{subfigure}[b]{0.32\textwidth}
    \centering
    \includegraphics[width=\textwidth]{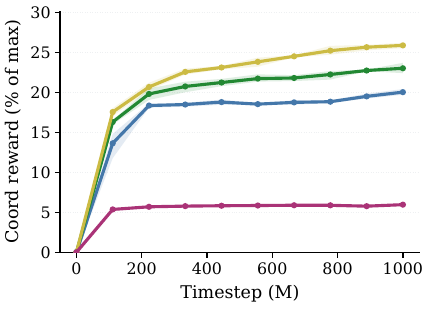}
    \caption{Coord Reward -- Medium}
    \label{fig:coord_med}
\end{subfigure}\hfill
\begin{subfigure}[b]{0.32\textwidth}
    \centering
    \includegraphics[width=\textwidth]{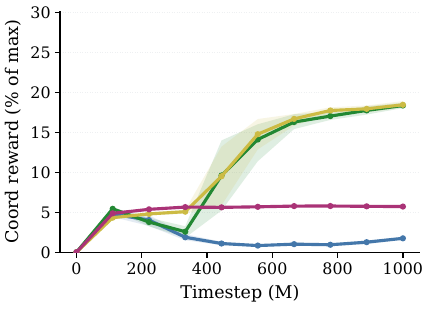}
    \caption{Coord Reward -- Hard}
    \label{fig:coord_hard}
\end{subfigure}
\\[1em]

\begin{subfigure}[b]{0.32\textwidth}
    \centering
    \includegraphics[width=\textwidth]{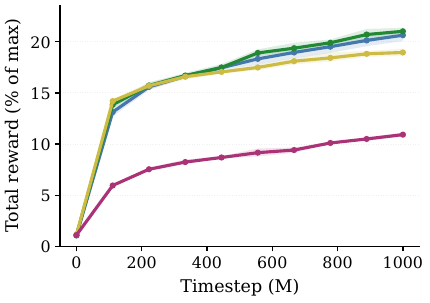}
    \caption{Total Reward -- Easy}
    \label{fig:total_easy}
\end{subfigure}\hfill
\begin{subfigure}[b]{0.32\textwidth}
    \centering
    \includegraphics[width=\textwidth]{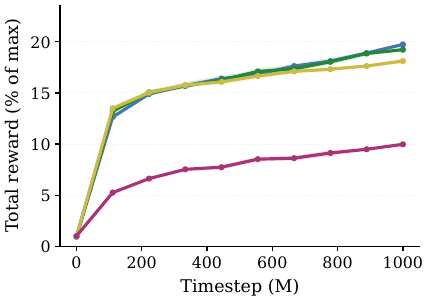}
    \caption{Total Reward -- Medium}
    \label{fig:total_med}
\end{subfigure}\hfill
\begin{subfigure}[b]{0.32\textwidth}
    \centering
    \includegraphics[width=\textwidth]{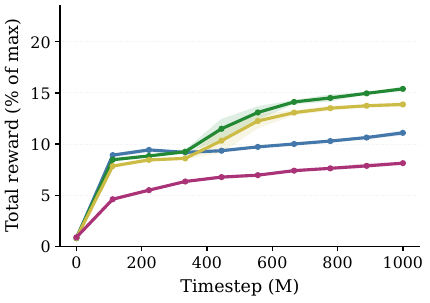}
    \caption{Total Reward -- Hard}
    \label{fig:total_hard}
\end{subfigure}

\caption{
\textbf{Learning curves during training for MARL baselines across coordination difficulty, trained for one billion steps.} 
Individual panels (a)-(i) break down performance by environment difficulty (Easy, Medium, Hard) and reward type (Base, Coordination, Total). Curves show the mean across 5 independent training seeds; shaded regions are 95\% bootstrap confidence intervals. Base and coordination scores are normalised by their category-specific achievable maxima, while total reward is normalised by the full achievable reward.
}
\label{fig:rl_learning_curves}
\end{figure*}

\let\rlcurvepanel\relax
\let\rlcurverowlabel\relax

\clearpage
\section{Additional LLM Experiments and Ablations}

\subsection{Environment calibration: multi-agent structure adds difficulty beyond the base game.}\label{app:env_calibration}

\begin{figure}
  \centering
  \includegraphics[width=0.5\linewidth]{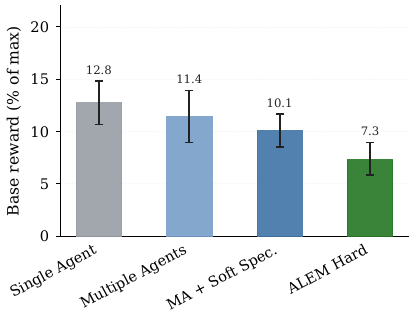}
  \caption{\textbf{Environment calibration.}
Base reward for Gemma-4-31B-it across settings sharing the same underlying world. Cooperative achievements are excluded so all settings are directly comparable. Error bars show $95\%$ bootstrap CIs.}
  \label{fig:llm-env-ablations-no-easy}
\end{figure}

Before evaluating full \textsc{alem}, we isolate how much difficulty is introduced by the multi-agent structure beyond the underlying open-ended environment dynamics. We evaluate Gemma-4-31B-it in four settings that share the same base world: a single-agent setting, a multi-agent setting without role pressure or coordination tasks, a multi-agent setting with soft specialisation, and full \textsc{alem} at $\alpha=0.9$ (\textsc{Hard}).

Fig.~\ref{fig:llm-env-ablations-no-easy} shows a steady decline in Base reward, the only reward category directly comparable across all settings: $12.8\%$ in the single-agent setting, $11.4\%$ with multiple agents, $10.1\%$ with soft specialisation, and $7.3\%$ in full \textsc{alem}. The key comparison is the single-agent setting versus full \textsc{alem}: a $43\%$ relative drop on the same underlying world. This calibration suggests that \textsc{alem}'s difficulty is not reducible to the base game. Adding agents changes the dynamics as each agent's transition depends on the actions of the others, soft specialisation adds role-allocation pressure, and full \textsc{alem} adds explicit coordination requirements. We do not over-interpret the intermediate steps, since adjacent confidence intervals overlap and this calibration uses one open-weight model. Instead, the result serves as a sanity check that full \textsc{alem} introduces multi-agent failure modes absent from single-agent progression.

\subsection{How much do game rules and coordination cues help?}
\label{app:prompt-disclosure}

Alem varies layouts and coordination assignments across episodes, but it inherits much of its base progression from Craftax. A model familiar with Craftax may therefore already know, for example, which tool it needs to mine a particular resource. Our default prompt explains this progression (App.~C.7). To examine how much agents gain from the information we provide, we evaluate Gemma-4-31B-it on Hard under three disclosure settings.

In \texttt{general}, agents receive a list of actions with descriptions and the names of achievements. In \texttt{specific}, they receive a structured guide to survival, roles, crafting, and progression. The guide also explains how to request and give resources. The default setting, \texttt{specific\_collaborative}, includes this guide and explains how to carry out our joint actions, requiring coordination. It also adds cues to each agent's observations during play. For example, an agent can see how many teammates must act on a visible target at the same time or how many steps remain to complete a handover. Agents can communicate in all three settings.

\begin{table}[t]
  \centering
  \caption{\textbf{Effect of game rules and coordination cues on Gemma-4-31B-it on Hard.}
    Brackets give confidence intervals.}
  \label{tab:prompt-disclosure}
  \small
  \setlength{\tabcolsep}{4pt}
  \begin{tabular}{@{}lccc@{}}
    \toprule
    Disclosure setting & Base\% & Coord.\% & Total\% \\
    \midrule
    \texttt{general}
      & 7.1 [5.9, 8.6] & 1.8 [0.6, 3.0] & 4.8 [3.8, 6.0] \\
    \texttt{specific}
      & 10.1 [8.1, 12.3] & 0.7 [0.0, 1.7] & 6.1 [5.0, 7.3] \\
    \texttt{specific\_collaborative} (default)
      & 7.3 [5.9, 8.9]  & 8.8 [6.9, 10.9]  &  8.0 [6.7, 9.1] \\ 
    \bottomrule
  \end{tabular}
\end{table}

With the structured game guide, mean Base\% rises from 7.1 to 10.1, while Coord.\% remains low under both settings (Table~\ref{tab:prompt-disclosure}). Providing the coordination rules and in-game cues together raises Coord.\% from 0.7 to 8.8, a gain of 8.1 percentage points, while Base\% falls from 10.1 to 7.3.

Prior familiarity with Craftax may contribute to base-task performance in all three settings. What distinguishes the two settings with the game guide is the information agents receive about working together. Even when Gemma has the progression rules, it coordinates better when it also knows how joint actions work and when a particular target requires coordination. 

\clearpage
\section{Qualitative Analysis}

Our qualitative analysis of the LLM agents' memory, reasoning and communication can be found under sections \ref{app:mem_qual}, \ref{app:reason_qual} and \ref{app:comm_qual} respectively. App.~\ref{app:functional-traces} complements these lexical analyses by annotating what each message, scratchpad entry, and reasoning trace does, using an LLM annotator.

\section{Memory entry usage: detailed analysis}
\label{app:memory_scratchpad}

\subsection{Setup}
\label{app:mem_setup}

We compare how Gemini~3.1~Pro and Gemma use the persistent memory field across episodes. This is the same trajectory set used in the communication analysis: $10$ Gemini episodes ($34{,}080$ agent-step records) and $20$ Gemma episodes ($24{,}747$ records). Each record contains the persisted \texttt{scratchpad} string, which we refer to as the agent's \emph{memory entry}, together with the parsed action and local observation. Parse failures are negligible, under $0.5\%$ for both models.

\paragraph{Lexical detectors.}
Structural features, such as character counts and line counts, are computed directly. The remaining features in Table~\ref{tab:mem_clean} are simple regular-expression detectors over the raw memory-entry text. These detectors are intended as interpretable proxies rather than full discourse parsers. We define them as follows:
\begin{itemize}\itemsep2pt
  \item \textbf{Multi-line memory entry ($\geq 2$ lines).} The memory entry contains at least one newline.
  \item \textbf{Turn-indexed plan.} The memory entry has at least two lines whose first non-whitespace tokens match
        ``\texttt{T}\,$N$\,\texttt{:}'' or ``\texttt{Turn}\,$N$\,\texttt{:}''
        for an integer $N$. This captures explicit turn-by-turn plans,
        such as ``\emph{T9: Do (completes handover \dots)}''.
  \item \textbf{Compact state summary.} The memory entry contains at least two
        ``\texttt{Field: value}'' fields drawn from a fixed vocabulary of status labels
        (\emph{Pos, Position, HP, Health, Inv, Inventory, Energy, Hunger, Thirst, A1, A2, Agent}).
        This captures compact one-line entries that mainly record position, resources, health, or teammate state.
  \item \textbf{Plan / future-tense word.} \emph{will, going to,
        gonna, plan, planning, next, then, after, once, when}, plus
        future-tense modal forms (\emph{i'll, we'll, i will, we
        will}). ``Plan terms / entry'' counts every match, while
        ``\% of entries'' fires if at least one match is present.
  \item \textbf{Past-tense / memory word.} \emph{tried, failed, was,
        were, did, didn't, couldn't, earlier, before, already, just,
        noticed, saw}.
  \item \textbf{Coordinate reference.} A token matching
        \texttt{\textbackslash(?\textbackslash s*-?\textbackslash d+\textbackslash s*,\textbackslash s*-?\textbackslash d+\textbackslash s*\textbackslash)?},
        i.e.\ a pair of integers, possibly wrapped in parentheses,
        such as ``$25,23$'' or ``$(0,-1)$''.
  \item \textbf{Teammate reference.} An ``\texttt{Agent}\,$X$'' or
        ``\texttt{A}$X$'' token for $X \in \{0,1,2\}$, optionally
        followed by a comma or a status colon.
  \item \textbf{Step-to-step turnover.} For every consecutive pair of
        controlled memory entries, we compute (a) the SequenceMatcher
        character similarity ratio and (b) Jaccard similarity over lower-cased
        tokens. Higher values indicate more overlap between consecutive entries.
\end{itemize}
The absolute values depend on the exact detector definitions, but the qualitative differences below are stable across seeds and robust to reasonable variants of the regexes.

\subsection{Quantitative results}
\label{app:mem_quant}

Table~\ref{tab:mem_clean} shows that the two models use the memory field in different ways. Gemini tends to write longer, multi-line entries with explicit future plans. Gemma writes shorter entries that more often summarise the current state.

\begin{table}[t]
  \centering
  \small
  \captionsetup{position=top,skip=6pt}
  \caption{Memory entry usage on controlled steps (mean~$\pm$~std across runs;
  Gemini $n=10$, Gemma $n=20$). Top: structural style. Middle: content.
  Bottom: step-to-step turnover from one controlled memory entry to the next.}
  \label{tab:mem_clean}
  \begin{tabular}{lrr}
    \toprule
    & Gemini~3.1~Pro & Gemma \\
    \midrule
    \multicolumn{3}{l}{\emph{Structural style}} \\
    entry characters                      & $206.7 \pm 16.4$  & $153.0 \pm 9.6$  \\
    entry lines                           & $3.86 \pm 0.72$   & $1.08 \pm 0.06$  \\
    multi-line entry ($\geq 2$ lines), \% of entries & $71.5 \pm 7.2$ & $2.0 \pm 1.3$ \\
    turn-indexed plan ($\geq 2$ lines), \% of entries & $12.4 \pm 9.4$ & $0.2 \pm 0.2$ \\
    compact state summary ($\geq 2$ status fields), \% & $4.7 \pm 1.8$ & $16.1 \pm 4.3$ \\
    \midrule
    \multicolumn{3}{l}{\emph{Content}} \\
    plan / future-tense word, \% of entries     & $75.3 \pm 2.5$ & $36.6 \pm 5.2$ \\
    past-tense / memory word, \% of entries     & $9.0 \pm 1.1$  & $6.7 \pm 1.4$ \\
    coordinate reference, \% of entries          & $90.7 \pm 3.1$ & $98.8 \pm 1.1$ \\
    teammate reference, \% of entries            & $68.0 \pm 7.0$ & $70.8 \pm 7.1$ \\
    coord references / entry                    & $3.52 \pm 0.71$ & $4.01 \pm 0.49$ \\
    plan terms / entry                          & $1.60 \pm 0.11$ & $0.52 \pm 0.08$ \\
    \midrule
    \multicolumn{3}{l}{\emph{Step-to-step turnover (controlled $\to$ next controlled entry)}} \\
    char similarity (sequence ratio) & $0.44 \pm 0.02$ & $0.58 \pm 0.04$ \\
    Jaccard on tokens                & $0.37 \pm 0.01$ & $0.47 \pm 0.04$ \\
    \bottomrule
  \end{tabular}
\end{table}

Three patterns stand out.

\paragraph{Gemini writes plans; Gemma writes state summaries.}
The structural differences are large. Gemini writes $3.86$ lines per memory entry on average, compared with $1.08$ for Gemma. Similarly, $71.5\%$ of Gemini entries have at least two lines, compared with $2.0\%$ for Gemma. Gemini also uses explicit turn-indexed plans in $12.4\%$ of entries, while Gemma almost never uses this format ($0.2\%$). Reading the entries gives the same picture: Gemini often lays out the next few turns line by line, whereas Gemma usually writes a compact one-line summary of position, recent obstacles, next movement, and teammate state. 

\paragraph{Neither model uses the pad as memory.}
Both models use the memory entry more for immediate planning or state tracking than for recording past failures or lessons. Past-tense or retrospective tokens, such as \emph{tried}, \emph{failed}, \emph{earlier}, \emph{before}, \emph{already}, and \emph{noticed}, appear in only $9.0\%$ of Gemini entries and $6.7\%$ of Gemma entries. This is notable because the memory entry is one of the few persistent within-episode channels available to the model. It could be used to record facts such as a failed craft attempt, a previous handover, or an already-tested route, but the models rarely use it in this way. This could also be due to the history (last 8 observations and actions) agents receive in each observation. 

\paragraph{Both models actively rewrite their memory entries.}
The memory entries are not simply copied forward unchanged. Consecutive controlled entries have token Jaccard similarity of $0.37$ for Gemini and $0.47$ for Gemma, with character-level similarity of $0.44$ and $0.58$, respectively. This indicates substantial rewriting from one controlled step to the next. Gemini rewrites more than Gemma, consistent with its planning-oriented usage: it often recomputes the next few actions rather than incrementally updating a fixed state summary.

\subsection{Qualitative examples}
\label{app:mem_qual}

\paragraph{Example selection.}
The examples below are illustrative rather than randomly sampled. We hold the role fixed across models by showing agent~$0$, the warrior, so that role-gated penalties and action affordances are comparable. We only show controlled steps, since forced \texttt{Noop} steps copy the previous memory entry. The Gemini example is a representative turn-indexed plan from run~02, step~9. The two Gemma examples, from run~00 step~0 and run~06 step~473, illustrate that the compact state-summary format appears both early and later in the episode. We did not search for unusually coherent or unusually successful entries. In our sample, we found no Gemma entries with the turn-indexed planning pattern.

\definecolor{geminiblue}{HTML}{1F6FEB}
\definecolor{geminitint}{HTML}{E8F1FD}
\definecolor{gemmagreen}{HTML}{1F8B4C}
\definecolor{gemmatint}{HTML}{E8F5EE}

\begin{tcolorbox}[
  enhanced,
  colback=geminitint,
  colframe=geminiblue,
  arc=2pt,
  boxrule=0.6pt,
  left=8pt,right=8pt,top=6pt,bottom=6pt,
  title={\textbf{Gemini~3.1~Pro}: run 02, agent 0
         (warrior), step 9, action: \texttt{Do}},
  fonttitle=\small\bfseries,
  coltitle=white,
  colbacktitle=geminiblue,
  attach boxed title to top left={yshift=-2mm,xshift=4mm},
  boxed title style={arc=2pt,boxrule=0pt}
]
\footnotesize
T9: Do (completes handover at 26,23)\\
T10: Move West (towards 21,23)\\
T11: Give wood to Agent 1 (since A1 will Request Wood on T10)\\
T12+: Move West to table, then Make Wood Pickaxe.\\
Since I'm a warrior, pickaxe crafting has a chance to fail (40\% or whatever the config is), so I'll keep trying if it does.
\end{tcolorbox}

\vspace{2pt}

\begin{tcolorbox}[
  enhanced,
  colback=gemmatint,
  colframe=gemmagreen,
  arc=2pt,
  boxrule=0.6pt,
  left=8pt,right=8pt,top=6pt,bottom=6pt,
  title={\textbf{Gemma~4}: agent 0 (warrior), runs 00 \& 06},
  fonttitle=\small\bfseries,
  coltitle=white,
  colbacktitle=gemmagreen,
  attach boxed title to top left={yshift=-2mm,xshift=4mm},
  boxed title style={arc=2pt,boxrule=0pt}
]
\footnotesize
\textbf{Run 00, step 0} \;\;action: \texttt{Move North}\\[2pt]
Starting resource collection. Agent 1 (Forager) is best for wood (25, 23). Agent 2 (Miner) is best for stone (20, 23). I will move to (24, 23) to assist with wood gathering.

\medskip
\textbf{Run 06, step 473} \;\;action: \texttt{Move North}\\[2pt]
Current (0, 19). Hit stone at (0, 20) and edge at x=-1. Moving North to (0, 18) to reposition and start searching East. A2 is searching North. A1 still sleeping.
\end{tcolorbox}

The examples illustrate the quantitative differences in Table~\ref{tab:mem_clean}. The Gemini entry is an explicit turn-indexed plan: it records the current coordinated action, predicts a teammate request, schedules a resource transfer, and includes a contingency for the warrior crafting penalty. The Gemma entries instead compress the current situation into a short state summary: position, recent obstacle, next movement, and teammate status. These examples reflect the broader pattern: Gemini tends to use memory entries to maintain near-future plans, while Gemma tends to use them to restate the current state.

\section{Reasoning ablation: detailed analysis}
\label{app:reasoning}

\subsection{Setup}
\label{app:reason_setup}

The reasoning ablation differs across models because of API constraints. Gemini~3.1~Pro does not allow reasoning to be disabled, so we compare the high-reasoning baseline to a low-reasoning setting, denoted \emph{Gemini-Low}. Gemma allows the reasoning trace to be disabled entirely, denoted \emph{Gemma-No-Reasoning}. Gemini-Low still records $879 \pm 44$ characters and $198 \pm 20$ tokens of \texttt{raw\_reasoning} per controlled step on average, so it should be interpreted as a partial ablation.

We collected $10$ Gemini-Low episodes and $18$ Gemma-No-Reasoning episodes. We reuse the same metrics as in App.~\ref{app:memory_scratchpad} for memory entries on controlled steps, and App.~\ref{app:comm_style} for messages across all records. The lexical detectors are exactly those defined in the corresponding appendices: turn-indexed plans, plan-word and past-word lexicons, coordinate references, imperative verbs, and related markers. The baseline columns in Table~\ref{tab:reason_clean} are copied from Tables~\ref{tab:mem_clean} and~\ref{tab:comm_clean}.

\subsection{Quantitative results}
\label{app:reason_quant}

\begin{table}[t]
  \centering
  \small
  \captionsetup{position=top,skip=6pt}
  \caption{Effect of reducing reasoning on memory-entry and message style
  (mean~$\pm$~std across runs). Baseline columns are from
  Tables~\ref{tab:mem_clean}--\ref{tab:comm_clean}; ablated columns are
  the new runs ($n=10$ Gemini-Low, $n=18$ Gemma-No-Reasoning). The
  memory-entry block uses controlled steps only; the message block uses
  all records.}
  \label{tab:reason_clean}
  \begin{tabular}{lrrrr}
    \toprule
    & \multicolumn{2}{c}{Gemini~3.1~Pro} & \multicolumn{2}{c}{Gemma} \\
    \cmidrule(lr){2-3}\cmidrule(lr){4-5}
    & Baseline & Low-Reason & Baseline & No-Reason \\
    \midrule
    \multicolumn{5}{l}{\emph{Memory entries (controlled steps only)}} \\
    entry characters                          & $206.7 \pm 16.4$ & $\mathbf{69.2 \pm 1.9}$  & $153.0 \pm 9.6$  & $\mathbf{228.5 \pm 22.4}$ \\
    entry lines                               & $3.86 \pm 0.72$  & $\mathbf{1.04 \pm 0.01}$ & $1.08 \pm 0.06$  & $1.10 \pm 0.12$  \\
    multi-line entry ($\geq 2$ lines), \%     & $71.5 \pm 7.2$   & $\mathbf{2.1 \pm 0.5}$   & $2.0 \pm 1.3$    & $1.9 \pm 2.3$    \\
    turn-indexed plan, \%                     & $12.4 \pm 9.4$   & $\mathbf{0.1 \pm 0.0}$   & $0.2 \pm 0.2$    & $0.2 \pm 0.1$    \\
    plan / future word, \%                    & $75.3 \pm 2.5$   & $\mathbf{35.5 \pm 2.5}$  & $36.6 \pm 5.2$   & $\mathbf{73.9 \pm 5.7}$ \\
    past / memory word, \%                    & $9.0 \pm 1.1$    & $2.9 \pm 0.5$            & $6.7 \pm 1.4$    & $\mathbf{18.3 \pm 2.8}$ \\
    coordinate ref, \%                        & $90.7 \pm 3.1$   & $\mathbf{45.5 \pm 5.4}$  & $98.8 \pm 1.1$   & $96.1 \pm 2.5$   \\
    teammate ref, \%                          & $68.0 \pm 7.0$   & $\mathbf{27.8 \pm 6.4}$  & $70.8 \pm 7.1$   & $77.2 \pm 7.2$   \\
    plan terms / entry                        & $1.60 \pm 0.11$  & $\mathbf{0.49 \pm 0.04}$ & $0.52 \pm 0.08$  & $\mathbf{1.28 \pm 0.16}$ \\
    \midrule
    \multicolumn{5}{l}{\emph{Messages (all records)}} \\
    msg characters                            & $132.4 \pm 16.0$ & $\mathbf{52.9 \pm 3.4}$  & $104.1 \pm 7.9$  & $98.3 \pm 13.1$  \\
    sentences / msg                           & $3.22 \pm 0.40$  & $\mathbf{1.55 \pm 0.09}$ & $2.60 \pm 0.16$  & $2.28 \pm 0.18$  \\
    multi-sentence ($\geq 2$), \%             & $93.7 \pm 2.0$   & $\mathbf{47.3 \pm 6.7}$  & $87.9 \pm 4.5$   & $88.7 \pm 4.9$   \\
    imperative verb, \%                       & $73.7 \pm 6.6$   & $\mathbf{38.7 \pm 5.4}$  & $52.2 \pm 7.3$   & $\mathbf{62.2 \pm 10.8}$ \\
    plan / future word, \%                    & $59.1 \pm 5.3$   & $\mathbf{22.1 \pm 3.4}$  & $27.0 \pm 3.8$   & $\mathbf{37.6 \pm 7.2}$ \\
    contains coordinate, \%                   & $56.5 \pm 10.0$  & $\mathbf{25.6 \pm 6.8}$  & $66.8 \pm 10.7$  & $67.6 \pm 8.7$   \\
    politeness (\emph{please/thanks}), \%     & $14.0 \pm 3.1$   & $10.4 \pm 3.0$           & $20.9 \pm 4.8$   & $\mathbf{27.1 \pm 8.8}$ \\
    self-refs / msg                           & $1.65 \pm 0.18$  & $\mathbf{0.67 \pm 0.12}$ & $0.92 \pm 0.17$  & $1.16 \pm 0.22$  \\
    \bottomrule
  \end{tabular}
\end{table}

\paragraph{Gemini loses much of its planner-like behaviour.}
Under low reasoning, the memory-entry features that characterised Gemini in the baseline largely disappear. Entry length falls from $207$ to $69$ characters, multi-line entries from $71.5\%$ to $2.1\%$, and turn-indexed plans from $12.4\%$ to $0.1\%$. The message channel changes in the same direction: messages fall from $132$ to $53$ characters, from $3.22$ to $1.55$ sentences, from $73.7\%$ to $38.7\%$ imperative verb usage, and from $56.5\%$ to $25.6\%$ references to coordinates. In short, the low-reasoning setting appears to remove much of the outward planning and directive communication observed in the baseline.

\paragraph{Gemma shifts more deliberation into memory entries.}
Gemma shows a different pattern. With reasoning disabled, memory entries become longer rather than shorter ($153 \to 228$ characters), often approaching the harness's $\sim\!1000$-character cap. Their content also becomes more plan-like: plan-word rate rises from $36.6\%$ to $73.9\%$, plan terms per entry from $0.52$ to $1.28$, and past-tense or reflective tokens from $6.7\%$ to $18.3\%$. The entries still average roughly one line ($1.10$ vs.\ $1.08$), so the change is not toward Gemini-style turn-indexed plans. Instead, Gemma writes longer single-line deliberations inside the memory field. The messages shift modestly in the same direction, with higher imperative usage ($52.2\% \to 62.2\%$) and more plan words ($27.0\% \to 37.6\%$).

\paragraph{This compensation is incomplete.}
Although Gemma appears to move some deliberation into the memory and message fields, performance still falls in the main ablation results: Coord.\% drops by $3.3$ points and Base\% by $2.7$ points. Gemini, which does not show the same expansion of memory entries, loses more Coord.\% ($-7.8$ points) and also loses Base\% ($-3.5$ points). Thus, the behavioural traces match the quantitative ablation: reducing reasoning changes how models use the remaining harness channels, but those changes do not recover the lost performance.

\subsection{Qualitative examples}
\label{app:reason_qual}

\paragraph{Example selection.}
The examples below are illustrative rather than randomly sampled. For Gemini-Low, we show a typical one-line memory entry and a rare surviving turn-indexed plan. For Gemma-No-Reasoning, we show a long single-line deliberation and a reflective debugging entry. We hold the agent role fixed where possible, using agent~$0$ as the warrior in three of the four excerpts. We use agent~$2$ for the long Gemma deliberation example because that excerpt better illustrates the near-cap memory-entry behaviour. Within these constraints, we selected matching examples from fixed runs rather than searching across runs for the most extreme cases.

\definecolor{geminiblue}{HTML}{1F6FEB}
\definecolor{geminitint}{HTML}{E8F1FD}
\definecolor{gemmagreen}{HTML}{1F8B4C}
\definecolor{gemmatint}{HTML}{E8F5EE}

\begin{tcolorbox}[
  enhanced,
  colback=geminitint,
  colframe=geminiblue,
  arc=2pt,
  boxrule=0.6pt,
  left=8pt,right=8pt,top=6pt,bottom=6pt,
  title={\textbf{Gemini-Low}: run 00, agent 0
         (warrior)},
  fonttitle=\small\bfseries,
  coltitle=white,
  colbacktitle=geminiblue,
  attach boxed title to top left={yshift=-2mm,xshift=4mm},
  boxed title style={arc=2pt,boxrule=0pt}
]
\footnotesize
\textbf{Step 1} \;\;action: \texttt{Move East} \;\;\emph{(typical one-line memory entry)}\\[2pt]
\textbf{Memory entry:} \emph{``Moving East to reach tree at 27,23 to
coordinate with agents.''}\\
\textbf{Message:} \emph{``Moving to tree at 27,23 for 3-agent
gather.''}

\medskip
\textbf{Step 60} \;\;action: \texttt{Move South} \;\;\emph{(rare surviving turn-indexed plan)}\\[2pt]
\emph{Pathing to 29,28 to attack cow at 29,27.\\
Current: 27,26\\
T60: Move South (to 27,27)\\
T61: Move South (to 27,28)\\
T62: Move East (to 28,28)\\
T63: Move East (to 29,28)\\
T64: Do}
\end{tcolorbox}

\vspace{2pt}

\begin{tcolorbox}[
  enhanced,
  colback=gemmatint,
  colframe=gemmagreen,
  arc=2pt,
  boxrule=0.6pt,
  left=8pt,right=8pt,top=6pt,bottom=6pt,
  title={\textbf{Gemma-No-Reasoning}: agent 0
         (warrior) \& agent 2},
  fonttitle=\small\bfseries,
  coltitle=white,
  colbacktitle=gemmagreen,
  attach boxed title to top left={yshift=-2mm,xshift=4mm},
  boxed title style={arc=2pt,boxrule=0pt}
]
\footnotesize
\textbf{Run 00, agent 2, step 9} \;\;action: \texttt{Move East}
\;\;\emph{(long single-line deliberation)}\\[2pt]
\emph{Current position (25, 25). Target tree (27, 23).\\
Path: (25, 25) -> (26, 25) -> (26, 24) [facing North] -> Do.\\
Agent 0 is at (26, 23) facing North/West? (Wait, Agent 0 is at 26,23,
tree is at 27,23, so Agent 0 is West of tree and facing East). Agent 1
is at (27, 24) facing North. I will move to (26, 24) and face North
(wait, if I'm at 26, 24 and tree is 27, 23, I should face East).
Correct target: Tree (27, 23). My current pos: (25, 25). Action: Move
East to (26, 25). Next: Move North to (26, 24), then Move East to face
(27, 23)\dots Actually, if I move to (26, 23) I'd be on top of Agent 0.
I'll go (26, 25) -> (26, 24) -> face East to target (27, 24)? No, tree
is at (27, 23). [\dots]}

\medskip
\textbf{Run 02, agent 0, step 211} \;\;action: \texttt{Move North}
\;\;\emph{(reflective debugging)}\\[2pt]
\emph{[\dots] Wait, I just tried Move West several times and stayed at
(25, 22). Let me check why. The construction site is at (24, 22).
Construction sites are solid. I must go around. [\dots] Actually,
looking at the observation, my last 5 moves were Move West and I
stayed at (25, 22). This confirms (24, 22) is blocked.}
\end{tcolorbox}

The examples reflect the quantitative pattern in Table~\ref{tab:reason_clean}. Gemini-Low usually writes short memory entries and short messages; the turn-indexed format remains possible, but is rare. Gemma-No-Reasoning, by contrast, keeps the one-line memory format but fills it with path enumeration, corrections, and retrospective debugging. This suggests that disabling the reasoning trace changes not only performance, but also where each model places its intermediate deliberation.

\section{Inter-agent communication style}
\label{app:comm_style}

\subsection{Setup}
\label{app:comm_setup}

We analyse the per-step \texttt{message\_sent} field, which is the open communication channel available to agents during an episode. We use the same trajectories as in the scratchpad analysis: $10$ Gemini~3.1~Pro episodes and $20$ Gemma~4 episodes, resulting in $34{,}080$ and $24{,}747$ agent-step records respectively.

\paragraph{Lexical detectors.}
To characterise how agents use the channel, we compute a small set of lexical features from the raw message text. These are deliberately simple probes rather than discourse parsers. Their purpose is to capture stable differences in communication style, such as whether a message gives an instruction, names a teammate, asks a question, or records a plan. Table~\ref{tab:comm_clean} reports the resulting rates. For completeness, the detectors are defined as follows:
\begin{itemize}\itemsep2pt
  \item \textbf{Imperative verb.} A sentence-initial base-form verb
        from a fixed action lexicon
        (\emph{move, go, do, place, make, drink, eat, attack, give,
        request, mine, chop, gather, face, wait, stay, build, kill,
        head, bring, get, take, drop, equip, craft, sleep, rest, hold,
        return, follow, stop, switch, prioritize, focus, defend}),
        optionally preceded by an addressee (``\emph{Agent X,}''). We
        also count the same verbs when they immediately follow a
        teammate addressee elsewhere in the message. ``Imperatives per
        message'' counts all matches; ``contains an imperative'' is
        binary.
  \item \textbf{Broadcast phrasing.} A leading first-person plural
        cohortative: \emph{let's, lets, let us, we should, we will, we
        need to, we must, we'll}.
  \item \textbf{Addresses specific teammate.} A token of the form
        ``\emph{Agent~$X$,}'' or ``\emph{Agent\_$X$,}'' followed by a
        clause, where $X \in \{0,1,2\}$.
  \item \textbf{Acknowledgement.} \emph{ok, okay, got it, copy,
        acknowledged, agreed, sounds good, on it, will do, roger}.
  \item \textbf{Question.} A question mark outside coordinate tokens.
  \item \textbf{Politeness.} \emph{please, thanks, thank you, ty}.
  \item \textbf{Urgency.} \emph{now, urgent, urgently, immediately,
        asap, critical, critically, hurry, quickly, fast,
        right now}.
  \item \textbf{Plan / future-tense word.} \emph{will, going to,
        gonna, plan, planning, next, then, after, once, when}, plus
        future-tense modal forms (\emph{i'll, we'll, i will, we will}).
  \item \textbf{Past-tense / memory word.} \emph{tried, failed, was,
        were, did, didn't, couldn't, earlier, before, already, just,
        noticed, saw}.
  \item \textbf{Coordinate.} A token matching
        \texttt{\textbackslash(?\textbackslash s*-?\textbackslash d+\textbackslash s*,\textbackslash s*-?\textbackslash d+\textbackslash s*\textbackslash)?},
        i.e.\ a pair of integers possibly wrapped in parentheses,
        such as ``$25,23$'' or ``$(0,-1)$''.
  \item \textbf{Self-reference density.} Count of \emph{i, my, me,
        i'm, i'll, mine} tokens, normalised per message.
\end{itemize}
The absolute values should be interpreted as proxy measurements, since they depend on the chosen lexicons. The main comparisons below are more important: the gaps are stable across seeds and robust to reasonable variants of the detectors.

\subsection{Quantitative results}
\label{app:comm_quant}

Table~\ref{tab:comm_clean} groups the results into structural style, speech-act and pragmatic markers, and content turnover between consecutive messages.

\begin{table}[t]
  \centering
  \small
  \captionsetup{position=top,skip=6pt}
  \caption{Inter-agent message style across all records (mean~$\pm$~std
  across runs; Gemini $n=10$, Gemma $n=20$). Top: structural style.
  Middle: speech-act and pragmatic markers. Bottom: content and turnover.}
  \label{tab:comm_clean}
  \begin{tabular}{lrr}
    \toprule
    & Gemini~3.1~Pro & Gemma \\
    \midrule
    \multicolumn{3}{l}{\emph{Structural style}} \\
    message characters                  & $132.4 \pm 16.0$ & $104.1 \pm 7.9$ \\
    message words                       & $28.2 \pm 3.8$   & $20.9 \pm 1.5$  \\
    sentences / message                 & $3.22 \pm 0.40$  & $2.60 \pm 0.16$ \\
    multi-sentence ($\geq 2$), \% of msgs & $93.7 \pm 2.0$  & $87.9 \pm 4.5$ \\
    \midrule
    \multicolumn{3}{l}{\emph{Speech-act and pragmatic markers}} \\
    contains imperative verb, \% of msgs           & $73.7 \pm 6.6$ & $52.2 \pm 7.3$ \\
    imperative verbs / message                     & $1.80 \pm 0.41$ & $1.06 \pm 0.20$ \\
    contains broadcast phrasing (let's / we), \%   & $12.8 \pm 4.0$ & $9.7 \pm 3.7$  \\
    addresses specific teammate (``Agent~$X$,''), \% & $43.4 \pm 7.0$ & $44.1 \pm 8.4$ \\
    contains acknowledgement, \%                   & $5.3 \pm 1.9$  & $6.1 \pm 2.2$  \\
    contains a question, \%                        & $1.6 \pm 0.4$  & $1.7 \pm 0.7$  \\
    contains an exclamation mark, \%               & $57.7 \pm 7.1$ & $43.8 \pm 7.9$ \\
    contains \emph{please/thanks}, \%              & $14.0 \pm 3.1$ & $20.9 \pm 4.8$ \\
    contains urgency word (\emph{now, urgent,} \dots), \% & $14.1 \pm 2.5$ & $21.2 \pm 3.2$ \\
    \midrule
    \multicolumn{3}{l}{\emph{Content and turnover}} \\
    contains coordinate, \% of msgs                & $56.5 \pm 10.0$ & $66.8 \pm 10.7$ \\
    contains future-tense / plan word, \% of msgs & $59.1 \pm 5.3$  & $27.0 \pm 3.8$  \\
    contains past-tense / memory word, \% of msgs & $8.5 \pm 1.4$   & $6.7 \pm 1.7$   \\
    self-references (\emph{I, my, me}) / message  & $1.65 \pm 0.18$ & $0.92 \pm 0.17$ \\
    char similarity vs.\ previous message          & $0.39 \pm 0.02$ & $0.48 \pm 0.02$ \\
    Jaccard vs.\ previous message                  & $0.27 \pm 0.01$ & $0.32 \pm 0.03$ \\
    verbatim same as previous message, \%          & $0.5 \pm 1.0$   & $2.0 \pm 2.3$   \\
    \bottomrule
  \end{tabular}
\end{table}

Three patterns emerge.

\paragraph{Gemini communicates more like a planner.}
Gemini sends longer and more directive messages than Gemma. Its messages contain about $27\%$ more characters, about $24\%$ more sentences, and more imperative verbs per message ($1.80$ vs.\ $1.06$). They also contain more self-references ($1.65$ vs.\ $0.92$ per message). In the trajectories, this often looks like a model maintaining a team plan: Gemini states what it is doing, assigns actions to particular teammates, and marks when the next coordinated step should happen. Gemma's messages are shorter and more often read as local status updates.

\paragraph{The channel is mostly used to broadcast intent, not to deliberate.}
The two models differ in tone. Gemini uses direct commands and exclamation marks more often. Gemma more often uses polite or urgent language: \emph{please/thanks} appears in $20.9\%$ of Gemma messages compared with $14.0\%$ for Gemini, and urgency words appear in $21.2\%$ compared with $14.1\%$. The shared pattern is equally important. Both models address a specific teammate at similar rates ($43.4\%$ vs.\ $44.1\%$), while questions are rare for both ($1.6\%$ and $1.7\%$). Thus, agents mainly use communication to announce intent, request help, and assign actions, rather than to conduct back-and-forth discussion.

\paragraph{Communication style mirrors scratchpad style.}
The message channel reflects the same model-specific tendencies seen in the scratchpads. Planning words appear in $59.1\%$ of Gemini messages but only $27.0\%$ of Gemma messages; in scratchpads, the corresponding rates are $75.3\%$ and $36.6\%$. Both models mention coordinates less often in their messages than in their scratchpads. Still, the overall pattern is consistent. Gemini uses its scratchpad as a private plan and its messages as outward-facing instructions. Gemma uses its scratchpad more like a state snapshot and its messages more like status reports or requests for help.

\subsection{Qualitative examples}
\label{app:comm_qual}

\paragraph{How we sampled the examples.}
The excerpts below are illustrative, but we sample them with fixed criteria. We compare the same role in both models, agent~$0$ (the warrior), so available actions, crafting penalties, and team responsibilities are held constant. We also distinguish between controlled steps, where the agent has a real action choice, and forced \texttt{Noop} steps, where the agent is asleep, resting, or dead, but the harness still asks for a message.

Given these criteria, we take the first qualifying examples in chronological order from Gemini run~00 and Gemma run~00, rather than choosing messages that best fit the table. The only exception is the late HP-critical Gemma message: run~00 does not contain such a forced step, so we use the first matching example from Gemma run~06 at step~700. The Gemini box contains two controlled steps and one forced step; the Gemma box contains one controlled step and two forced steps.

\definecolor{geminiblue}{HTML}{1F6FEB}
\definecolor{geminitint}{HTML}{E8F1FD}
\definecolor{gemmagreen}{HTML}{1F8B4C}
\definecolor{gemmatint}{HTML}{E8F5EE}

\begin{tcolorbox}[
  enhanced,
  colback=geminitint,
  colframe=geminiblue,
  arc=2pt,
  boxrule=0.6pt,
  left=8pt,right=8pt,top=6pt,bottom=6pt,
  title={\textbf{Gemini~3.1~Pro}: run 00, agent 0
         (warrior)},
  fonttitle=\small\bfseries,
  coltitle=white,
  colbacktitle=geminiblue,
  attach boxed title to top left={yshift=-2mm,xshift=4mm},
  boxed title style={arc=2pt,boxrule=0pt}
]
\footnotesize
\textbf{Step 1} \;\;\texttt{ctrl} \;\;action: \texttt{Move East}\\[1pt]
\emph{``I chopped 25,23. Let's do the 3-agent tree at 27,23! I'm moving
to its West side (26,23). Agent 0, go to its North side (27,22). Agent
2, go to its South side (27,24). Face the tree and wait, we will `Do'
together when everyone is ready.''}

\medskip
\textbf{Step 10} \;\;\texttt{ctrl} \;\;action: \texttt{Make Wood Pickaxe}\\[1pt]
\emph{``My wood pickaxe craft failed due to the warrior penalty! Trying
again this turn. Once I get it, I will make a wood sword and then join
you for the 3-agent stone.''}

\medskip
\textbf{Step 230} \;\;\texttt{forced} \;\;action: \texttt{Noop} \;\;hp $0$\\[1pt]
\emph{``Agent 2, your health is 1! You are about to die! Drink a potion
IMMEDIATELY to heal if you have one! Place stone to block the orc!''}
\end{tcolorbox}

\vspace{2pt}

\begin{tcolorbox}[
  enhanced,
  colback=gemmatint,
  colframe=gemmagreen,
  arc=2pt,
  boxrule=0.6pt,
  left=8pt,right=8pt,top=6pt,bottom=6pt,
  title={\textbf{Gemma~4}: runs 00 \& 06, agent 0
         (warrior)},
  fonttitle=\small\bfseries,
  coltitle=white,
  colbacktitle=gemmagreen,
  attach boxed title to top left={yshift=-2mm,xshift=4mm},
  boxed title style={arc=2pt,boxrule=0pt}
]
\footnotesize
\textbf{Run 00, step 0} \;\;\texttt{ctrl} \;\;action: \texttt{Move
North}\\[1pt]
\emph{``Agent 1: face North and Do the tree at (25, 23). Agent 2: move
to the stone at (20, 23). I'm moving North to (24, 23) to help gather
wood.''}

\medskip
\textbf{Run 00, step 100} \;\;\texttt{forced} \;\;hp $5$\\[1pt]
\emph{``Almost there, thanks for waiting.''}

\medskip
\textbf{Run 06, step 700} \;\;\texttt{forced} \;\;hp $3$\\[1pt]
\emph{``Agent 1, please prioritize me for food as well, I am at 0 and
my health is critical (3). Staying put.''}
\end{tcolorbox}

These examples make the lexical patterns easier to interpret. Gemini's opening message contains a complete coordination proposal: it names the target, assigns positions to each teammate, and specifies when the joint \texttt{Do} action should happen. Its later messages keep this outward-facing style, either explaining the next step in its own plan or giving urgent instructions to another agent. Gemma's opening message also assigns roles and locations, showing that it can use the channel for coordination. However, its forced-step messages better reflect its aggregate style: concise status updates, polite acknowledgements, and requests tied to local health or resource needs.

\section{Functional coding of messages, scratchpads, and reasoning}
\label{app:functional-traces}

Apps.~\ref{app:memory_scratchpad}--\ref{app:comm_style} describe the agents' text with lexical detectors, such as plan words, coordinates, and imperative verbs. These detectors are simple to reproduce, but the presence of a word such as \emph{next} does not show that an entry contains a plan. This appendix instead codes what each piece of text does: whether a message proposes a joint target, asks a teammate to act, or coordinates timing, and whether a scratchpad entry or reasoning trace lays out a plan or tracks what a teammate is doing. Sec.~\ref{subsec:q2_ablate_mem_coms_reason} quotes the results.

\subsection{Setup}
\label{app:func_setup}

\paragraph{Traces.}
We use the full-harness evaluation runs of Gemini-3.1-Pro-High and Gemma-4-31B-it from Sec.~\ref{subsec:q1} on all three difficulties: $10$ Gemini episodes and $20$ Gemma episodes per difficulty. The Hard runs are the trajectories analysed in Apps.~\ref{app:memory_scratchpad}--\ref{app:comm_style}. Each logged agent step contains the message the agent broadcast to its teammates, its scratchpad entry, and its reasoning. For Gemini, the reasoning is the thought summary returned by the API; for Gemma, it is the full reasoning text.

\paragraph{Sampling.}
We sample only controlled agent steps, i.e.\ steps on which the agent is not asleep, resting, or dead. On the other steps, the harness forces a Noop and copies the previous scratchpad, so the text does not reflect a decision by the model. For each model and difficulty, we draw $400$ controlled steps, split equally across episodes ($40$ per Gemini episode and $20$ per Gemma episode; seed~$0$). One Gemini Easy episode failed, with an empty model output on every step, so we exclude it and draw $44$ steps from each of the remaining nine episodes. This gives $2{,}396$ steps ($1{,}196$ for Gemini and $1{,}200$ for Gemma). Each non-empty channel of a step becomes one item, giving $7{,}167$ items: $2{,}375$ messages, $2{,}396$ scratchpad entries, and $2{,}396$ reasoning traces.

\paragraph{Blinding.}
The annotator sees an opaque item identifier, the channel, the author's agent index and role (e.g.\ ``Agent~1 (forager)''), and the text. It does not see the model, difficulty, episode, or step. Before annotation, we mask model names and any number attached to the word \emph{reward} ($117$ spans in total), and we check that no such string remains anywhere in the corpus. Items are shuffled before batching, so each request mixes models, difficulties, and channels.

\paragraph{Annotator.}
We use Gemini~3~Flash (\texttt{gemini-3-flash-preview}) through the Gemini API, with the thinking level set to low and the default temperature. Each request contains up to $8$ items and asks for a true or false value for every code on every item, returned as JSON that must match a fixed schema. We reject and retry any response that does not return every item identifier with a boolean for every code. All $896$ requests succeeded. Besides the code definitions in Table~\ref{tab:func_codebook}, the prompt describes the game in one sentence and gives the following rules: code each item on its own text only; a code applies only when the text performs that function, and mentioning a word is not enough; an ordered plan needs at least two concrete future steps after the current action; plans count whether they are for the agent alone or for the team; coordinates such as $(21, 23)$ are map positions, and ``T9'' means game step~$9$.

\paragraph{Codebook.}
Table~\ref{tab:func_codebook} lists the nine codes, grouped into three capabilities. A capability is present in an item if any of its codes is, and the codes are not mutually exclusive. \texttt{teammate\_reference} records only that a teammate is named, so it counts toward no capability.

\begin{table}[t]
  \centering
  \small
  \captionsetup{position=top,skip=6pt}
  \caption{Codebook used for annotation. Each code is a function that the text performs, not a word that it contains.}
  \label{tab:func_codebook}
  \begin{tabular}{@{}l p{0.68\linewidth}@{}}
  \toprule
  Code & Definition given to the annotator \\
  \midrule
  \multicolumn{2}{@{}l}{\emph{Theory of mind}} \\
  \texttt{teammate\_state} & Attributes a current state, need or limitation to a teammate, beyond restating their coordinates: e.g. low health, needs wood, is asleep or dead, cannot see the map, is out of range of a target. \\
  \texttt{teammate\_intent} & Infers or predicts what a teammate wants, knows, believes or will do next, from their messages or behaviour (e.g. `Agent 1 is heading to the tree', `A2 probably hasn't seen the zombie'). \\
  \texttt{teammate\_commitment} & Records, restates or acknowledges something a teammate has agreed or committed to do (e.g. `A1 will Build at T12', `Agent 2 said they will bring stone'). \\
  \midrule
  \multicolumn{2}{@{}l}{\emph{Shared intent}} \\
  \texttt{joint\_goal} & Proposes, accepts or refers to a goal or target that two or more agents pursue together (e.g. `let's all Do the tree', `we need to build the beacon together', `OK, I'll join you at the site'). \\
  \texttt{action\_assignment} & Asks or tells a specific teammate to take a specific action or role (e.g. `Agent 1, start the Build', `you mine, I'll craft'). \\
  \texttt{timing\_alignment} & Coordinates when actions happen relative to teammates: on the same turn, on a signal, within a window, or after another agent acts (e.g. `Do on the same turn', `complete it within 6 steps'). \\
  \midrule
  \multicolumn{2}{@{}l}{\emph{Multi-step planning}} \\
  \texttt{ordered\_plan} & Lays out, in order, at least two concrete future actions or sub-goals beyond the action being taken now, for the agent itself or the team (e.g. `move west, move west, then Do', `T9: Build, T10: wait for A2, T11: Do', `after drinking, go to the plant, then find the ladder'). Not this code: the current action plus a goal or direction (`Moving North to search for the ladder', `Next pos (27, 25)'). \\
  \texttt{conditional\_plan} & States a contingency for the future: if or when X happens, do Y (e.g. `if A2 doesn't arrive by T15, I'll go alone'). \\
  \midrule
  \multicolumn{2}{@{}l}{\emph{No capability}} \\
  \texttt{teammate\_reference} & Names or clearly refers to a specific teammate (e.g. `Agent 2', `A1', `the miner'). Listing a teammate's coordinates alone is only this code. \\
  \bottomrule
  \end{tabular}
\end{table}

\paragraph{Statistics.}
The episode is the resampling unit. For each model and quantity, we compute one rate per episode and arrange these rates in an (episodes $\times$ difficulties) matrix. We report 95\% percentile intervals from the stratified bootstrap of \citet{agarwal2021deep}, which resamples episodes within each difficulty ($50{,}000$ replicates). For differences between models, the two models' episodes are resampled independently. Because every episode contributes the same number of sampled steps, each estimate equals the pooled percentage of steps, and nearly equals the pooled percentage of items, since a few steps have no message. The plan-carried rate is computed only over steps whose reasoning contains a plan: for each episode, we take the share of these steps whose scratchpad entry or message also contains a plan, and then average over episodes. The main text pools all three difficulties; Table~\ref{tab:func_main} also reports Hard alone, the setting of the ablations.

\paragraph{Reliability.}
To measure how consistently the annotator applies the codebook, we annotate a random $10\%$ of the items ($717$) a second time in new batches, so that each item is coded alongside different neighbours. Table~\ref{tab:func_kappa} reports Cohen's $\kappa$ between the two passes. Agreement is at least $\kappa = 0.66$ for every code, and between $0.70$ and $0.85$ for the codes quoted in the main text. This measures the consistency of a single annotator, not its agreement with human coders.

\begin{table}[t]
  \centering
  \small
  \captionsetup{position=top,skip=6pt}
  \caption{Test-retest reliability on $717$ items annotated twice, in different batches. ``Positive'' is the percentage of items for which the first pass assigns the code.}
  \label{tab:func_kappa}
  \begin{tabular}{@{}l rrr@{}}
  \toprule
  Code or capability & Cohen's $\kappa$ & Agreement (\%) & Positive (\%) \\
  \midrule
  \texttt{teammate\_reference} & 0.97 & 99.2 & 80.2 \\
  \texttt{teammate\_state} & 0.75 & 88.3 & 38.6 \\
  \texttt{teammate\_intent} & 0.75 & 89.0 & 33.3 \\
  \texttt{teammate\_commitment} & 0.66 & 95.1 & 7.9 \\
  \texttt{joint\_goal} & 0.77 & 90.4 & 30.7 \\
  \texttt{action\_assignment} & 0.85 & 94.8 & 22.3 \\
  \texttt{timing\_alignment} & 0.80 & 92.7 & 23.6 \\
  \texttt{ordered\_plan} & 0.74 & 87.0 & 49.7 \\
  \texttt{conditional\_plan} & 0.70 & 91.6 & 17.0 \\
  \midrule
  Theory of mind (any code) & 0.82 & 90.9 & 54.4 \\
  Shared intent (any code) & 0.77 & 88.7 & 46.7 \\
  Multi-step planning (any code) & 0.73 & 86.5 & 56.1 \\
  \bottomrule
  \end{tabular}
\end{table}

\subsection{Results}
\label{app:func_results}

Table~\ref{tab:func_channels} reports every code in every channel. Three patterns stand out.

\paragraph{Coordination requests are made in messages.}
Agents rarely use the scratchpad to direct teammates. Action assignments appear in $41.1\%$ of Gemini's messages and $31.1\%$ of Gemma's, but in only $4.6\%$ and $2.0\%$ of their scratchpad entries. Joint targets and timing also appear in the scratchpads, but less often than in messages (for Gemini, $24.6\%$ versus $39.1\%$ for joint targets and $21.7\%$ versus $35.7\%$ for timing).

\paragraph{The planning gap is largest in the channels that persist.}
Gemini plans more often than Gemma in every channel, but the gap is about twice as large in the scratchpad ($+39.3$ points) and in messages ($+34.9$) as in reasoning ($+17.3$). Both models plan in most of their reasoning traces; they differ mainly in whether a plan also reaches the scratchpad, where it persists across turns, or the messages, where teammates can see it.

\paragraph{Both models name teammates equally often but say different things about them.}
The two models name a specific teammate at similar rates in messages ($64.7\%$ and $61.4\%$) and scratchpads ($72.1\%$ and $75.2\%$), in line with the lexical results in Apps.~\ref{app:memory_scratchpad} and~\ref{app:comm_style}. Gemini more often describes a teammate's state, both in messages ($27.6\%$ versus $14.4\%$) and in scratchpads ($36.5\%$ versus $28.0\%$), and more often records what a teammate intends to do in its scratchpad ($28.4\%$ versus $20.1\%$).

\begin{table}[t]
  \centering
  \small
  \captionsetup{position=top,skip=6pt}
  \caption{Percentage of items in each channel that carry each code, pooled over all three difficulties (Gemini-3.1-Pro-High and Gemma-4-31B-it). $\Delta$ is Gemini minus Gemma, in percentage points; differences whose 95\% interval includes zero are shown in grey. ``Any of the above'' is the capability.}
  \label{tab:func_channels}
  \begin{adjustbox}{max width=\linewidth}
  \begin{tabular}{@{}l rrr rrr rrr@{}}
  \toprule
   & \multicolumn{3}{c}{Message} & \multicolumn{3}{c}{Scratchpad} & \multicolumn{3}{c}{Reasoning} \\
  \cmidrule(lr){2-4}\cmidrule(lr){5-7}\cmidrule(lr){8-10}
  Code & Gemini & Gemma & $\Delta$ & Gemini & Gemma & $\Delta$ & Gemini & Gemma & $\Delta$ \\
  \midrule
  \texttt{teammate\_state} & 27.6 & 14.4 & $+13.2$ & 36.5 & 28.0 & $+8.5$ & 65.7 & 56.2 & $+9.4$ \\
  \texttt{teammate\_intent} & 5.9 & 2.4 & $+3.5$ & 28.4 & 20.1 & $+8.3$ & 67.1 & 67.6 & \textcolor{gray}{$-0.5$} \\
  \texttt{teammate\_commitment} & 2.7 & 1.8 & \textcolor{gray}{$+1.0$} & 3.5 & 1.9 & $+1.6$ & 21.9 & 18.3 & \textcolor{gray}{$+3.6$} \\
  \emph{any of the above} & 31.8 & 17.6 & $+14.2$ & 51.9 & 41.4 & $+10.5$ & 86.8 & 83.8 & \textcolor{gray}{$+3.0$} \\
  \midrule
  \texttt{joint\_goal} & 39.1 & 30.6 & $+8.5$ & 24.6 & 18.6 & $+6.1$ & 38.2 & 42.9 & \textcolor{gray}{$-4.8$} \\
  \texttt{action\_assignment} & 41.1 & 31.1 & $+10.0$ & 4.6 & 2.0 & $+2.6$ & 30.7 & 22.8 & $+8.0$ \\
  \texttt{timing\_alignment} & 35.7 & 21.7 & $+14.0$ & 21.7 & 11.2 & $+10.5$ & 32.3 & 23.9 & $+8.4$ \\
  \emph{any of the above} & 63.9 & 53.8 & $+10.1$ & 33.9 & 23.4 & $+10.5$ & 54.8 & 54.2 & \textcolor{gray}{$+0.6$} \\
  \midrule
  \texttt{ordered\_plan} & 50.6 & 20.0 & $+30.6$ & 68.7 & 32.5 & $+36.2$ & 79.1 & 60.2 & $+19.0$ \\
  \texttt{conditional\_plan} & 19.6 & 9.6 & $+10.0$ & 17.9 & 3.5 & $+14.4$ & 28.3 & 21.2 & $+7.2$ \\
  \emph{any of the above} & 63.0 & 28.1 & $+34.9$ & 73.5 & 34.2 & $+39.3$ & 83.6 & 66.3 & $+17.3$ \\
  \midrule
  \texttt{teammate\_reference} & 64.7 & 61.4 & \textcolor{gray}{$+3.2$} & 72.1 & 75.2 & \textcolor{gray}{$-3.1$} & 94.2 & 97.9 & $-3.7$ \\
  \bottomrule
  \end{tabular}
  \end{adjustbox}
\end{table}

\paragraph{Hard only.}
Table~\ref{tab:func_main} repeats the quantities quoted in the main text for Hard alone, the setting of the ablations. Every difference has the same sign as in the pooled data. With a third of the sample, the gaps in joint targets and action assignments in messages are no longer distinguishable from zero, while the gap in timing still is.

\begin{table}[t]
  \centering
  \small
  \captionsetup{position=top,skip=6pt}
  \caption{Quantities quoted in Sec.~\ref{subsec:q2_ablate_mem_coms_reason}, pooled over all three difficulties and on Hard alone. $\Delta$ is Gemini minus Gemma, in percentage points, with its 95\% stratified bootstrap interval; differences whose interval includes zero are shown in grey. $^{a}$Among steps whose reasoning contains a plan, the percentage whose scratchpad entry or message also contains one (not necessarily the same plan).}
  \label{tab:func_main}
  \begin{adjustbox}{max width=\linewidth}
  \begin{tabular}{@{}l rrl rrl@{}}
  \toprule
   & \multicolumn{3}{c}{All three difficulties} & \multicolumn{3}{c}{Hard only} \\
  \cmidrule(lr){2-4}\cmidrule(lr){5-7}
  Quantity (\% of items) & Gemini & Gemma & $\Delta$ [95\% CI] & Gemini & Gemma & $\Delta$ [95\% CI] \\
  \midrule
  Message: joint target & 39.1 & 30.6 & $+8.5$ {\scriptsize$[3.6, 13.6]$} & 38.6 & 32.2 & \textcolor{gray}{$+6.4$ {\scriptsize$[-3.0, 16.5]$}} \\
  Message: action assignment & 41.1 & 31.1 & $+10.0$ {\scriptsize$[5.7, 14.4]$} & 39.8 & 32.0 & \textcolor{gray}{$+7.9$ {\scriptsize$[0.0, 16.0]$}} \\
  Message: timing & 35.7 & 21.7 & $+14.0$ {\scriptsize$[9.7, 18.5]$} & 36.8 & 25.1 & $+11.7$ {\scriptsize$[3.3, 20.8]$} \\
  Scratchpad: ordered plan & 68.7 & 32.5 & $+36.2$ {\scriptsize$[31.1, 41.3]$} & 65.0 & 36.3 & $+28.7$ {\scriptsize$[19.2, 38.8]$} \\
  Scratchpad: contingency plan & 17.9 & 3.5 & $+14.4$ {\scriptsize$[11.6, 17.1]$} & 17.5 & 3.0 & $+14.5$ {\scriptsize$[9.8, 19.5]$} \\
  Reasoning: any plan & 83.6 & 66.3 & $+17.3$ {\scriptsize$[13.3, 21.3]$} & 82.2 & 67.2 & $+15.0$ {\scriptsize$[7.7, 22.5]$} \\
  Plan carried from reasoning$^{a}$ & 88.8 & 60.3 & $+28.5$ {\scriptsize$[24.2, 32.9]$} & 84.8 & 62.0 & $+22.8$ {\scriptsize$[14.4, 31.6]$} \\
  \bottomrule
  \end{tabular}
  \end{adjustbox}
\end{table}

\paragraph{Examples.}
Table~\ref{tab:func_examples} shows annotated items. We chose them by hand to illustrate the codes, so they are not a random sample.

\begin{table}[t]
  \centering
  \small
  \captionsetup{position=top,skip=6pt}
  \caption{Annotated items, with every code that the annotator assigned. The last item shows the boundary of \texttt{ordered\_plan}: the current action plus a goal is not an ordered plan.}
  \label{tab:func_examples}
  \begin{tabular}{@{}p{0.62\linewidth} p{0.33\linewidth}@{}}
    \toprule
    Item & Codes assigned \\
    \midrule
    \emph{Gemini, message:} ``Eating cow this turn! I will head to 24,17 next to join you guys for the 3-agent stone. Will take the East or North position.''
      & \parbox[t]{\linewidth}{\raggedright \texttt{joint\_goal}, \texttt{ordered\_plan}} \\
    \addlinespace
    \emph{Gemma, message:} ``Agent 1, please clear the stone at (26, 19) now. Once it's gone, I'll move up to (26, 19) and wait for Agent 0 at (25, 18) for the sync on (26, 18).''
      & \parbox[t]{\linewidth}{\raggedright \texttt{teammate\_reference}, \texttt{joint\_goal}, \texttt{action\_assignment}, \texttt{timing\_alignment}, \texttt{ordered\_plan}} \\
    \addlinespace
    \emph{Gemma, message:} ``Agent 1, I'm still here at (30, 29) facing East. Please let me know exactly when you reach (32, 29) and face West. Once you're in position, we will sync the handover. Agent 0, still waiting on Agent 1 to arrive for the iron.''
      & \parbox[t]{\linewidth}{\raggedright \texttt{teammate\_reference}, \texttt{joint\_goal}, \texttt{action\_assignment}, \texttt{timing\_alignment}, \texttt{conditional\_plan}} \\
    \addlinespace
    \emph{Gemini, scratchpad:} ``Agent 2 is at 18,14, going to face North to target cow at 18,13. / I am at 17,11. / Turn 1: Move South to 17,12 (current) / Turn 2: Move South to 17,13 / Turn 3: Move East to face 18,13 / Turn 4: Do (sync attack)''
      & \parbox[t]{\linewidth}{\raggedright \texttt{teammate\_reference}, \texttt{teammate\_intent}, \texttt{joint\_goal}, \texttt{timing\_alignment}, \texttt{ordered\_plan}} \\
    \addlinespace
    \emph{Gemini, scratchpad:} ``Moving to 27,36. Next turn I'll move South to 27,37 to see if there is a plant at 27,38. If not, I'll ask A1 to Give. If A1 Gives me a sapling, I'll ascend and plant it if Place Plant fails here. I need food badly.''
      & \parbox[t]{\linewidth}{\raggedright \texttt{teammate\_reference}, \texttt{ordered\_plan}, \texttt{conditional\_plan}} \\
    \addlinespace
    \emph{Gemma, scratchpad:} ``Pos: (20, 5) $\to$ (20, 4). Agent 0 at (19, 4). Agent 1 at (21, 13). Searching for ladder on x=20.''
      & \parbox[t]{\linewidth}{\raggedright \texttt{teammate\_reference}} \\
    \bottomrule
  \end{tabular}
\end{table}

\subsection{Limitations}
\label{app:func_limits}

\begin{itemize}\itemsep2pt
  \item \textbf{One annotator, no human validation.} Test-retest agreement shows that the annotator applies the codebook consistently, not that it agrees with human coders. We read samples of its output while developing the codebook, but did not measure agreement with human annotations.
  \item \textbf{Annotator family.} The annotator belongs to the same model family as Gemini-3.1-Pro-High. Masking hides model names, but writing style may still reveal the author. The codes describe what a text does rather than how good it is, which limits the scope for self-preference, but we cannot rule it out.
  \item \textbf{Reasoning traces differ in kind.} Gemini's reasoning traces are API thought summaries, whereas Gemma's are its full reasoning, so a summary may omit or reorganise content that the full reasoning contains. Comparisons between the models within the reasoning channel should be read with this in mind.
  \item \textbf{Co-occurrence, not tracking.} The plan-carried rate checks whether a plan appears in the scratchpad or messages on the same step, not whether it is the same plan.
  \item \textbf{Imperfect masking.} A reward value phrased loosely, without the word \emph{reward} next to it, can survive the mask.
  \item \textbf{Full-harness runs only.} We did not annotate the ablation runs. The results therefore describe how each model uses a component when it is available, and they offer an explanation of the ablation effects rather than a direct measurement of what changes when the component is removed.
\end{itemize}

\clearpage
\section{Human Play Details}\label{app:human_play}

The human rows of Table~\ref{tab:alem_results} come from the human play interface we built for \textsc{alem}; the interface and session runner come with the environment (\texttt{examples/play\_with\_human\_interface.py}).

\begin{figure}[h]
  \centering
  \includegraphics[width=\linewidth]{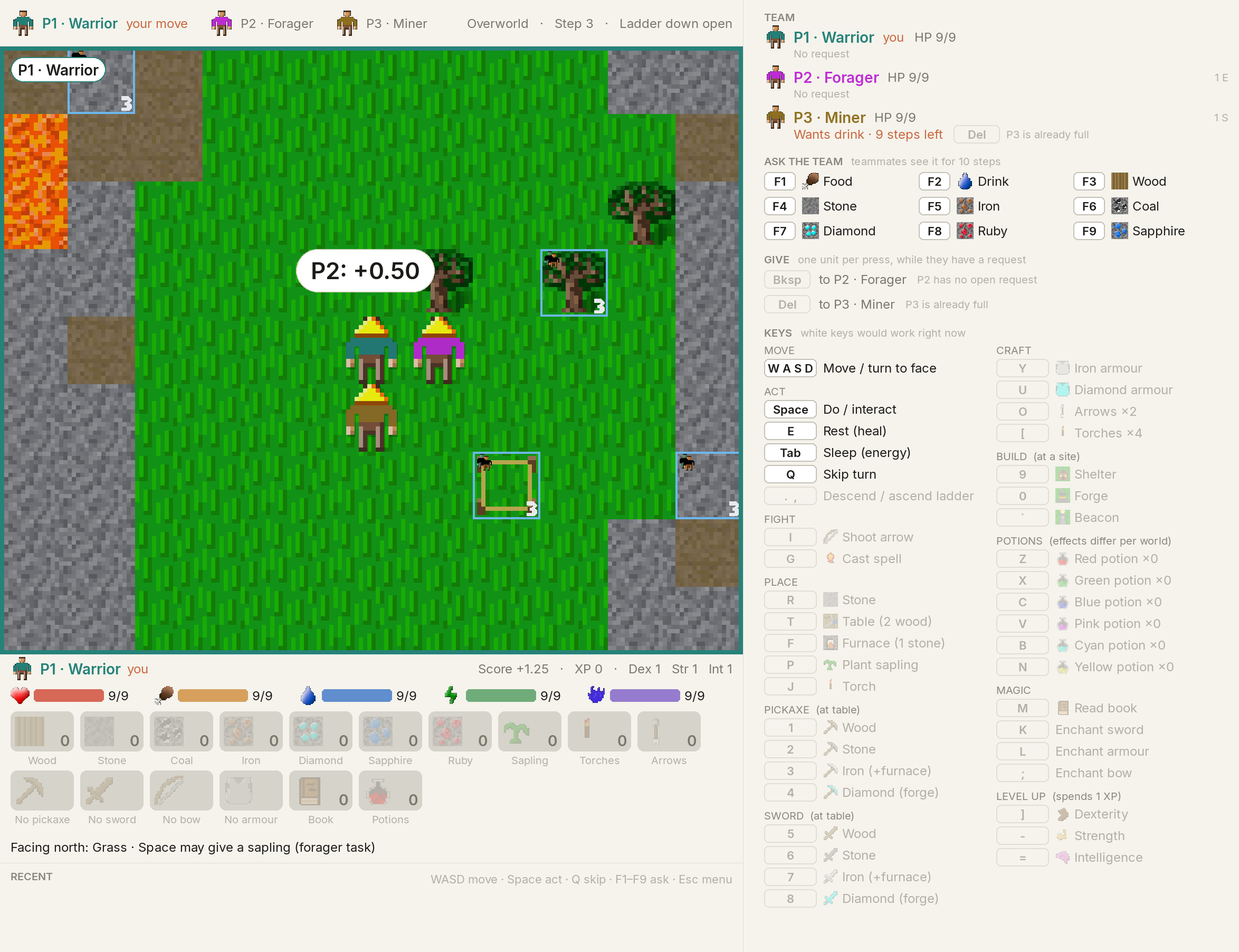}
  \caption{The human-play interface. Left: the agents' egocentric map view, framed in the colour of the player whose turn it is; below it, that player's stat bars, inventory and a description of the faced tile. Right: the team's health and open requests, the request and give keys, and every other key, highlighted when its action would succeed this step.}
  \label{fig:human_interface}
\end{figure}

\paragraph{Interface.}
Figure~\ref{fig:human_interface} shows the screen. The map is the same $9\times11$ egocentric view the pixel renderer produces, centred on the player whose turn it is, including the coordination borders and elite-creature markings. Around it we expose, in labelled form, the information the text observation gives LLM agents and nothing more: the player's stats and inventory, each teammate's health, position and open resource request, and a one-line description of the faced tile (its coordination rule, the pickaxe tier it needs, whether a ladder is open, or the attack rule for an elite creature). The sidebar lists every key and marks those whose action would succeed on the current step using the environment's legal-action mask, so players do not need to memorise recipes. No hidden state is shown.

\paragraph{Protocol.}
Three authors shared one keyboard, each controlling one player, acting in turn; the world advances once all three have chosen, so one environment step is one round of three decisions, as for the agents. Talking was allowed throughout, as the human counterpart of the agents' communication channel. Each session used the evaluation settings (three players, soft specialisation, individual reward, the difficulty preset, a cap of 10{,}000 steps) and the first evaluation world, seed 9999, so the team played the same world as the agents' first episode. Before playing, the team read the same \texttt{<game\_rules>} briefing the LLM agents receive. There was no practice round. The Easy session, played under a one-hour limit, ended when all three players had died, after 478 steps and having reached the second dungeon floor; the Hard session, played under a two-hour limit, was stopped by the team after 455 steps, also on the second floor with no deaths. Both episodes were scored as played with the same \texttt{compute\_score} function as the agents. Every decision, its latency and the full environment state per step were logged in the agent trajectory format, so both sessions can be replayed.

\clearpage
\section{Privileged Heuristic Policy Details}\label{app:oracle}

The privileged heuristic row of Table~\ref{tab:alem_results} is a hand-designed, centralised policy written to check how much of the nominal maximum reward is attainable at all. It reads the complete environment state, including the whole map, all creature positions and health, the coordination requirements of every block and creature, and the potion identity mapping, and it chooses the joint action of all three players. It acts only through the normal action interface under the evaluation settings: no god mode, no state edits, the same \texttt{EnvParams}, difficulty presets and 10{,}000-step cap as the agents. Every rule that binds an RL or LLM agent (tool tiers, role gates, coordination requirements, hunger and thirst, combat, kill-gated ladders) binds it too; what it removes is partial observability and the coordination problem, not the game.

\paragraph{Policy.}
The policy is a priority-ordered set of rule-based macros (navigate, mine, craft, place, request and give, fight, sleep, descend) over pathfinding on the true map. Its priorities were fixed by iterating on the evaluation seeds: equip every player with a stone pickaxe and sword and leave the overworld before its night spawns; fight every hostile three-on-one and never leave a player alone near creatures; hold position near a fountain on dungeon floors so kills, water and request-and-give provisioning stay close; attempt only those coordination tasks whose geometry admits the required number of adjacent players; and, late in the horizon, descend for the floor-entry achievements. Because the episode runner controls the random-key schedule, the policy also uses the simulator as an exact one-step model: it simulates its scripted joint action, and when the result is damage or a death it tries per-player alternatives (attack, dodge, hold) and commits the best simulated outcome. 
\clearpage

\newpage

\end{document}